\documentclass{article}

\usepackage{xcolor}
\usepackage[preprint]{corl_2026}

\usepackage{amsmath,amsfonts,amssymb,mathrsfs,amsthm}
\usepackage{mathtools}
\usepackage{enumitem}

\usepackage{tikz}
\usetikzlibrary{quotes,arrows.meta,angles,calc,3d,shapes,shadows,intersections,plotmarks,positioning,backgrounds,external}
\usepackage{nicefrac, subcaption}

\usepackage{array}
\usepackage{tabularx}
\usepackage{booktabs}
\usepackage{longtable}
\usepackage{multirow}
\usepackage{siunitx}
\newcolumntype{Y}{>{\arraybackslash}X}
\newcolumntype{L}[1]{>{\arraybackslash}p{#1}}
\usepackage{makecell}

\usepackage{microtype}
\usepackage{xspace}
\usepackage{wrapfig}
\usepackage{dsfont}

\hypersetup{colorlinks=true}
\usepackage[capitalize]{cleveref}
\crefname{paragraph}{\S}{\S\S}

\usepackage[most]{tcolorbox}

\newtcolorbox{questionbox}{
  colback=questionBoxBack,
  colframe=questionBoxFrame,
  boxrule=0.4pt,
  arc=2pt,
  left=6pt,right=6pt,
  top=3pt,bottom=3pt,
  boxsep=1pt,
  before skip=4pt,
  after skip=4pt,
  enhanced
}

\usetikzlibrary{backgrounds,fit}
\pgfdeclarelayer{underlay}
\pgfsetlayers{underlay,background,main}

\definecolor{questionBoxBack}{RGB}{248,248,246}
\definecolor{questionBoxFrame}{RGB}{112,116,122}

\usepackage{wrapfig}

\newtheorem{lemma}{Lemma}

\newtheorem{assumption}{Assumption}

\newcommand{\mathbbm}[1]{{\mathds{#1}}}
\newcommand{\R}{\mathbb{R}}

\newcommand{\E}{\mathbb{E}}

\newcommand{\TV}{\operatorname{TV}}

\newcommand{\freelabel}{\mathrm{free}}
\newcommand{\succlabel}{\mathrm{succ}}

\newcommand{\gtprior}{\chi_{\mathrm{GT}}}

\newcommand{\wspace}{\mathcal{W}}
\newcommand{\winst}{\xi}
\newcommand{\cspace}{\mathcal{X}}
\newcommand{\state}{x}

\newcommand{\traj}{\pi}
\newcommand{\object}{o}

\newcommand{\consem}[1]{\mathbbm{1}_{#1}}
\newcommand{\ind}[1]{\mathbbm{1}\left({#1}\right)}
\newcommand{\wsubsubset}{\mathcal{E}}

\newcommand{\wrand}{\mathbbm{W}}
\newcommand{\probspace}{\mathbb{P}}
\newcommand{\qspace}{\mathbb{Q}}

\newcommand{\gprobspace}{\bar{\probspace}}

\newcommand{\observ}{\Theta}

\newcommand{\cB}{\mathcal{B}}
\newcommand{\cP}{\mathcal{P}}
\newcommand{\cA}{\mathcal{A}}
\newcommand{\cS}{\mathcal{S}}

\newcommand{\cC}{\mathcal{C}}

\newcommand{\obs}{\mathrm{obs}}

\newcommand{\KL}{\mathcal{D}_{\mathrm{KL}}}

\newcommand{\mytexttilde}{\ensuremath{\sim}}

\newcommand{\Chamfer}{\mathrm{CD}}
\newcommand{\depthmap}{D}
\newcommand{\neardepth}{d_{\mathrm{near}}}
\newcommand{\fardepth}{d_{\mathrm{far}}}
\newcommand{\croparea}{A}
\newcommand{\imgarea}{A_{\mathrm{img}}}

\hypersetup{
  pdftitle={MatterDoor: Sampling Zero-shot Spatio-semantic Priors using Generative Models},
  pdfauthor={Subhransu S. Bhattacharjee, Hao Lu, Dylan Campbell, Rahul Shome}
}

\title{MatterDoor: Sampling Zero-shot Spatio-semantic Priors using Generative Models
}

\author{Subhransu~S.~Bhattacharjee$^{*}$, Hao~Lu, Dylan~Campbell, Rahul~Shome\\
School of Computing, Australian National University, Canberra, ACT 2600, Australia\\
\small $^{*}$Corresponding author:
\texttt{Subhransu.Bhattacharjee@anu.edu.au}.
}

\begin{document}
\maketitle

\vspace{-2.5em}
\begin{abstract}
Autonomous robots often view rooms only partially, through a doorway, where the walls and scene structure hide the geometry and task-relevant semantics needed for safe navigation and goal-directed action. We ask whether off-the-shelf pretrained generative vision models can derive this missing structure as zero-shot offline priors for robot reasoning. Such priors should support spatio-semantic queries over unobserved structure, estimating the target object likelihood in hidden regions and the probability that those regions are occupied. Given an egocentric RGB observation and target query, our pipeline uses VLM-guided outpainting, monocular depth estimation, and semantic segmentation to sample semantically labeled 3D point cloud hypotheses of the hidden room. We introduce \emph{MatterDoor}, a Matterport3D-derived benchmark of doorway-occluded indoor scenes, and evaluate the resulting priors with generative metrics and simulated Stretch robot object-reaching tasks. Our results suggest that useful spatio-semantic priors for planning can be derived without problem-specific fine-tuning.
\end{abstract}
\vspace{-0.5em}

\begin{figure}[!hb]
    \centering
    \includegraphics[
        width=0.835\linewidth,
        trim=30pt 0 25pt 0,
        clip
    ]{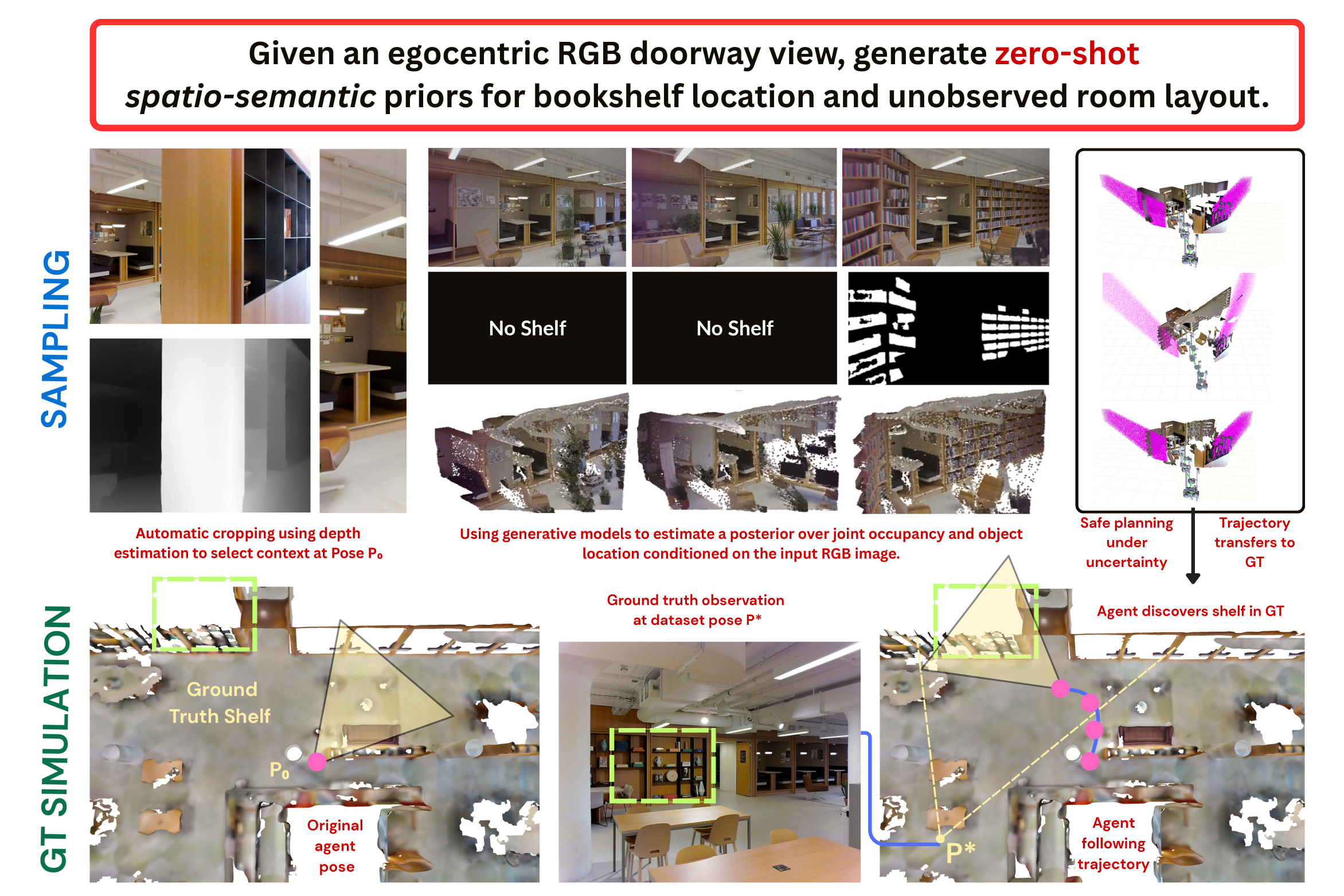}
\caption{\small \textbf{Overview.}
Given a partial doorway view and a shelf query, the agent plans zero-shot without ground-truth geometry or semantic annotations. The crop is lifted with monocular depth; a generative model samples hidden completions; and an image segmenter converts each completion into 3D free-space, obstacle, and target hypotheses. These sampled worlds define the belief for a sampling-based planner. Evaluation is against the ground-truth Matterport3D scene: the robot localizes the fully unobserved shelf within one meter, \textit{without} access to the true environment during offline planning in simulation.
}
    \label{fig:fig1}
\end{figure}

\section{Introduction}
\label{sec:intro}
Autonomous robots in cluttered indoor environments often need to reason under heavy occlusion, beyond what is directly visible from partial observations, because direct exploration can be costly~\cite{placed2023active}. An RGB view through a doorway may reveal room type and local traversability while hiding obstacles, free space, and goal locations. Planning therefore requires a belief over unseen semantic structure, not only a map of visible space~\cite{elfes1989occupancy,vandenberg2011lqgmp,silver2010pomcp,dimsam}. Existing priors are often handcrafted, domain-specific, or based on discrete priors~\cite{chad,assist}, tied to coarse workspace abstractions~\cite{unaware}, or difficult to use directly for configuration-space planning with geometric constraints and semantic goals~\cite{lu2024sampling,axelrod2018provably}.

Semantic robot planning, belief-space planning, and open-vocabulary exploration show that learned priors can improve robotic decision-making~\cite{toberg2024commonsense,zhao2025seeingbelievingbeliefspaceplanning,bogenberger2026glasses}. Recent generative vision models~\cite{rombach2022high,podell2024sdxl,saharia2022photorealistic,BFL2025Kontext} offer such priors beyond the visible scene by implicitly encoding layout and object regularities~\cite{bhattacharjee2025uod}. We study doorway-occluded room entry, where a robot must reason about target-object locations and hidden occupancy \emph{before entering}. From a partial RGB view, we infer an offline, zero-shot prior over the unobserved interior: where the target may lie and which regions may be occupied. Unlike deterministic completion~\cite{hartley2003multiple,Song2017SSCNet,mildenhall2021nerf,splatting}, we sample plausible spatio-semantic completions~\cite{ruizsarmiento2017multiversal,sakagami2023robotic,pipe}, forming an empirical belief for estimating target-reaching probability. 
Our results show that these generative priors improve spatial estimates of target semantics and occupancy in simulation, compared to non-generative and deterministic baselines, supporting sampling-based priors for robotic reasoning under uncertainty.

\textbf{Contributions.}
We make three contributions:
(i) a zero-shot pipeline that composes pretrained generative and discriminative models to produce labeled 3D point-cloud hypotheses from a single RGB doorway view (\cref{fig:fig1}), directly queryable for collision, visibility, target evidence, costs, and constraints;
(ii) \emph{MatterDoor}, a Matterport3D-derived benchmark~\cite{mp3d} with 1000 fixed room--object queries, 100 hand-annotated crops, target-presence labels, and room-level plausibility labels; and
(iii) an evaluation protocol for measuring safe object-reaching under uncertainty in sampled hypotheses, together with simulated Stretch robot experiments.

\section{Related Work}
\label{sec:background}
\textbf{Planning and search under partial observability.}
Planning under uncertainty is commonly framed with beliefs, POMDPs, or probabilistic objectives~\cite{planning,kurniawati2022partially,silver2010pomcp}, where planners reason over possible worlds rather than a single reconstruction. Since exact belief maintenance is intractable in rich 3D scenes, practical systems use occupancy maps~\cite{elfes1989occupancy,octomap}, semantic maps and scene graphs~\cite{armeni2019scenograph,gu2024conceptgraphs}, object-search priors from spatial relations and commonsense semantics~\cite{vsearch,vsn,chad,toberg2024commonsense}, or active perception policies that reduce uncertainty~\cite{bajcsy1988active,chaplot2020objectgoal,gervet2023navigating}. While these works typically focus on closed-loop policies, they often require an initial prior. We instead study the offline phase before execution, deriving zero-shot priors from pretrained generative vision models.

\textbf{Motion planning under environment uncertainty.}
Sampling-based planners are widely used for configuration-space planning~\cite{kavraki1996prm,lavalle1998rrt,karaman2011sampling}. Under environment uncertainty, paths must be evaluated over possible worlds, motivating uncertain roadmaps, chance constraints, collision-probability constraints, robust planning, and safe navigation with obstacle uncertainty~\cite{pomdp,missiuro2006uncertainprm,blackmore2011chance,patil2012estimating,huang2009collisionprm,quintero2021robust,axelrod2018provably}. We build on sampled-world planning interfaces~\cite{hauser2014mcr,lu2024sampling}; our contribution is a zero-shot generative perception module that supplies sampled worlds, \textit{not} a new planner.

\textbf{Generative world priors.}
Learned world models compress observations into predictive representations for control~\cite{hafner2020dreamer,hansen2022tdmpc,berg2025semanticworldmodels,zheng2023occworld}, while foundation-model and generative-sampling methods estimate beliefs for planning, reasoning, motion generation, and object detection~\cite{wow,zhao2025seeingbelievingbeliefspaceplanning,dimsam,carvalho2024motion,bhattacharjee2025uod}. Scene completion, amodal perception, novel-view synthesis, and generative reconstruction also infer beyond visible evidence~\cite{Song2017SSCNet,unaware,scene_completion,scenesense,mildenhall2021nerf,splatting,gen3d,vmem}. Planning requires samples that are semantically grounded, geometrically explicit, and queryable for collision, visibility, and target evidence. We therefore represent generated 3D spatio-semantic hypotheses as semantically labeled point clouds, providing an embodied planning interface rather than a post-hoc latent simulation.

\section{Problem Formulation}
\label{sec:problem}
\textbf{Setting.} An autonomous robot observes a room from outside a doorway and receives a queried
object category $\object$. From the partial egocentric RGB image, it extracts a
doorway crop $\observ$, which serves as context. Given $\observ$ and $\object$,
our goal is to infer a spatio-semantic prior over the unobserved workspace
$\wspace$ behind the doorway. This prior should describe plausible geometry,
occupancy, and target-object locations so that the robot can choose a
collision-free trajectory that reaches or observes $\object$. Because the object
may be visible, partially visible, fully hidden, or absent, planning must reason
over possible completed worlds rather than the visible crop alone.

We model the unknown room by sampling completed worlds. 
A world $\winst$ is an
element of the measurable scene space $(\cS,\cA)$ and contains the observed crop
evidence together with a hypothesized completion of hidden geometry, free space,
occupied space, and semantic labels. For a fixed observation-query pair
$(\observ,\object)$, let
$Q_{\observ,\object}(A):=\qspace(A\mid\observ,\object)$ for $A\in\cA$, and draw
$\winst^1,\ldots,\winst^N\overset{\mathrm{i.i.d.}}{\sim}Q_{\observ,\object}$.
These samples form the empirical belief used by the planner. Sampling is not
intended to recover a single completion, but to estimate the spatio-semantic
events that determine planning success; details in \cref{app:setup}.

\textbf{Spatio-semantic probability.}
For a world $\winst$, let $E_{\object}(\winst)\subseteq\wspace$ be the 3D
region labeled as the queried object, whether lifted from visible evidence or
generated in the unobserved part of the room.  For a measurable region
$r\subseteq\wspace$, we define the detection probability as
\begin{equation}
\label{eq:pdet_main}
\begin{aligned}
p_{\mathrm{det}}^{\mathrm{reg}}(\observ,\object;r)
&:=
\frac{1}{N}\sum_{i=1}^{N}
\ind{E_{\object}(\winst^i)\cap r\neq\emptyset},\\
p_{\mathrm{det}}(\observ,\object)
&:=
\frac{1}{N}\sum_{i=1}^{N}
\ind{E_{\object}(\winst^i)\neq\emptyset}.
\end{aligned}
\end{equation}
Here $\ind{\phi}$ equals $1$ when condition $\phi$ holds and $0$ otherwise.
In \cref{eq:pdet_main}, the former quantity localizes object evidence to $r$,
whereas the latter is the reported scene-level quantity: it counts whether each
completed world contains the queried object anywhere, retaining the ambiguity
induced by the room label and partial view.

\textbf{Object-reaching probability.}
Let $\cspace$ be the robot configuration space.
Let $F(\winst)\subseteq\cspace$ be the collision-free configuration set, and let $G_{\object}(\winst)\subseteq\cspace$ be the configurations from which the robot can reach or observe $E_{\object}(\winst)$, with $G_{\object}(\winst)=\emptyset$ when $E_{\object}(\winst)=\emptyset$. For a single configuration, \begin{equation} \label{eq:pcoll_main} p_{\mathrm{coll}}^{N}(\state\mid\observ,\object) := \frac{1}{N}\sum_{i=1}^{N}\ind{\state\notin F(\winst^i)}. \end{equation}
For a candidate trajectory $\traj=(\state_0,\ldots,\state_K)$, let
$\mathcal{K}(\traj)\subseteq\cspace$ be the finite set of checked
configurations, including interpolated edge checks.  The empirical planning
probability is
\begin{equation}
\label{eq:pplan_main}
p_{\mathrm{plan}}^{N}(\traj\mid\observ,\object)
:=
\frac{1}{N}\sum_{i=1}^{N}
\ind{\state_K\in G_{\object}(\winst^i)}
\prod_{\state\in\mathcal{K}(\traj)}
\ind{\state\in F(\winst^i)}.
\end{equation}
The product is the indicator of the finite intersection of collision-free
checks within one sampled world.  Hence
$0\le p_{\mathrm{plan}}^{N}(\traj\mid\observ,\object)
\le p_{\mathrm{det}}(\observ,\object)$, and the sampled worlds keep object
semantics, occupancy, and path feasibility coupled.

\textbf{Offline planning objective.} Let $p_{\mathrm{plan}}^{\mathrm{GT}}$ denote the inaccessible true posterior law over completed worlds. Given a partial RGB observation $\observ$ of $\wspace$ and target object $\object$, the ideal object-reaching problem is to find a trajectory
\begin{equation}
\traj^\star \in \operatorname*{argmax}_{\traj}\, p_{\mathrm{plan}}^{\mathrm{GT}}(\traj\mid\observ,\object),
\end{equation}
maximizing the probability of reaching the target collision-free from the robot's initial state. Since $p_{\mathrm{plan}}^{\mathrm{GT}}$ is unavailable, we evaluate the empirical priors above through the spatio-semantic queries and plans they induce.

\section{Dataset}
\label{sec:benchmark}

\begin{figure*}[t]
\centering
\footnotesize

\begin{minipage}{\textwidth}\centering
\begin{subfigure}{0.175\textwidth}
  \centering
  \includegraphics[width=\linewidth]{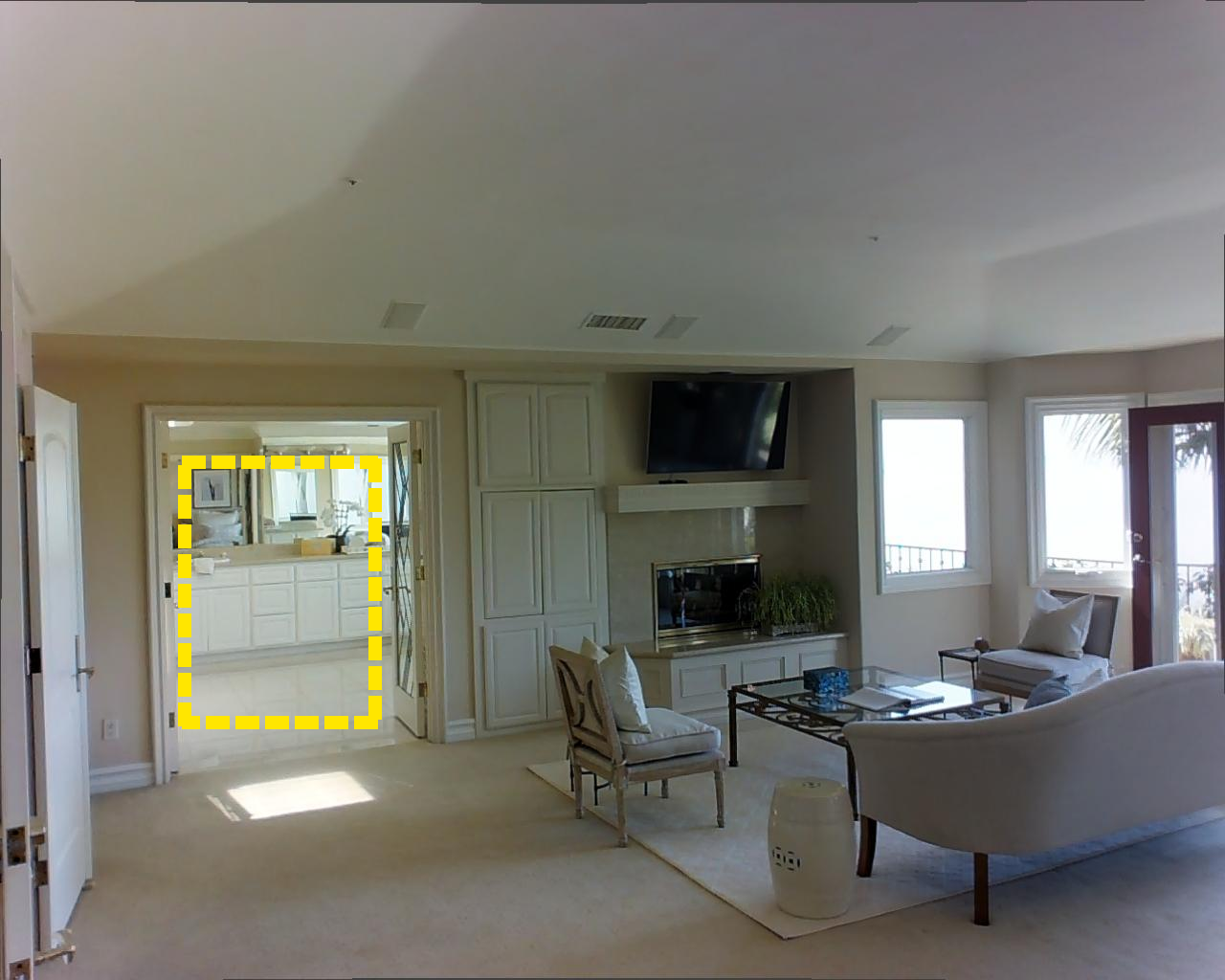}
  \subcaption{Bathroom}\label{subfig:bath1}
\end{subfigure}\hfill
\begin{subfigure}{0.175\textwidth}
  \centering
  \includegraphics[width=\linewidth]{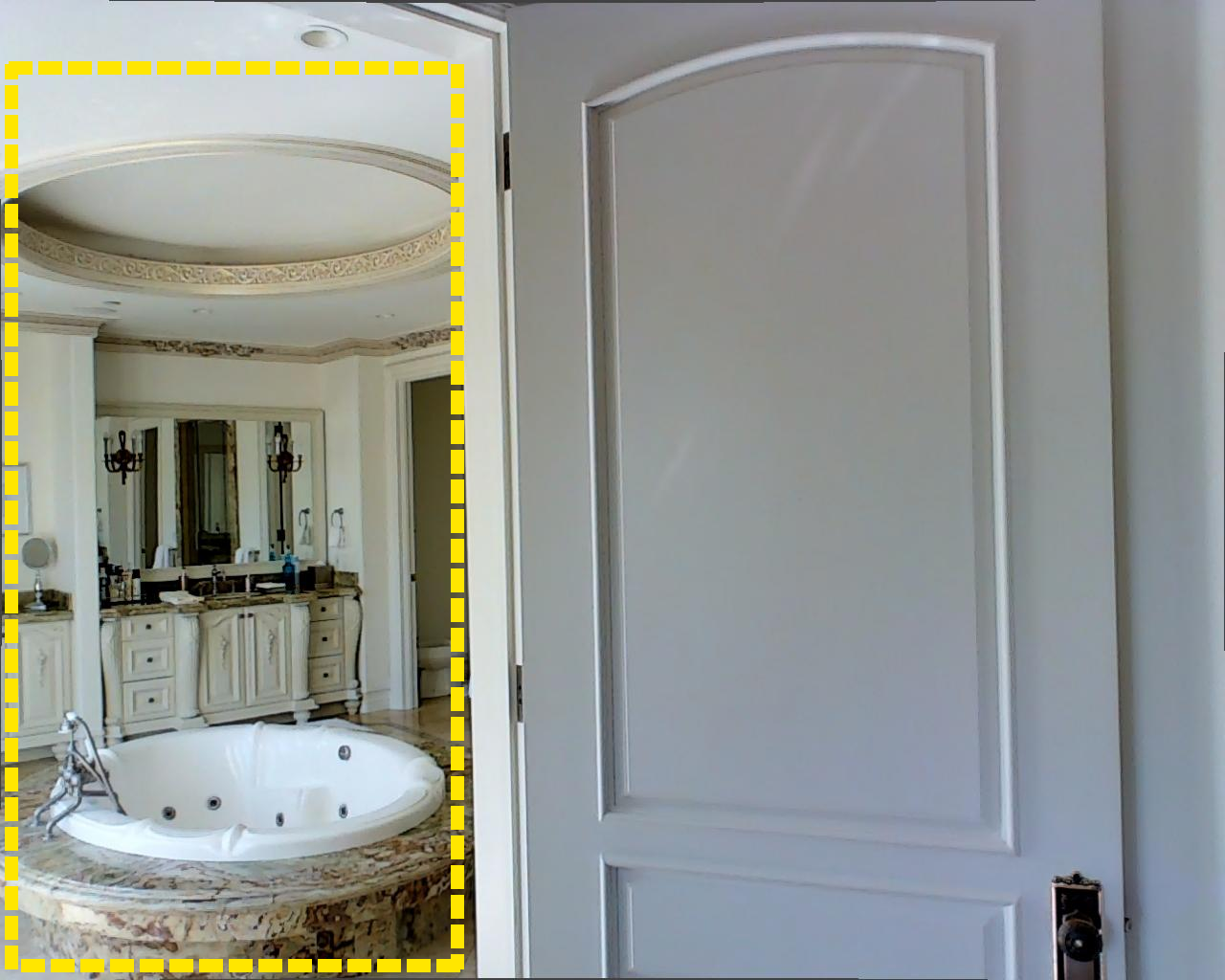}
  \subcaption{Bathroom}\label{subfig:bath2}
\end{subfigure}\hfill
\begin{subfigure}{0.175\textwidth}
  \centering
  \includegraphics[width=\linewidth]{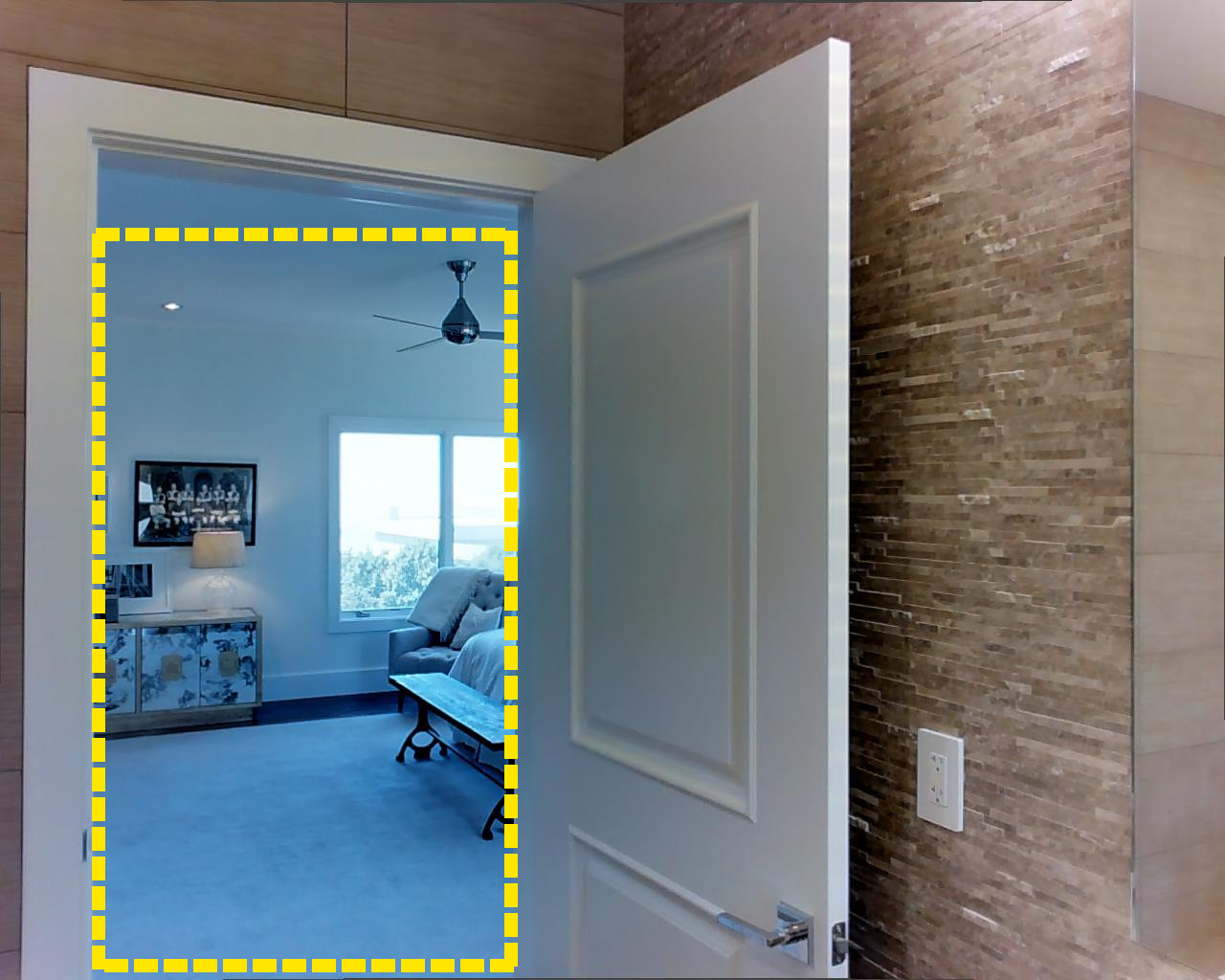}
  \subcaption{Bedroom}\label{subfig:bed1}
\end{subfigure}\hfill
\begin{subfigure}{0.175\textwidth}
  \centering
  \includegraphics[width=\linewidth]{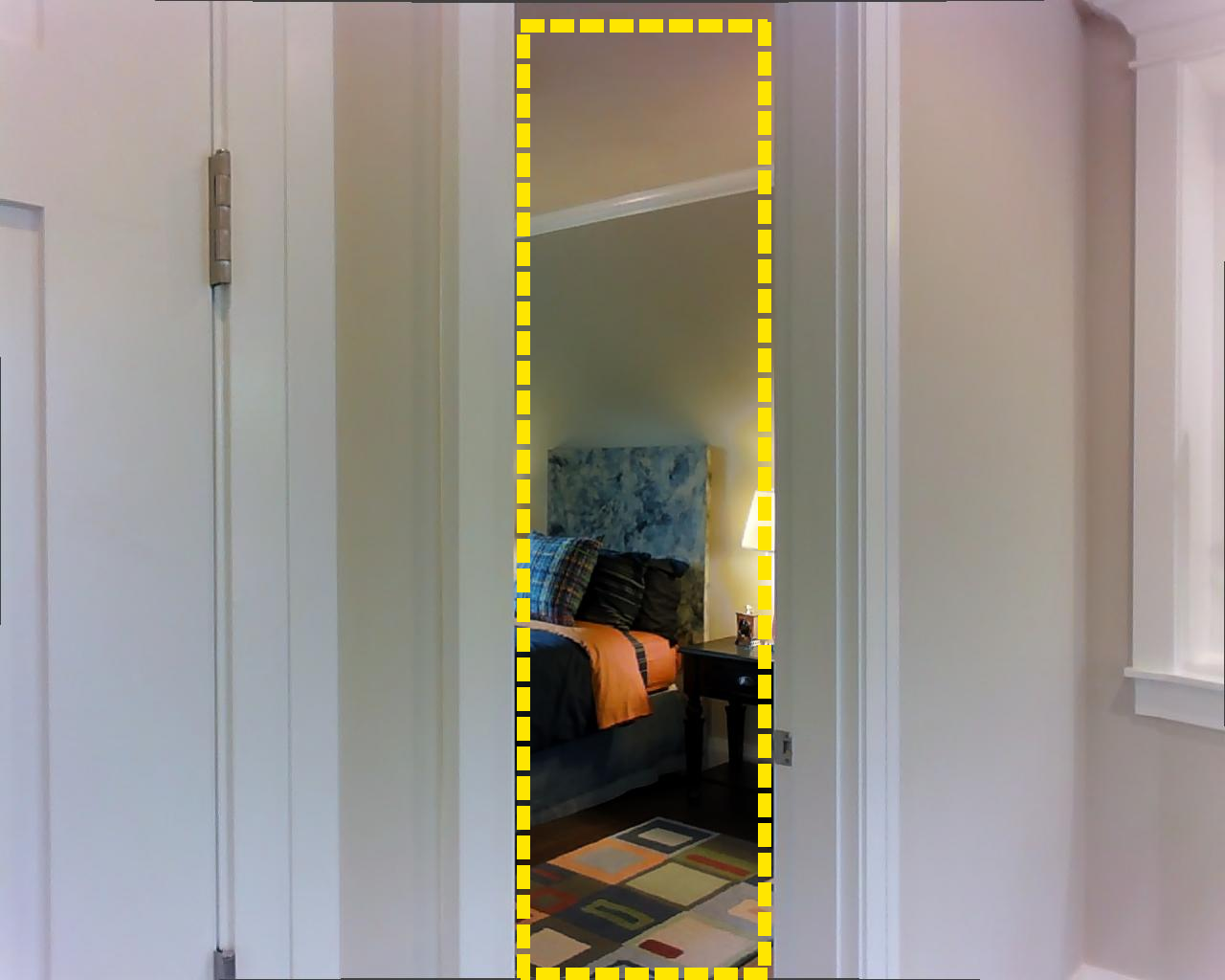}
  \subcaption{Bedroom}\label{subfig:bed2}
\end{subfigure}\hfill
\begin{subfigure}{0.175\textwidth}
  \centering
  \includegraphics[width=\linewidth]{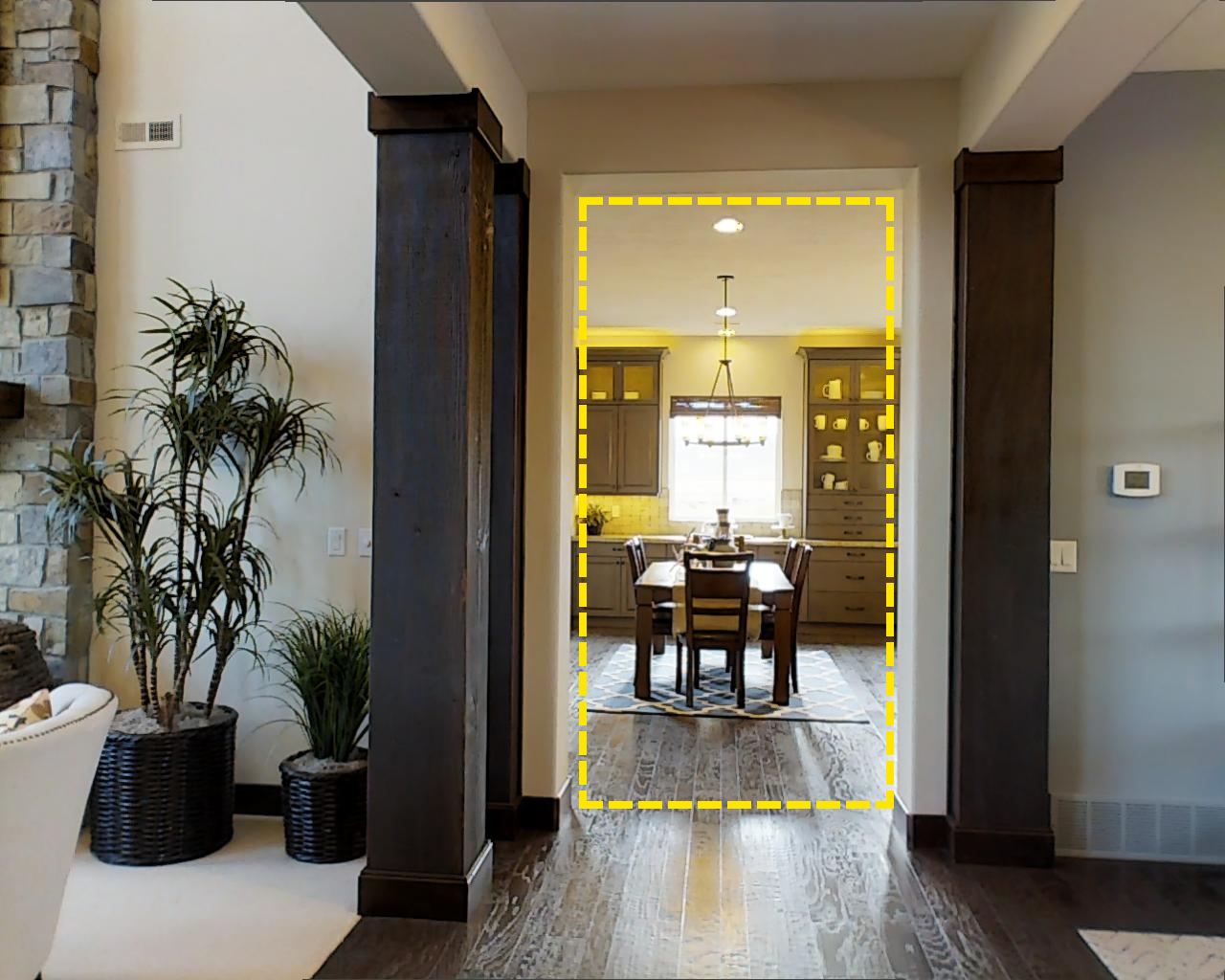}
  \subcaption{Kitchen}\label{subfig:kitchen1}
\end{subfigure}
\end{minipage}

\begin{minipage}{\textwidth}\centering
\begin{subfigure}{0.175\textwidth}
  \centering
  \includegraphics[width=\linewidth]{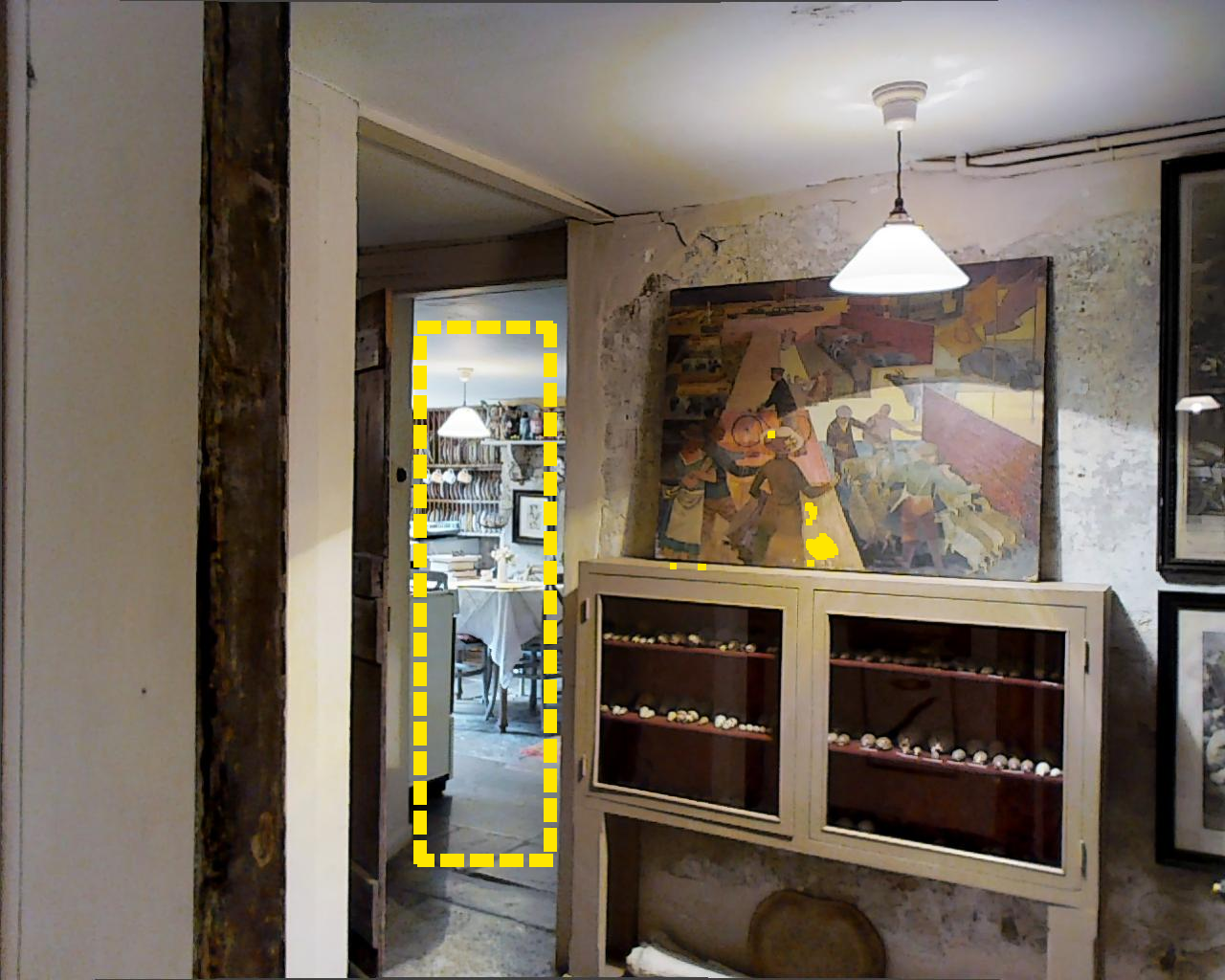}
  \subcaption{Kitchen}\label{subfig:kitchen2}
\end{subfigure}\hfill
\begin{subfigure}{0.175\textwidth}
  \centering
  \includegraphics[width=\linewidth]{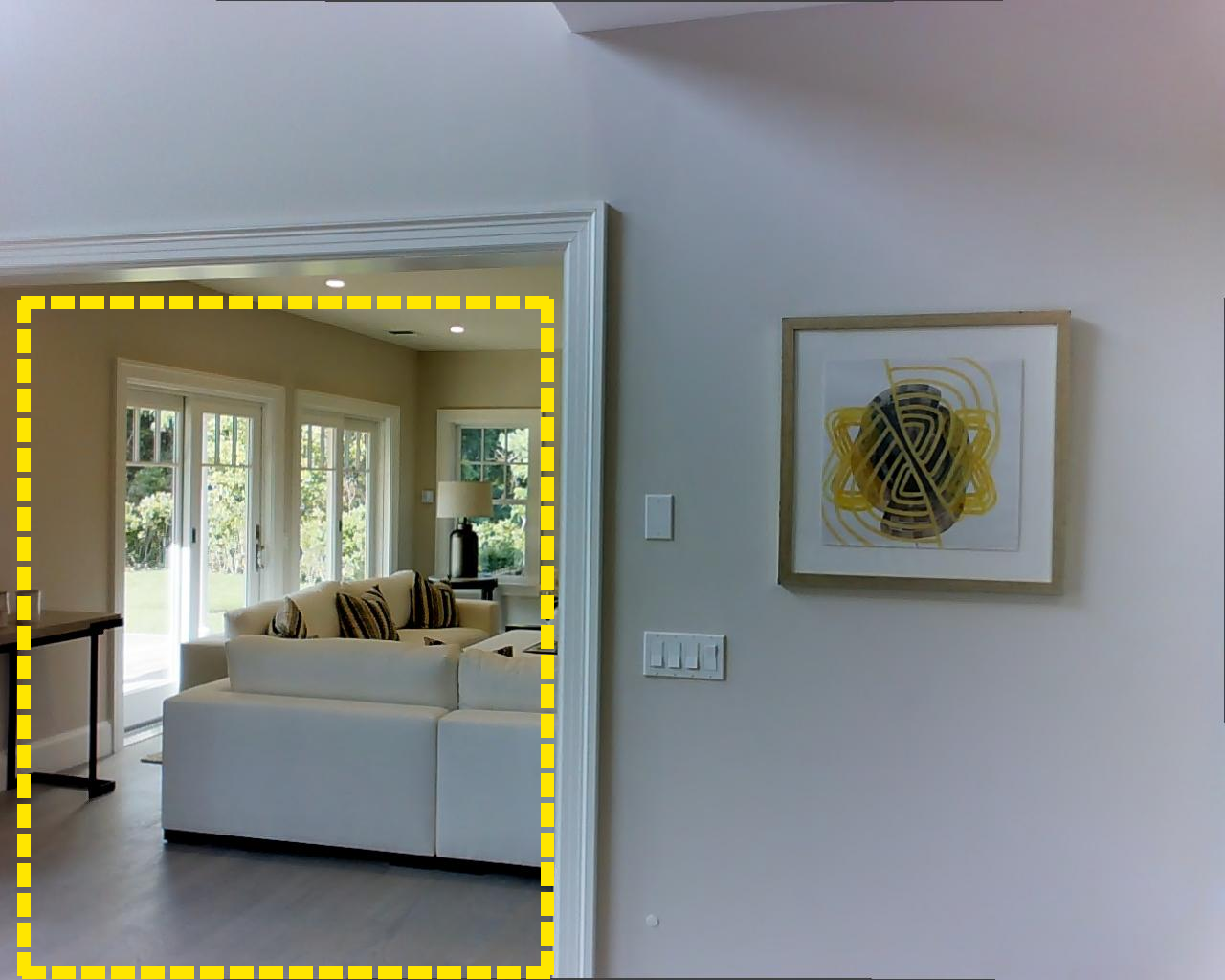}
  \subcaption{Living room}\label{subfig:living1}
\end{subfigure}\hfill
\begin{subfigure}{0.175\textwidth}
  \centering
  \includegraphics[width=\linewidth]{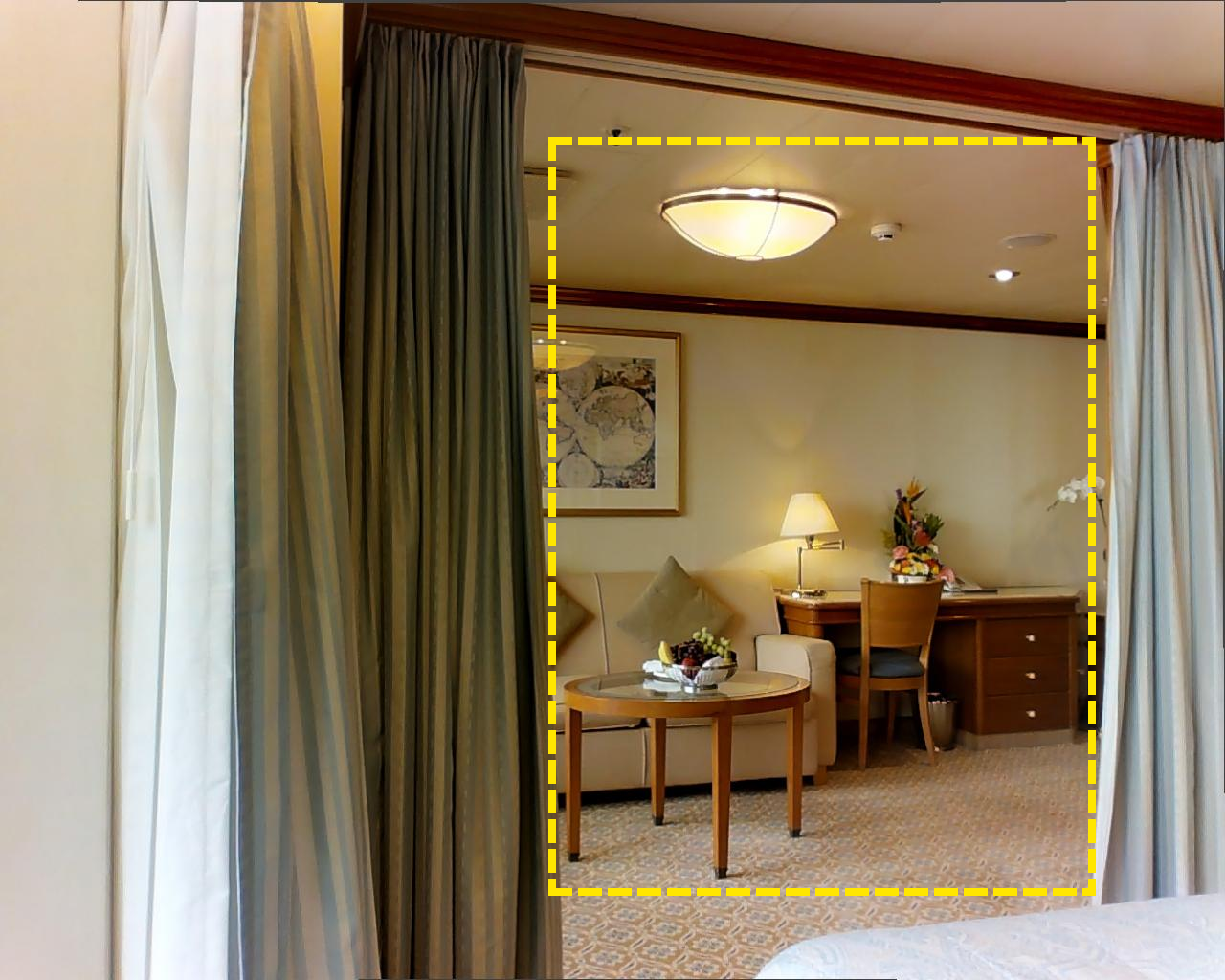}
  \subcaption{Living room}\label{subfig:living2}
\end{subfigure}\hfill
\begin{subfigure}{0.175\textwidth}
  \centering
  \includegraphics[width=\linewidth]{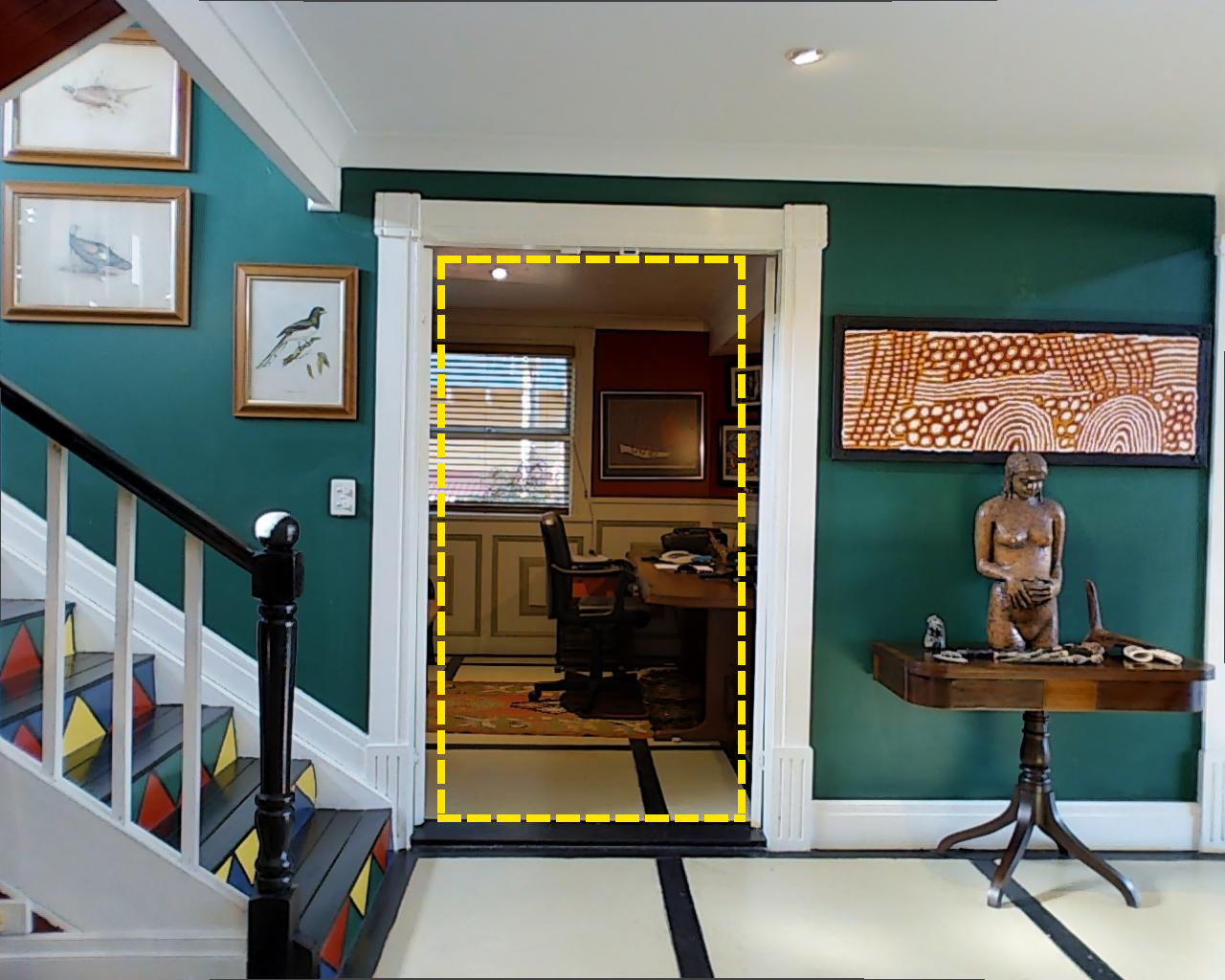}
  \subcaption{Office}\label{subfig:office1}
\end{subfigure}\hfill
\begin{subfigure}{0.175\textwidth}
  \centering
  \includegraphics[width=\linewidth]{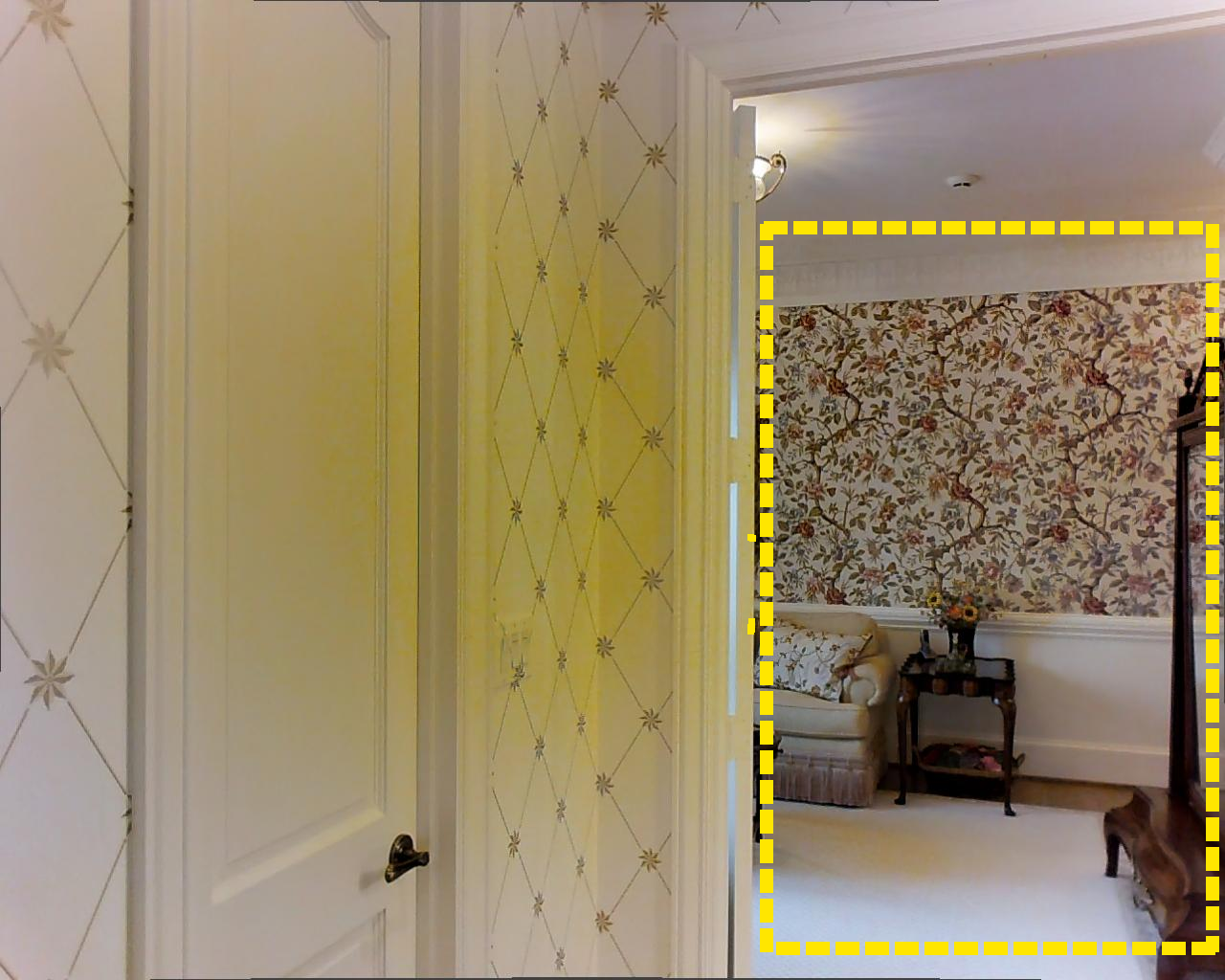}
  \subcaption{Office}\label{subfig:office2}
\end{subfigure}
\end{minipage}
\begin{center}
\resizebox{0.86\textwidth}{!}{\begin{tikzpicture}[font=\sffamily\scriptsize,x=1cm,y=1cm]
  \definecolor{mdInk}{RGB}{35,40,48}
  \definecolor{mdMuted}{RGB}{92,98,108}
  \definecolor{mdBlue}{RGB}{53,126,184}
  \definecolor{mdSky}{RGB}{88,171,227}
  \definecolor{mdGreen}{RGB}{63,160,128}
  \definecolor{mdYellow}{RGB}{229,174,75}
  \definecolor{mdOrange}{RGB}{213,122,73}
  \definecolor{mdRed}{RGB}{191,83,83}
  \definecolor{mdPurple}{RGB}{139,112,177}
  \definecolor{mdPink}{RGB}{197,111,156}
  \definecolor{mdTeal}{RGB}{70,154,151}
  \definecolor{mdGray}{RGB}{151,157,167}
  \tikzset{
    donutedge/.style={draw=white,line width=0.55pt},
    paneltitle/.style={font=\sffamily\bfseries\small,text=mdInk,align=center},
    panelsubtitle/.style={font=\sffamily\tiny,text=mdMuted,align=center},
    legendtext/.style={font=\sffamily\tiny,text=mdInk,anchor=west},
    centerbig/.style={font=\sffamily\bfseries\small,text=mdInk,align=center},
    centersmall/.style={font=\sffamily\tiny,text=mdMuted,align=center}
  }
  \def\donutouter{0.74}
  \def\donutinner{0.44}
  \def\mdslice#1#2#3{%
    \path[fill=#3,donutedge]
      (#1:\donutouter) arc[start angle=#1,end angle=#2,radius=\donutouter]
      -- (#2:\donutinner) arc[start angle=#2,end angle=#1,radius=\donutinner]
      -- cycle;%
  }

  \begin{scope}[shift={(0.00,0)}]
    \node[paneltitle] at (0,1.25) {Target rooms};
    \node[panelsubtitle] at (0,1.01) {100 doorway-side contexts};
    \mdslice{90.000}{-43.200}{mdBlue}
    \mdslice{-43.200}{-133.200}{mdGreen}
    \mdslice{-133.200}{-198.000}{mdYellow}
    \mdslice{-198.000}{-255.600}{mdPurple}
    \mdslice{-255.600}{-270.000}{mdSky}
    \node[centerbig] at (0,0.06) {100};
    \node[centersmall] at (0,-0.15) {contexts};
    \fill[mdBlue] (-1.20,-0.98) rectangle ++(0.10,0.10);
    \node[legendtext] at (-1.05,-0.93) {living 37};
    \fill[mdGreen] (-1.20,-1.20) rectangle ++(0.10,0.10);
    \node[legendtext] at (-1.05,-1.15) {bedroom 25};
    \fill[mdYellow] (-1.20,-1.42) rectangle ++(0.10,0.10);
    \node[legendtext] at (-1.05,-1.37) {kitchen 18};
    \fill[mdPurple] (-1.20,-1.64) rectangle ++(0.10,0.10);
    \node[legendtext] at (-1.05,-1.59) {office 16};
    \fill[mdSky] (-1.20,-1.86) rectangle ++(0.10,0.10);
    \node[legendtext] at (-1.05,-1.81) {bath 4};
  \end{scope}

  \begin{scope}[shift={(3.50,0)}]
    \node[paneltitle] at (0,1.25) {Queried objects};
    \node[panelsubtitle] at (0,1.01) {184 selected GT-present rows};
    \mdslice{90.000}{1.957}{mdBlue}
    \mdslice{1.957}{-78.261}{mdSky}
    \mdslice{-78.261}{-129.130}{mdGreen}
    \mdslice{-129.130}{-160.435}{mdYellow}
    \mdslice{-160.435}{-187.826}{mdOrange}
    \mdslice{-187.826}{-213.261}{mdRed}
    \mdslice{-213.261}{-238.696}{mdPurple}
    \mdslice{-238.696}{-254.348}{mdPink}
    \mdslice{-254.348}{-264.130}{mdTeal}
    \mdslice{-264.130}{-270.000}{mdGray}
    \node[centerbig] at (0,0.06) {184};
    \node[centersmall] at (0,-0.15) {rows};
    \fill[mdBlue] (-1.42,-0.98) rectangle ++(0.10,0.10);
    \node[legendtext] at (-1.27,-0.93) {table 45};
    \fill[mdSky] (-1.42,-1.20) rectangle ++(0.10,0.10);
    \node[legendtext] at (-1.27,-1.15) {chair 41};
    \fill[mdGreen] (-1.42,-1.42) rectangle ++(0.10,0.10);
    \node[legendtext] at (-1.27,-1.37) {cabinet 26};
    \fill[mdYellow] (-1.42,-1.64) rectangle ++(0.10,0.10);
    \node[legendtext] at (-1.27,-1.59) {plant 16};
    \fill[mdOrange] (-1.42,-1.86) rectangle ++(0.10,0.10);
    \node[legendtext] at (-1.27,-1.81) {shelf 14};
    \fill[mdRed] (-0.08,-0.98) rectangle ++(0.10,0.10);
    \node[legendtext] at (0.07,-0.93) {sink 13};
    \fill[mdPurple] (-0.08,-1.20) rectangle ++(0.10,0.10);
    \node[legendtext] at (0.07,-1.15) {TV 13};
    \fill[mdPink] (-0.08,-1.42) rectangle ++(0.10,0.10);
    \node[legendtext] at (0.07,-1.37) {bed 8};
    \fill[mdTeal] (-0.08,-1.64) rectangle ++(0.10,0.10);
    \node[legendtext] at (0.07,-1.59) {micro. 5};
    \fill[mdGray] (-0.08,-1.86) rectangle ++(0.10,0.10);
    \node[legendtext] at (0.07,-1.81) {fridge 3};
  \end{scope}

  \begin{scope}[shift={(7.00,0)}]
    \node[paneltitle] at (0,1.25) {Visibility};
    \node[panelsubtitle] at (0,1.01) {184 selected GT-present rows};
    \mdslice{90.000}{-88.043}{mdOrange}
    \mdslice{-88.043}{-270.000}{mdBlue}
    \node[centerbig] at (0,0.06) {184};
    \node[centersmall] at (0,-0.15) {rows};
    \fill[mdOrange] (-1.26,-0.98) rectangle ++(0.10,0.10);
    \node[legendtext] at (-1.11,-0.93) {partial 91};
    \fill[mdBlue] (-1.26,-1.23) rectangle ++(0.10,0.10);
    \node[legendtext] at (-1.11,-1.18) {hidden 93};
  \end{scope}

\begin{scope}[shift={(10.50,0)}]
    \node[paneltitle] at (0,1.25) {Query policy};
    \node[panelsubtitle] at (0,1.01) {1000 crop--object cells};
    \mdslice{90.000}{23.760}{mdBlue}
    \mdslice{23.760}{-78.120}{mdGreen}
    \mdslice{-78.120}{-81.000}{mdRed}
    \mdslice{-81.000}{-270.000}{mdGray}
    \node[centerbig] at (0,0.06) {1000};
    \node[centersmall] at (0,-0.15) {cells};
    \fill[mdBlue] (-1.38,-0.98) rectangle ++(0.10,0.10);
    \node[legendtext] at (-1.23,-0.93) {GT present 184};
    \fill[mdGreen] (-1.38,-1.22) rectangle ++(0.10,0.10);
    \node[legendtext] at (-1.23,-1.17) {plaus. gen. 283};
    \fill[mdRed] (-1.38,-1.46) rectangle ++(0.10,0.10);
    \node[legendtext] at (-1.23,-1.41) {unlikely gen. 8};
    \fill[mdGray] (-1.38,-1.70) rectangle ++(0.10,0.10);
    \node[legendtext] at (-1.23,-1.65) {skipped 525};
  \end{scope}

\end{tikzpicture}}
\end{center}
\caption{\small \textbf{MatterDoor examples and benchmark diversity.}
Top: Representative MatterDoor doorway observations. The dotted yellow box marks the consensus doorway crop used as the robot observation. 
Bottom: Donut charts summarize room coverage, queried-object coverage, crop visibility split, and query-policy composition across present, plausible absent, unlikely absent, and skipped cases as found after VLM gating (see \cref{app:vlm_study}).}
\label{fig:matterdoor1k_distribution}
\end{figure*}

\textbf{Scene Curation.}
MatterDoor adds a fixed-observation robotics task to Matterport3D~\cite{mp3d}: each query contains a doorway RGB crop and a target object, requiring inference of hidden free space and object evidence beyond the doorway. From undistorted RGB images, we detect indoor doorway views, remove near-duplicates and unreliable RGB-D cases, and retain the \textit{manual crops} by majority vote from three annotators. 
The robot observation is the conservative intersection of accepted doorway boxes. Majority camera rays select the target room\footnote{This is possible because Matterport3D provides full-house semantic reconstructions, global camera poses, and region annotations for most rooms.}, while projected Matterport object 3D bounding boxes provide target visibility in 2D, crop intersection, and semantic labels, yielding $457$ shortlisted scenes from all forward facing (\texttt{i1}) images from all 90 buildings in Matterport.
Since these labels rely on projected geometry and crowd-sourced semantic instances~\cite{mp3d}, automated filtering alone is insufficient: projected targets can leak across room boundaries or outside the intended hidden room. Following the annotation-quality motivation of HM3D-Semantics~\cite{hm3d,hm3dsem}, we verify and \textit{curate $100$ doorway scenes} from $377$ consensus crops on RGB images, camera calibration, room-object labels, mesh geometry, and projected MP40cat target labels.

\textbf{Scene--Object Pairs.}
We pair each verified context with ten object classes shared by MP40cat~\cite{mp3d} and ADE20K~\cite{Zhou2017ADE20K}, giving the fixed $1000$-query MatterDoor dataset (\cref{fig:matterdoor1k_distribution}) motivated by object-search benchmarks~\cite{chad,zeng2018semantic_robot_programming,batra2020objectnav}. For each query, we record GT realization, namely a verified 3D target box and 2D projection in the ray-selected target room, and room-level plausibility, namely non-zero object--room support under Matterport3D co-occurrence and spatial data-mining cues following \citet{hm3dsem}. We remove GT-realized objects whose projected 2D box lies more than $50\%$ inside the crop, since these are mostly visible rather than hidden doorway targets. This separates GT-present queries for realization and 3D localization, plausible GT-absent queries for generated hidden-room completions, and skipped GT-absent queries for gate analysis only. 
MatterDoor spans five common room types (bathroom, bedroom, kitchen, living room, office) and ten object classes (bed, cabinet, chair, microwave, plant, refrigerator, shelf, sink, table, tv). 
The $1000$ crop--object table has $144$ unique GT-present queries; strict 3D
scoring uses $184$ selected GT-present object instances, of which $173$ have
reliable region-based NavMesh support~\cite{navmesh} for planning replay.
Under the selected VLM gate, $291$ GT-absent queries are generated ($283$
room-plausible and $8$ room-unlikely) and $565$ GT-absent queries are skipped.
Dataset provenance, construction, structure, split statistics, taxonomy, geometry checks, reproducibility notes, and plausibility heatmaps are in \cref{app:dataset}. The dataset will be released publicly.

\section{Method}
\label{sec:method}
\begin{figure*}[!t]
    \centering
    \includegraphics[width=0.94\linewidth, trim=60pt 380pt 30pt 380pt, clip]{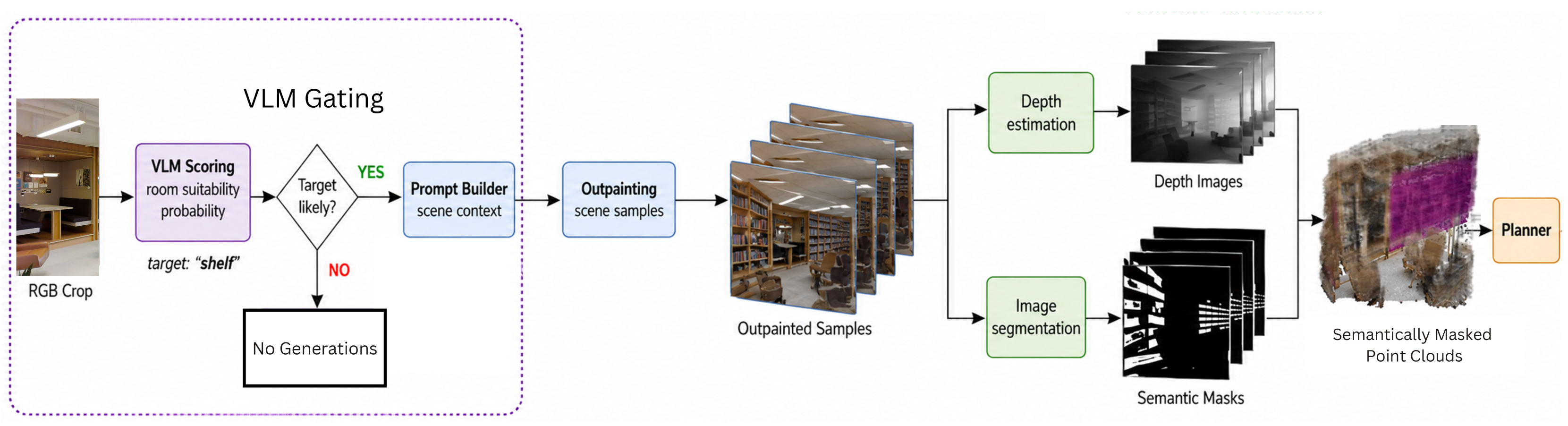}
    \caption{\small \textbf{Pipeline.}
    A MatterDoor query is processed in four stages: a VLM gate scores the queried
    object and constructs a fixed prompt; the outpainting samples hidden RGB
    completions; segmentation and depth estimation lift each completion into an aligned
    spatio-semantic point cloud; and the sampled worlds are passed to the planner
    for collision and target-evidence queries.}
    \label{fig:flowchart}
\end{figure*}

\textbf{Sampling Pipeline.} For each MatterDoor query, the zero-shot pipeline maps a doorway crop, completion side, and queried object to up to $N=100$ 3D world samples (\cref{fig:flowchart}), with no retraining or fine-tuning. Qwen2.5-VL 7B~\cite{qwen25vl} first scores target plausibility behind the doorway from the room crop and object label. We use a fixed gate $\gamma=0.40$ (\cref{app:vlm_study}): accepted queries receive target-conditioned outpainting, while rejected queries normally skip generation. GT-present VLM misses remain in the strict 3D evaluation denominator; if they lack target-cloud samples, a uniform bounding-box fallback is used. Accepted crops are outpainted by $500$ pixels on each side using a 4-bit quantized FLUX.1-Fill-dev model with a compiled transformer kernel and layer pre-caching~\cite{BFL2024Fill,nunchaku,cache}, while preserving observed doorway evidence through the $10$--$50\%$ estimated-depth band. For each query, $N=100$ RGB completions are segmented with ADE20K SegFormer-B5~\cite{Zhou2017ADE20K,segformer}, mapped to the ten MatterDoor classes, filtered at $0.25$ confidence and $0.5\%$ image area, lifted with DepthPro metric depth~\cite{Bochkovskii2024}, and affinely aligned to the estimated crop depth using RANSAC~\cite{Fischler1981RANSAC}, to prevent scale ambiguity. 
We disable DepthPro's focal-length prediction head and use known MatterDoor intrinsics for ray-preserving back-projection; intermediate resizing~\cite{resize} retains detail at lower working resolution. Pipeline implementation details are in \cref{app:pipeline_stages}, with the gate, outpainting, segmentation, lifting, and resource/sample-budget convention detailed in \cref{app:stage0_vlm,app:stage1,app:stage2,app:stage3,app:throughput_budget}. Excluding amortized pre-processing, sampling takes $1\pm0.25~\mathrm{s}$ per sample. Pipeline code will be released for reproducibility.

\textbf{Planner.}
The planner treats each lifted (statistical outlier removed) point cloud as a sampled world $\winst^i$ and
uses the notation of \cref{eq:pcoll_main,eq:pplan_main} for collision,
free-space, and target-evidence checks. 
At evaluation, we use the region mesh to define a bounded, floor-aligned convex search volume.
We instantiate a Stretch mobile robot~\cite{stretch}, a common
indoor mobile-manipulation platform supported by embodied
simulators~\cite{omnigibson}, and restrict planning to
$\cspace_F=\{(t,R)\in\mathrm{SE}(2):t\in F\}\subseteq\cspace$, where
$F\subset\mathbb{R}^2$ is the union of sampled floor regions clipped to the
convex floor-plane footprint of the search volume.  Goal reachability is
evaluated by $G_{\object}(\winst^i)\subseteq\cspace_F$.  We build a roadmap in
$\cspace_F$ using the sampling-based planner of~\citet{lu2024sampling}
\cite{hauser2014mcr}, with FCL~\cite{fcl} checking feasibility in each sampled
world.  The selected trajectory maximizes
$p_{\mathrm{plan}}^{N}(\traj\mid\observ,\object)$ in \cref{eq:pplan_main} and
is evaluated post hoc on the Matterport3D ground-truth reconstruction.  Planning
takes $1\pm0.5~\mathrm{min}$ for the $100$ samples used per task; details are
in~\cref{app:eval_plan}.

\section{Experimental Setup}
\label{subsec:setup}

\textbf{VLM gating.}
MatterDoor evaluates $1000$ doorway--object queries.  The gate-accounting table
contains $144$ unique GT-present queries; strict 3D metrics use $184$ selected
GT-present object instances.  At $\gamma=0.40$, $37$ GT-present queries enter
the conservative fallback path, and $291$ GT-absent queries are generated for
semantic plausibility analysis.  Fallback keeps real-target VLM misses in the
denominator without giving the generator or planner target-location information.

\textbf{Generative Metrics.}
\label{subsec:metrics}
For task $t$, let $\mathcal S_t=\{1,\ldots,N_t\}$ be the generated samples and
$\mathcal D_t=\{i\in\mathcal S_t:|\hat Y_t^i|>0\}$ the detections.
We compare smoothed empirical room--object distributions using
$p^{\rm GT}_{r,\alpha}(o)=(n^{\rm GT}_{r,o}+\alpha)/(\sum_{o_i}n^{\rm GT}_{r,o_i}+\alpha|\mathcal O|)$
and
$p^{\rm VLM}_{r,\alpha}(o)=(s_{r,o}+\alpha)/(\sum_{o_i}s_{r,o_i}+\alpha|\mathcal O|)$,
where $\alpha=1$~\cite{smooth} and $s_{r,o}$ is the crop-conditioned VLM score
(\cref{app:vlm_study}). We report
\begin{equation}
\label{eq:semantic_label_kl_main}
D_{\rm sem}=
\frac{1}{|\mathcal R|}
\sum_{r\in\mathcal R}
\KL\!\left(p^{\rm GT}_{r,\alpha}\middle\|p^{\rm VLM}_{r,\alpha}\right),
\end{equation}
where lower values indicate closer agreement with Matterport3D room--object statistics; a uniform object prior is included as reference.
We measure object realization by
$p_{\rm det}(t)=|\mathcal D_t|/|\mathcal S_t|$ and report its benchmark average.
For localization, let $A_t$ and $\hat A_t^i$ be the GT and generated boxes,
$d_{ti}=\operatorname{dist}(\hat A_t^i,A_t)$, and assign $d_{ti}=\infty$ for misses.
For the first $K$ samples $\mathcal S_t^K$ and threshold $\tau$,
\begin{equation}
\label{eq:bbox_opr_main}
\operatorname{mOPR}_{K,\tau}
=
\frac{1}{|\mathcal T|K}
\sum_{t\in\mathcal T}\sum_{i\in\mathcal S_t^K}\ind{d_{ti}\le\tau},
\qquad
\operatorname{bOPR}_{K,\tau}
=
\frac{1}{|\mathcal T|}
\sum_{t\in\mathcal T}\ind{\min_{i\in\mathcal S_t^K}d_{ti}\le\tau}.
\end{equation}
mOPR estimates single-sample target recall, while bOPR asks whether any sampled hypothesis reaches the target region; they coincide for deterministic one-sample baselines.
Geometry is evaluated by symmetric Chamfer distance $\Chamfer$~\cite{chamfer}
between generated and GT point clouds inside the expanded frustum. Target localization uncertainty is summarized by the detected-sample energy score~\cite{gneiting2007strictly,szekely2013energy}:
\begin{equation}
\label{eq:placement_energy_main}
\mathcal E(t)=
\frac{1}{|\mathcal D_t|}
\sum_{i\in\mathcal D_t}\|c_t^i-c_t^{\rm GT}\|_2
-
\frac{1}{2|\mathcal D_t|^2}
\sum_{i,j\in\mathcal D_t}\|c_t^i-c_t^j\|_2 .
\end{equation}
The first term rewards accurate placement and the second accounts for diversity; for single hypotheses it reduces to center error. Lower Chamfer and energy are better, and the GT oracle scores $0$.

\textbf{Planning Metrics.}
Planning metrics test whether the sampled-world belief is useful for trajectory selection, beyond merely capturing hidden-room semantic evidence, as in belief-space planning under uncertainty~\cite{pomdp,zhao2025seeingbelievingbeliefspaceplanning,axelrod2018provably,vandenberg2011lqgmp}. Given an observation $\observ$, our planner optimizes a trajectory $\pi$ using the empirical joint prior over hidden occupancy and target location induced by sampled scenes (\cref{eq:pplan_main}). We evaluate the 173 tasks with a verified valid region-based NavMesh by replaying each planned trajectory in the corresponding ground-truth scene. We report $p_{\mathrm{target}}^{\mathrm{GT}}$, the fraction of terminal configurations that reach the ground-truth target; $p_{\mathrm{coll}}^{\mathrm{GT}}$, the fraction of trajectories that collide with the ground-truth environment; $d_{\mathrm{tgt}}$, the terminal distance to the ground-truth target; and $d_{\mathrm{traj}}$, the Fr\'echet distance~\cite{frechet} to the ground-truth-oracle trajectory. We also report $\chi_{\mathrm{GT}}\in[0,1]$, the trajectory-length-weighted collision-free fraction, and $C_f\in[0,1]$, the trajectory-length-weighted collision-free prefix fraction before the first ground-truth collision. Higher is better for $p_{\mathrm{target}}^{\mathrm{GT}}$, $\chi_{\mathrm{GT}}$, and $C_f$; lower is better for $p_{\mathrm{coll}}^{\mathrm{GT}}$, $d_{\mathrm{tgt}}$, and $d_{\mathrm{traj}}$.
\label{subsec:baselines}

\textbf{Generative Baselines.}
We separate discrete-label and 3D baselines because their outputs differ.  The
\emph{uniform object prior} assigns equal probability to every benchmark object
in every room and is evaluated only by $D_{\mathrm{sem}}$.  \emph{No generation}
emits no object or geometry, fixing the zero point for $p_{\mathrm{det}}$ and
OPR.  The \emph{uniform random search-volume} baseline samples target locations
uniformly in the valid hidden-room volume without RGB or object cues, reported
at $K=100$ and averaged over seeds.  Two deterministic one-hypothesis controls
place a target-sized box at the search-volume centroid or doorway-side
frontier~\cite{yamauchi1997frontier};
repeated copies are not counted as independent samples.  The \emph{GT oracle} is
a non-deployable ceiling using Matterport3D target boxes and scene geometry.
Degenerate sanity checks are omitted from the main table and reported in
\cref{app:additional_result_breakdowns}; direct-generation pilot alternatives
are discussed separately in \cref{app:other_pipe}.

\textbf{Planning Baselines.}
We compare four planning settings with the same robot, task, trajectory parameterization, and ground-truth replay evaluation, differing only in the planning prior. \emph{Uniform} plans from the observed scene and one target location sampled uniformly in the hidden search region, providing no learned occupancy or placement prior. \emph{1-sample} uses the observed scene plus the first generated completion and target hypothesis, collapsing the belief to one deterministic world. \emph{Ours} uses all 100 generated environment--target hypotheses to estimate joint task-success probability and select trajectories robust to hidden free-space and target-location uncertainty. \emph{GT oracle} is a non-deployable upper bound using ground-truth geometry and target location. 
All trajectories are evaluated against the same ground-truth scene and target with the metrics above.

\label{sec:results}
\section{Results \& Discussion}
\label{subsec:results}

\begin{table}[t]
\centering
\scriptsize
\setlength{\tabcolsep}{3.0pt}
\renewcommand{\arraystretch}{0.92}
\caption{\textbf{Generative priors evaluation.}
$D_s{=}D_{\rm sem}$, $p_d{=}p_{\rm det}$, m/bOPR are mean/best OPR@1m,
CD is symmetric Chamfer, and $\mathcal{E}$ is placement energy in meters.
Dashes denote inapplicability.}
\label{tab:gen_overall}
\begin{tabularx}{\linewidth}{
@{}
>{\arraybackslash}p{1.55cm}
*{6}{>{\centering\arraybackslash}X}
@{}
}
\toprule
Method
& $D_s{\downarrow}$
& $p_d{\uparrow}$
& mOPR${\uparrow}$
& bOPR${\uparrow}$
& CD${\downarrow}$
& $\mathcal{E}{\downarrow}$ \\
\midrule
Unif. obj. prior & 0.642 & --    & --    & --    & --   & --    \\
No generation    & --    & 0.000 & 0.000 & 0.000 & --   & --    \\
Unif. random     & --    & --    & 0.001 & 0.039 & --   & 3.42  \\
Centroid         & --    & --    & 0.006 & 0.006 & --   & 5.04  \\
Frontier         & --    & --    & 0.003 & 0.003 & --   & 10.07 \\
\textbf{Ours}    & \textbf{0.144} & \textbf{0.903} & \textbf{0.901}
                 & \textbf{0.925} & \textbf{1.81} & \textbf{2.26} \\
GT oracle        & 0.000 & 1.000 & 1.000 & 1.000 & 0.00 & 0.00 \\
\bottomrule
\end{tabularx}
\end{table}

\begin{table*}[t]
\centering
\scriptsize
\setlength{\tabcolsep}{3.0pt}
\renewcommand{\arraystretch}{0.92}
\caption{\textbf{Planning evaluation.}
Metrics are computed over 173 cases.
Here $d_{\mathrm{tgt}}=\|\pi(1),o_{\rm GT}\|$ and
$d_{\mathrm{traj}}=\|\pi,\pi_{\rm GT}\|$.
Bold indicates the best deployable method.}
\label{tab:planning_main}
\begin{tabularx}{\textwidth}{
@{}
>{\arraybackslash}p{1.25cm}
*{6}{>{\centering\arraybackslash}X}
@{}
}
\toprule
Method
& $p^{\rm GT}_{\rm tgt}{\uparrow}$
& $p^{\rm GT}_{\rm coll}{\downarrow}$
& $d_{\rm tgt}{\downarrow}$
& $d_{\rm traj}{\downarrow}$
& $\gtprior{\uparrow}$
& $C_f{\uparrow}$ \\
\midrule
Uniform
& 0.069 & 0.815 & 5.050 & 6.175 & 0.316 & 0.291 \\
1-sample
& 0.041 & 0.971 & 6.034 & 6.695 & 0.049 & 0.043 \\
\textbf{Ours}
& \textbf{0.231} & \textbf{0.428} & \textbf{2.693}
& \textbf{5.928} & \textbf{0.884} & \textbf{0.849} \\
GT oracle
& 1.000 & 0.075 & 1.098 & 0.000 & 0.925 & 0.925 \\
\bottomrule
\end{tabularx}
\end{table*}

\textbf{Generation Results.} \Cref{tab:gen_overall} shows that MatterDoor turns a single doorway crop into a
useful sampled 3D belief; qualitative results are provided in \cref{app:qualitative_results}. It realizes the queried target in $90.3\%$ of samples,
achieves $90.1\%$ mean OPR@1m for a single draw, and reaches $92.5\%$ best
OPR@1m over the sample budget. This is not explained by search-volume size:
uniform random reaches only $0.1\%$ mean OPR@1m and $3.9\%$ best OPR@1m, with
higher placement energy ($3.42$\,m vs. $2.26$\,m). MatterDoor also improves the
room--object semantic score over the uniform object prior ($0.144$ vs. $0.642$),
while the GT oracle marks the non-deployable ceiling. Undefined baseline metrics
are left blank rather than counted as zeros. The scene and object breakdowns in
\cref{tab:gen_breakdown} use the same metrics: scene rows include
$D_{\mathrm{sem}}$ because semantic KL is room-distribution based, while object
rows omit it because object-level KL is not defined. \Cref{tab:main_task_averages}
separates partially visible and out-of-crop targets; the out-of-crop split
remains strong in $p_{\mathrm{det}}$ and OPR@1m, but has worse $\Chamfer$
because less room geometry is constrained by the semantics of the input image.

\textbf{Ablations.}
The main results in~\cref{tab:gen_overall} compare against no-generation, uniform-random, centroid, frontier, and GT-oracle controls.
These isolate whether
success comes from target belief rather than box volume or fixed geometric
heuristics: uniform random reaches only $3.9\%$ best OPR@1m after $100$ draws, whereas MatterDoor reaches $90.1\%$ mean OPR@1m from a single draw and has lower placement energy. Appendix ablations then vary sample budget $K$, OPR tolerance,
scene, object, task type, and VLM gate threshold. The pattern is stable: additional
samples mainly help hard out-of-crop targets; visible targets often succeed
early; and the semantic gate trades coverage against skipped absent queries
without changing the placement-metric interpretation. Full results are in
\cref{app:generative_ablations,app:vlm_study}; ablation curves and
heatmaps are in
\cref{fig:app_opr_budget_thresholds,fig:app_opr_budget_baselines,fig:app_distance_budget_curve,fig:app_semantic_kl_by_room,fig:app_opr_budget_by_scene,fig:app_opr_budget_by_object,fig:app_opr_budget_by_task},
the gate-threshold sweep is in \cref{fig:gamma_threshold_study}, and numerical
details are in
\cref{tab:app_full_ablation_details,tab:app_bbox_budget_current,tab:app_bbox_scene_current,tab:app_bbox_object_current,tab:app_bbox_task_current,tab:gamma_threshold_breakpoints}.

\textbf{Planning Results.}
\label{subsec:planning}
Our sampled-belief planner consistently outperforms the deterministic single-world controls under ground-truth replay. Compared with \emph{Uniform}, it improves target-reaching success from 0.069 to 0.231, reduces collision rate from 0.815 to 0.428, and increases the weighted collision-free and prefix fractions from 0.316/0.291 to 0.884/0.849. It also reduces terminal target distance (2.693 versus 5.050) and produces trajectories closer to those of the GT oracle.
In contrast, \emph{1-sample} performs worse than \emph{Uniform}, showing that a single inaccurate completion can mislead planning. 
The GT oracle can still collide because target reachability does not guarantee collision-free reachability through the revealed room geometry.
Overall, planning over multiple plausible completions yields safer, more goal-directed trajectories than deterministic planning from either an uninformative prior or a single completion.

\section{Conclusion}
\label{sec:conclusion}
We introduced a generative sampling framework that converts partial RGB
observations into spatio-semantic priors for planning under uncertainty.
Instead of relying only on curated domain-knowledge priors~\cite{assist}, the
method treats pretrained vision models as conditional environment samplers whose
outputs are filtered, sampled, and lifted into labeled 3D point-cloud worlds.
MatterDoor provides curated scenes with semantic annotations and partial
observations for evaluating these hidden-room priors.  The benchmark tests
whether generated 3D hypotheses are plausible, target-aware, geometrically
grounded, and usable by a downstream planner.  Overall, the work shows that
pretrained vision models can provide planner-queryable beliefs for robust
sampling-based motion planning beyond the field of view.

\textbf{Future work.} Next steps include expanding this benchmark to more
datasets, including outdoor scenes~\cite{dl3dv10k}, to test whether
doorway-hidden priors transfer across broader architecture and sensing
conditions.  Connecting sampled-world planning to interactive world models that
revise beliefs from new observations~\cite{wow,bajcsy1988active,bohg2017interactive,huang2023vlmaps,shafiullah2023clipfields}
would let robots maintain semantic maps and decide when to gather evidence
rather than act only on priors.  Future work can also explore receding-horizon
planning or control~\cite{mayne2000constrained,janner2022diffuser,carvalho2024motion,chi2023diffusionpolicy}
and extend object queries to affordance queries~\cite{gibson1979ecological,omnigibson,ichter2023saycan,huang2023voxposer}.

\section{Limitations}
\begin{figure}[t]
    \centering
    \includegraphics[width=0.98\linewidth]{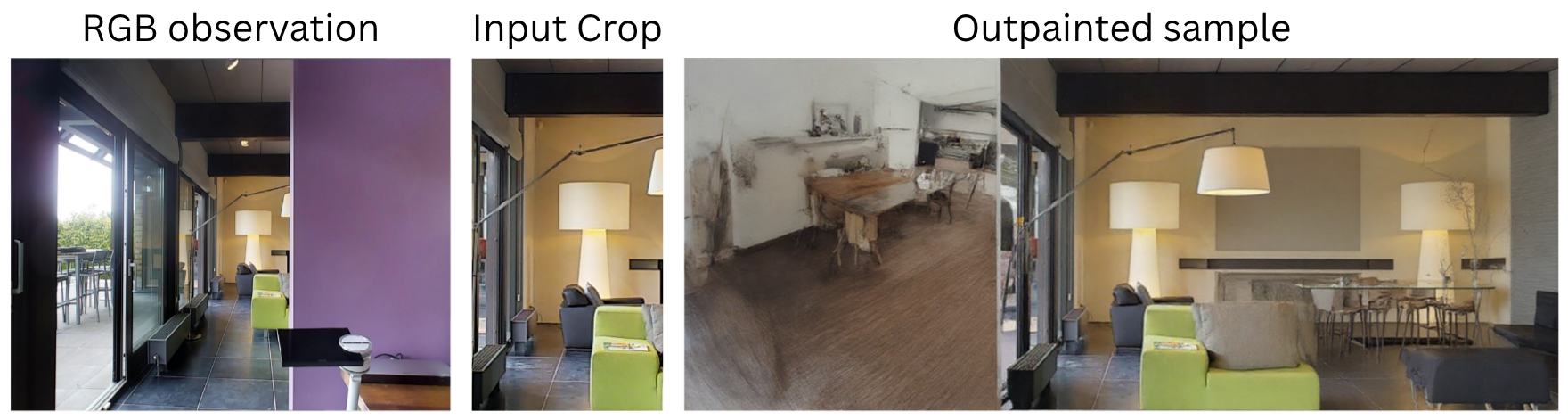}
    \caption{\textbf{Outpainting failure}: Generated sample loses context of the left expansion in the generation.}
    \label{fig:failure_outpaint}
\end{figure}

Our benchmark uses a single source dataset, limiting diversity in architecture, geography, furnishing style, and sensing conditions. We study fixed-observation priors rather than closed-loop robot policies, and the pipeline inherits biases from pretrained components; for example, outpainting can violate room layout~\cite{bhattacharjee2025uod}, as shown in  \cref{fig:failure_outpaint} (although in $<0.2\%$ of samples). 
We omit a matched direct RGB-to-3D baseline~\cite{tewari2023dfm,vmem} because current open-weight models have incompatible conditioning, high runtime or memory cost, and weak hidden-target localization in our pilot study; \cref{app:other_pipe,tab:vmem} reports these pilot alternatives rather than scored baselines.
Challenges beyond benchmarking remain, including sampling time, parallel planning implementation, and closed-loop real-world robot evaluations.

\clearpage
\appendix
\begin{center}
  \LARGE\bfseries Appendix
\end{center}

\setcounter{section}{0}
\renewcommand{\thesection}{\Alph{section}}
\setcounter{table}{0}
\renewcommand{\thetable}{S\arabic{table}}
\setcounter{figure}{0}
\renewcommand{\thefigure}{S\arabic{figure}}
\setcounter{equation}{0}
\renewcommand{\theequation}{S\arabic{equation}}
\renewcommand{\theHequation}{S\arabic{equation}}
\begingroup
\vbadness=10000

\section{Qualitative Results}
\label{app:qualitative_results}

\Cref{fig:qualitative-appendix} shows representative outputs.  Each example
includes the GT crop, one generated RGB hypothesis, the aggregated
target-support heatmap, and the GT 3D context.  The heatmap voxelizes generated
target-object points from all samples, max-pools them over height, and uses a
shared inferno scale.  The magenta dotted outline and star mark the GT footprint
and centroid, while the cyan outline marks the 1 m tolerance region.  Visual
agreement is strongest when high heatmap support overlaps the GT footprint or
the tolerance region.

\begin{figure*}[!th]
\centering
\includegraphics[
  width=\linewidth,
  trim=70pt 10pt 0 0,
  clip
]{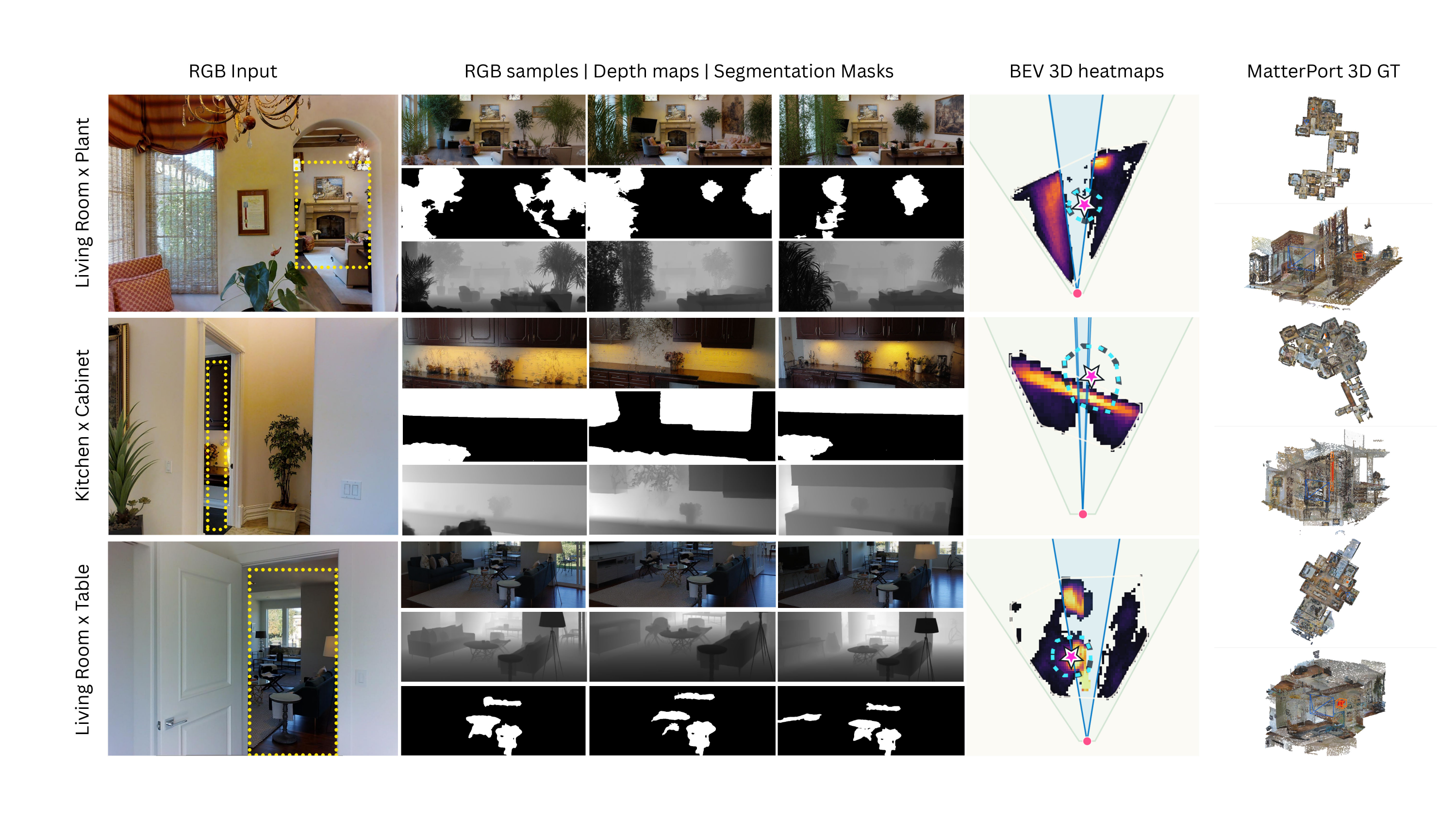}
\caption{Compact qualitative benchmark outputs.  Each example shows the GT RGB crop guide, one cropped generated RGB hypothesis, the aggregated target-support heatmap, and the GT 3D context.
In the heatmaps, inferno intensity denotes generated target-footprint support from the generated object points; the pale outline is the generated support hull; green and blue indicate the search volume and crop/frustum footprint; the magenta dotted outline and star mark the GT footprint and centroid; and the cyan dotted outline marks the 1 meter tolerance region.}
\label{fig:qualitative-appendix}
\end{figure*}

\section{Mathematical Formulation and Consistency of Empirical Priors}
\label{app:setup}

We fix the probability notation for the sampled-world prior used
in~\cref{sec:problem}.  For a fixed doorway observation and object query, the
generative pipeline defines a conditional law over completed worlds.  Collision,
target, and path-success quantities are events under that law, and their
empirical priors are sample averages of the corresponding indicators.
We use standard facts about standard Borel spaces, Markov kernels, finite
product measures for i.i.d.~samples, and regular conditional
probabilities~\cite{klenke2020,kechris1995descriptive,bogachev2007measure}.
For a Polish space $E$, $\cB(E)$ denotes its Borel $\sigma$-algebra and
$\cP(E)$ denotes the set of Borel probability measures on $E$.

\textbf{Conditioning, world space, and samples.} The conditioning input is the
finite-resolution data given to the generative pipeline.  The main text writes
this input as a doorway crop and object query, $(\observ,\object)$.  Here we
fold the pair into one symbol, still denoted by $\observ$:
\begin{equation}
\label{eq:obs_space}
E_{\obs}
:=
\R^{H\times W\times 3}\times\mathcal{O}.
\end{equation}
The set $\mathcal{O}$ is finite and has the discrete topology.  If the object
label is not part of a particular conditioning input, the same notation applies
with $\mathcal{O}$ a singleton.  With this convention, the law $Q_{\observ}$
below is the same object as $Q_{\observ,\object}$ in~\cref{sec:problem}.
A completed world is represented by a padded, semantically labeled point cloud.
For a fixed maximum size $M$ and finite semantic label set
$\cC=\{0,1,\ldots,K_{\mathrm{cat}}\}$, define
\begin{equation}
\label{eq:scene_space}
\cS:=\R^{M\times3}\times\cC^M,\qquad \cA:=\cB(\cS).
\end{equation}
Each point carries coordinates and a semantic label, $(x,y,z,\ell)$ with
$\ell\in\cC$; smaller clouds are padded with a fixed null label.  Since
$E_{\obs}$ and $\cS$ are finite-dimensional Borel spaces with a finite discrete
factor, both are standard Borel spaces.

\textbf{Model and true posterior laws.} The generative pipeline is represented
by a Markov kernel from observations to completed worlds:
\begin{equation}
\label{eq:model_kernel}
\qspace:E_{\obs}\times\cA\to[0,1].
\end{equation}
We write $\qspace(A\mid\observ)$ for $\qspace(\observ,A)$.
For a fixed observation value $\observ$ and $A\in\cA$, the model-induced law
over completed worlds is
\begin{equation}
\label{eq:model_law}
Q_{\observ}(A)
:=
\qspace(A\mid\observ).
\end{equation}
Thus
$\winst^1,\ldots,\winst^N\overset{\mathrm{i.i.d.}}{\sim}Q_{\observ}$ means
that every sampled world is generated independently from the same conditional
law.  The joint law of the $N$ samples is the product measure
$Q_{\observ}^{\otimes N}$.

Let $(\Omega,\mathcal{F},\mathbb{P})$ be the probability space for the true
scene. Let $O:\Omega\to E_{\obs}$ be the random observation, and let
$\wrand:\Omega\to\cS$ be the true completed world.  Because both spaces are
standard Borel, the joint law of $(\wrand,O)$ admits a regular conditional
probability of $\wrand$ given $O$.  Hence there is a kernel
\begin{equation}
\label{eq:true_kernel}
\kappa:E_{\obs}\times\cA\to[0,1],
\end{equation}
such that, for every world event $A\in\cA$ and observation event
$B\in\cB(E_{\obs})$,
\begin{equation}
\label{eq:disintegration}
\mathbb{P}(\wrand\in A,O\in B)
=
\int_B \kappa(\theta,A)(\mathbb{P}\circ O^{-1})(d\theta),
\end{equation}
where the integral averages the conditional mass $\kappa(\theta,A)$ over all
observation values $\theta\in B$ according to the observation law
$\mathbb{P}\circ O^{-1}$.  Equation~\eqref{eq:disintegration} is the
disintegration identity for the joint law of $(\wrand,O)$.  For a realized
observation value $\observ$, the true posterior is, for $A\in\cA$,
\begin{equation}
\label{eq:true_posterior}
P_{\observ}(A)
:=
\kappa(\observ,A).
\end{equation}
The posterior is unique for $(\mathbb{P}\circ O^{-1})$-almost every observation.
Below, $\observ$ is fixed, and the model law $Q_{\observ}$ is compared with
$P_{\observ}$.

\textbf{Spatio-semantic events.} Let $\cspace$ be the robot configuration
space with its Borel $\sigma$-algebra.  A spatio-semantic predicate is a
Boolean function of a robot configuration and a completed world:
\begin{equation}
\label{eq:indicator_def}
\consem{\mathrm{sem}}:\cspace\times\cS\to\{0,1\}.
\end{equation}
The predicate is one when the property $\mathrm{sem}$ holds.  Examples are
collision-freeness, target reachability or observability, and target evidence
inside a region.

\begin{assumption}[Measurable spatio-semantic predicates]
\label{ass:indicator_meas}
For each semantic property $\mathrm{sem}$, the map $\consem{\mathrm{sem}}$ is
$\cB(\cspace)\otimes\cA$-measurable.
\end{assumption}

In the discretized evaluation, these predicates are finite computations on
labeled points: coordinate projections, label comparisons, distance or
intersection tests, and finite Boolean operations.  Such operations are Borel
measurable.

For a fixed configuration $\state$, the predicate defines an event in the
completed-world space:
\begin{equation}
\label{eq:event_set}
\wsubsubset_{\mathrm{sem}}(\state)
:=
\{\winst\in\cS:\consem{\mathrm{sem}}(\state,\winst)=1\}.
\end{equation}
The model-side probability of that event is
\begin{equation}
\label{eq:model_event_probability}
\qspace_{\mathrm{sem}}(\state\mid\observ)
:=
Q_{\observ}\left(\wsubsubset_{\mathrm{sem}}(\state)\right)
=
\E_{\winst\sim Q_{\observ}}\left[
\consem{\mathrm{sem}}(\state,\winst)
\right].
\end{equation}
For a finite checked path, the corresponding event is the finite intersection
of the checked configuration events in the same completed world.  It is not a
product of marginal waypoint probabilities.
The detection, collision, and planning quantities
in~\cref{eq:pdet_main,eq:pcoll_main,eq:pplan_main} are special cases of these
events.

\textbf{Empirical priors.} For sampled worlds
$\winst^1,\ldots,\winst^N\overset{\mathrm{i.i.d.}}{\sim}Q_{\observ}$, the
empirical prior is the average of the predicate:
\begin{equation}
\label{eq:empirical_prior}
\gprobspace_{\mathrm{sem}}^{N}(\state\mid\observ)
:=
\frac{1}{N}\sum_{i=1}^{N}\consem{\mathrm{sem}}(\state,\winst^i).
\end{equation}
\begin{lemma}[Consistency of empirical indicator priors]
\label{lem:lln_indicator}
Fix $\observ\in E_{\obs}$, $\state\in\cspace$, and semantic property
$\mathrm{sem}$.  Under~\cref{ass:indicator_meas}, the empirical prior
in~\cref{eq:empirical_prior} satisfies
\begin{equation}
\label{eq:lln_convergence}
\gprobspace_{\mathrm{sem}}^{N}(\state\mid\observ)
\xrightarrow{\mathrm{a.s.}}
\qspace_{\mathrm{sem}}(\state\mid\observ),
\end{equation}
under the product sampling law $Q_{\observ}^{\otimes N}$.  Moreover, for every
$\varepsilon>0$,
\begin{equation}
\label{eq:hoeffding_rate}
Q_{\observ}^{\otimes N}\left(
\left|
\gprobspace_{\mathrm{sem}}^{N}(\state\mid\observ)
-
\qspace_{\mathrm{sem}}(\state\mid\observ)
\right|
\ge \varepsilon
\right)
\le
2\exp(-2N\varepsilon^2).
\end{equation}
\end{lemma}

\begin{proof}
Let $X_i:=\consem{\mathrm{sem}}(\state,\winst^i)$.  By measurability and the
i.i.d.~sampling assumption, $X_1,\ldots,X_N$ are i.i.d.~Bernoulli random
variables with mean
\[
\E[X_1]
=
Q_{\observ}\left(\wsubsubset_{\mathrm{sem}}(\state)\right)
=
\qspace_{\mathrm{sem}}(\state\mid\observ).
\]
The almost-sure convergence follows from the strong law of large
numbers~\cite{klenke2020}.  The concentration bound is Hoeffding's inequality
for bounded independent random variables~\cite{hoeffding1963}.
\end{proof}

\textbf{Finite path-success priors.} For a candidate trajectory
$\traj=(\state_0,\ldots,\state_K)$, let $\mathcal{K}(\traj)$ be the finite set
of checked configurations, including interpolated edge checks.  Define the
world-level success indicator
\begin{equation}
\label{eq:path_success_indicator_app}
H_{\traj}(\winst)
:=
\consem{\succlabel}(\state_K,\winst)
\prod_{\state\in\mathcal{K}(\traj)}
\consem{\freelabel}(\state,\winst).
\end{equation}
By~\cref{ass:indicator_meas}, $H_{\traj}$ is measurable.  The finite product is
an indicator because each factor is binary; it equals one exactly when the
terminal configuration succeeds and every checked configuration is
collision-free in the same sampled world.  The model-side path-success
probability is
\begin{equation}
\label{eq:path_success_law_app}
q_{\mathrm{plan}}(\traj\mid\observ)
:=
\E_{\winst\sim Q_{\observ}}\left[H_{\traj}(\winst)\right].
\end{equation}
For i.i.d.~sampled worlds, $H_{\traj}(\winst^1),\ldots,H_{\traj}(\winst^N)$
are i.i.d.~Bernoulli random variables, so
\begin{equation}
\label{eq:path_success_empirical_app}
\frac{1}{N}\sum_{i=1}^{N}H_{\traj}(\winst^i)
\xrightarrow{\mathrm{a.s.}}
q_{\mathrm{plan}}(\traj\mid\observ),
\end{equation}
and, for every $\varepsilon>0$,
\begin{equation}
\label{eq:path_success_hoeffding_app}
Q_{\observ}^{\otimes N}\left(
\left|
\frac{1}{N}\sum_{i=1}^{N}H_{\traj}(\winst^i)
-
q_{\mathrm{plan}}(\traj\mid\observ)
\right|
\ge \varepsilon
\right)
\le
2\exp(-2N\varepsilon^2).
\end{equation}
With $\consem{\succlabel}$ as terminal reach-or-observe success for the queried
object,~\cref{eq:path_success_law_app} is~\cref{eq:pplan_main}.

For any fixed finite collection $\mathcal{T}$ of candidate trajectories, a
union bound gives the uniform finite-class statement
\begin{equation}
\label{eq:finite_path_union_app}
Q_{\observ}^{\otimes N}\left(
\max_{\traj\in\mathcal{T}}
\left|
\frac{1}{N}\sum_{i=1}^{N}H_{\traj}(\winst^i)
-
q_{\mathrm{plan}}(\traj\mid\observ)
\right|
\ge \varepsilon
\right)
\le
2|\mathcal{T}|\exp(-2N\varepsilon^2).
\end{equation}
Roadmap details remain
in~\cref{app:eval_plan}.

\textbf{Model mismatch.} The empirical prior estimates probabilities under the
model law $Q_{\observ}$, while the target posterior law is $P_{\observ}$.
Total variation bounds~\cite{liese2006divergences} the difference between their
event probabilities.  For
probability measures $P$ and $Q$ on a measurable space $(E,\mathcal{E})$,
\begin{equation}
\label{eq:tv_def}
\TV(P,Q)
:=
\sup_{A\in\mathcal{E}}|P(A)-Q(A)|.
\end{equation}
Every completed-world event $A\in\cA$ satisfies
\begin{equation}
\label{eq:tv_event_bound}
\left|
P_{\observ}(A)-Q_{\observ}(A)
\right|
\le
\TV(P_{\observ},Q_{\observ}).
\end{equation}
Combining this inequality with~\cref{lem:lln_indicator}, if
$p_{\mathrm{true}}(\state\mid\observ):=
P_{\observ}(\wsubsubset_{\mathrm{sem}}(\state))$, then
\begin{equation}
\label{eq:empirical_posterior_bound}
Q_{\observ}^{\otimes N}\left(
\left|
\gprobspace_{\mathrm{sem}}^{N}(\state\mid\observ)
-
p_{\mathrm{true}}(\state\mid\observ)
\right|
\ge
\TV(P_{\observ},Q_{\observ})+\varepsilon
\right)
\le
2\exp(-2N\varepsilon^2).
\end{equation}
The total-variation term is posterior-model mismatch; the $\varepsilon$ term is
Monte Carlo error from finitely many sampled worlds.  The same decomposition
applies to the path-success event
\[
\{\winst\in\cS:H_{\traj}(\winst)=1\}.
\]

\section{Planning with Sampled Worlds}
\label{app:eval_plan}

\textbf{Planning objective.}
We formulate the planning objective as a continuous-space motion-planning
problem for a mobile robot in $\mathrm{SE}(2)$.
We use a PRM-style roadmap~\cite{kavraki1996prm} within the task-specific
search volume and evaluate candidate trajectories using the sampled worlds. Each
trajectory is scored by joint task success: it must remain collision-free with
respect to the sampled occupancy and reach the sampled target. For our
sampled-belief planner, all 100 generated occupancy--target hypotheses are
passed to the planner. A trajectory receives credit in a sampled world only when
it is collision-free in that world and its terminal state reaches the
corresponding sampled target. The planner therefore seeks a path that maximizes
the number of sampled worlds in which both conditions hold, following the
sampling-based uncertainty-planning framework of~\cite{lu2024sampling}.

Collision checking is performed along roadmap edges at finite interpolation
resolution, as is standard in sampling-based motion planning. In our
implementation, the interpolation step sizes are $0.05~\mathrm{m}$ for
translation and $0.10~\mathrm{rad}$ for rotation. Collision queries use
FCL~\cite{fcl}, the default collision-checking backend used by
OMPL~\cite{sucan2012open}. These checks approximate continuous edge validity by
dense interpolation, so the planner operates in continuous $\mathrm{SE}(2)$
while using a finite set of checked configurations for computation.

\textbf{Simulation setup.}
Each planning problem is instantiated for a Stretch mobile robot~\cite{stretch}.
The robot base is planned in $\mathrm{SE}(2)$, with configurations sampled
inside the task-specific search volume. The search volume comes from the
doorway-view setup and varies by task, rather than using a fixed global field of
view; details are in \cref{app:dataset}. For each generated world sample, we
fuse the observed ground-truth geometry from the visible region into the sampled
completion. As a result, all planners share the same observed geometry and differ only
in how they reason about the unobserved part of the room. We crop the floor
plane from each occupancy sample to avoid treating the floor as an obstacle,
then downsample and voxelize the occupancy point cloud for faster collision
checking.

The planning variants differ only in the occupancy representation and target
prior used during planning. 
\emph{Uniform prior} uses the visible scene and one
target location sampled uniformly from the hidden search region. \emph{1-sample}
uses the visible scene and one generated occupancy--target hypothesis.
\emph{Ours} uses all 100 generated occupancy--target hypotheses and optimizes
over a sampled belief. \emph{GT oracle} uses the ground-truth occupancy and
target location. The baselines therefore plan in a single hypothesized world,
whereas the sampled-belief planner plans over multiple plausible hidden worlds. A terminal state is
counted as reaching a planning hypothesis if it lies within $0.25~\mathrm{m}$ of
that hypothesis's target. All trajectories are then replayed against the same
ground-truth occupancy and target, and evaluated with the common
$1.0~\mathrm{m}$ success threshold in \cref{tab:planning_main}.

\textbf{Roadmap construction and search.}
For each task, we construct a roadmap using 3000 randomly sampled
configurations inside the search volume. The resulting roadmaps contain about
$5\times 10^4$ edges on average. Edge validity and edge weights are computed by
collision checking the interpolated edge configurations against the relevant
planning world or worlds. The final path is obtained by a greedy $A^{*}$ search
over the roadmap, using the edge and goal weights induced by the selected prior.
This objective is related to minimum-constraint-removal
planning~\cite{hauser2014mcr}, since a path is preferred when it violates fewer
sampled worlds.
\Cref{fig:plans} shows four representative cases where the sampled-belief
planner selects trajectories that terminate near the ground-truth object,
despite planning only from generated occupancy--target hypotheses.

\begin{figure}
    \centering
    \includegraphics[width=1\linewidth]{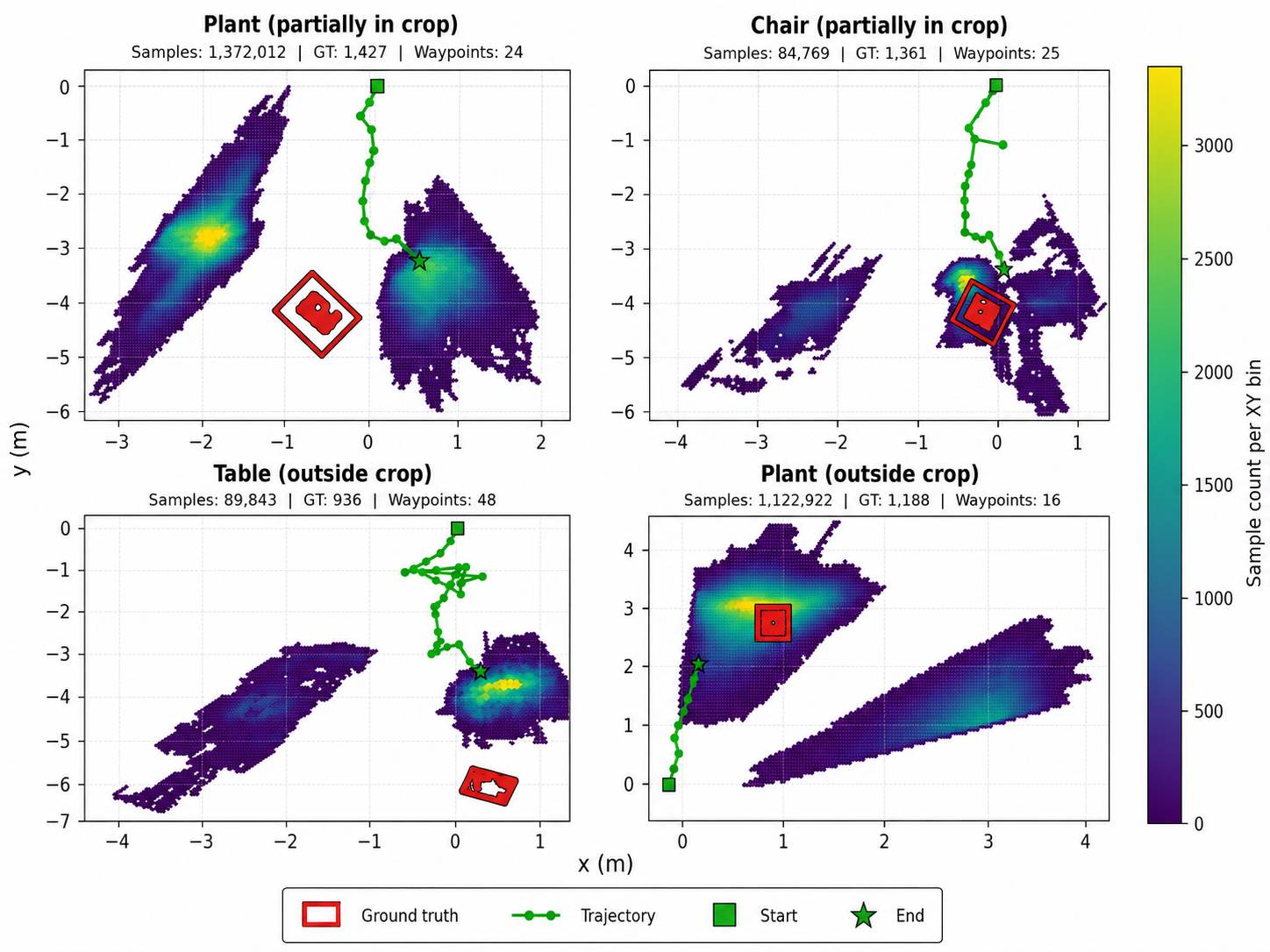}
    \caption{\textit{Qualitative plans.} 
    The plots show a 2D projection of the SE(2) paths visualized alongside the target object location heatmaps (projected to 2D), derived from 100 samples. Note that the occupancy probabilities are not visualized but are jointly optimized in the planner. Note that the sampled generative priors represent all the available information about the unobserved rooms (where the ground truth object location exists but is unknown). 
    Prominent heatmap regions being close to the ground truth represents the samples identifying that region to be likely for the object. 
    The path also ending close to such regions denote the ability of the robot to reach the location while accounting for its probability of collision. 
    The top-left and lower-left heatmaps show localization failure cases. 
    }
    \label{fig:plans}
\end{figure}

\begin{table}[t]
\centering
\small
\setlength{\tabcolsep}{5pt}
\renewcommand{\arraystretch}{1.08}
\caption{\small \textbf{Planning runtime.}
Runtime statistics are reported over the same 173 evaluated tasks. Time is reported here in milliseconds.}
\label{tab:runtime}
\begin{tabular}{lrrrr}
\toprule
Method
& Total mean
& Total median
& Roadmap mean
& $A^{*}$ mean \\
\midrule
Uniform prior
& 1567.0
& 1518.7
& 1498.9
& 0.73 \\
1-sample
& 1556.4
& 1490.8
& 1473.6
& 3.08 \\
Ours
& 79333.0
& 78023.1
& 79239.6
& 10.54 \\
GT oracle
& 1589.7
& 1517.6
& 1508.6
& 1.63 \\
\bottomrule
\end{tabular}
\end{table}

\textbf{Runtime analysis.}
The runtime breakdown in \cref{tab:runtime} shows that roadmap construction
dominates total planning time. These timings are for offline planning in the
benchmark simulator, not for a real-time deployment stack. For deterministic single-world planners,
including \emph{Uniform}, \emph{1-sample}, and \emph{GT oracle}, the mean total
runtime is about $1.6~\mathrm{s}$, with almost all of that time spent
constructing and collision-checking the roadmap. Our sampled-belief planner
checks each roadmap edge against 100 sampled worlds, increasing the mean runtime
to $79.3~\mathrm{s}$. The $A^{*}$ search itself remains comparatively small for
all methods: even for the sampled-belief planner, the mean search time is
$10.54~\mathrm{ms}$. The
timing therefore points to sequential collision checking over many sampled
worlds as the current computational bottleneck, rather than graph search. Since collision checks
across roadmap edges and sampled worlds are independent, parallelizing this
stage~\cite{vamp} is a candidate route to reduce sampled-belief planning time.
Quantifying that parallel implementation, or converting the planner into an
online controller, is beyond the current scope and is left to future work.

\section{Generative Ablations and Sensitivity Analysis}
\label{app:generative_ablations}

This section reports the ablations behind the fixed-budget generative results in
\cref{tab:gen_overall}.  Gate-threshold diagnostics use all 1000 queries and
are reported with the dataset-construction analysis in \cref{app:vlm_study}.

\textbf{Sample budget and placement tolerance.}
\Cref{fig:app_opr_budget_thresholds,fig:app_opr_budget_baselines} show how
OPR changes with sample budget.  We score the same samples at
three tolerances and compare OPR@1m against the geometry-only baselines from the
main paper.  \Cref{tab:app_bbox_budget_current} gives the numbers;
\cref{fig:app_distance_budget_curve} gives target-distance sensitivity.

\begin{figure}[!htbp]
\centering
\includegraphics[width=0.62\textwidth]{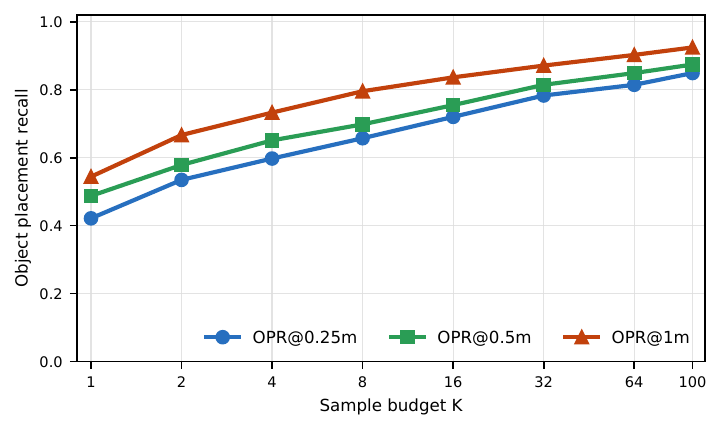}
\caption{Sample-budget ablation for object placement recall.  The x-axis is the
number of generated samples $K$ and the three curves vary only the placement
tolerance used by OPR.}
\label{fig:app_opr_budget_thresholds}
\end{figure}

\begin{figure}[!htbp]
\centering
\includegraphics[width=0.62\textwidth]{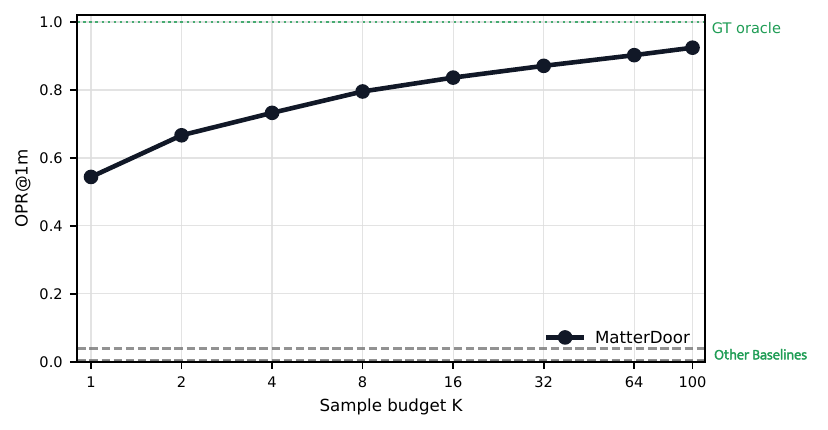}
\caption{OPR@1m against simple placement baselines.  MatterDoor's OPR increases with
sample budget, while the no-generation, uniform search-volume box,
search-volume centroid, and doorway-side frontier baselines remain low.}
\label{fig:app_opr_budget_baselines}
\end{figure}

\begin{figure}[!htbp]
\centering
\includegraphics[width=0.62\textwidth]{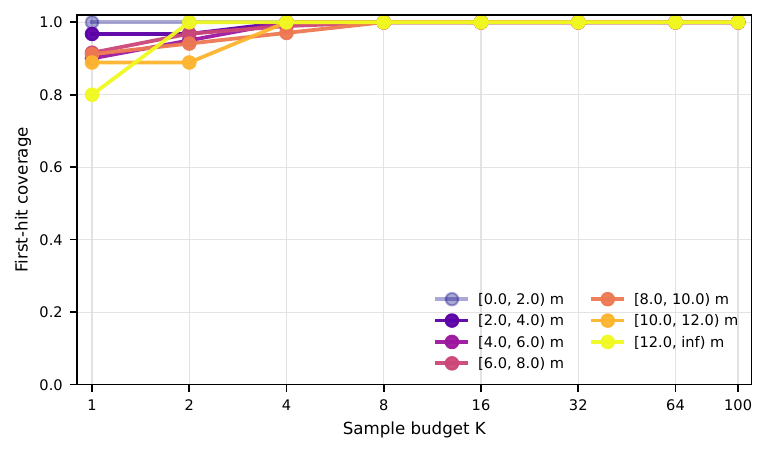}
\caption{Target-distance sensitivity for the sample-budget ablation.  The curve
tracks how the generated target-box distance to the GT box decreases as the
number of sampled hypotheses increases.}
\label{fig:app_distance_budget_curve}
\end{figure}

\textbf{Semantic calibration.}
\Cref{fig:app_semantic_kl_by_room} breaks down the room-wise contribution to the
semantic distribution score in \cref{tab:gen_overall}.  This is a generated
room--object ablation, not a dataset-construction diagnostic.

\begin{figure}[!htbp]
\centering
\includegraphics[width=0.62\textwidth]{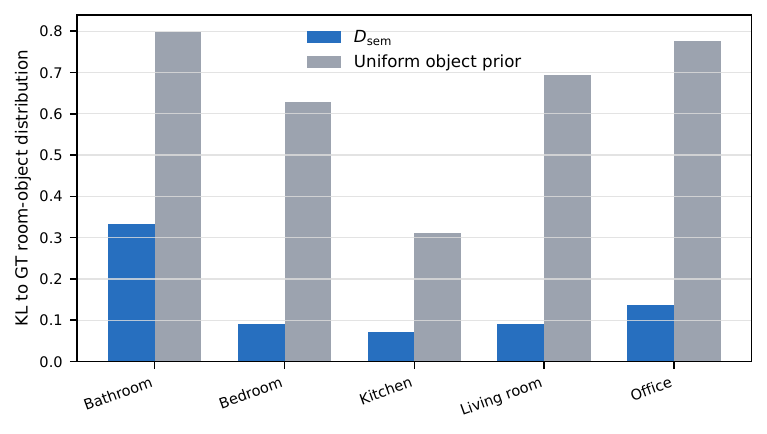}
\caption{Room-wise semantic calibration for the generated object distribution.
Lower KL indicates closer agreement with the Matterport3D room--object prior
used for MatterDoor scoring.}
\label{fig:app_semantic_kl_by_room}
\end{figure}

\textbf{Grouped sensitivity.}
Grouped sensitivities appear in
\cref{fig:app_opr_budget_by_scene,fig:app_opr_budget_by_object,fig:app_opr_budget_by_task}
and
\cref{tab:app_bbox_scene_current,tab:app_bbox_object_current,tab:app_bbox_task_current}.
Not-in-crop and farther targets generally need more samples, while visible or
nearby targets tend to hit earlier.

\begin{figure}[!htbp]
\centering
\includegraphics[width=0.72\textwidth]{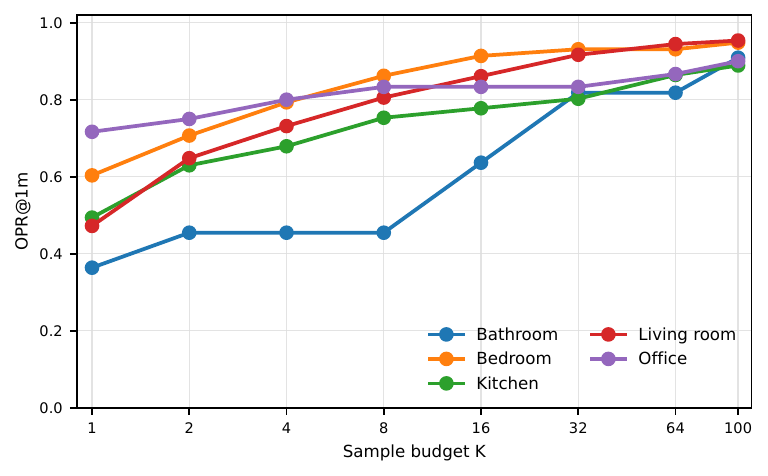}
\caption{Sample-budget OPR sensitivity by target room.  Each room uses the same
object-placement audit and OPR@1m scoring protocol.}
\label{fig:app_opr_budget_by_scene}
\end{figure}

\begin{figure}[!htbp]
\centering
\includegraphics[width=0.82\textwidth]{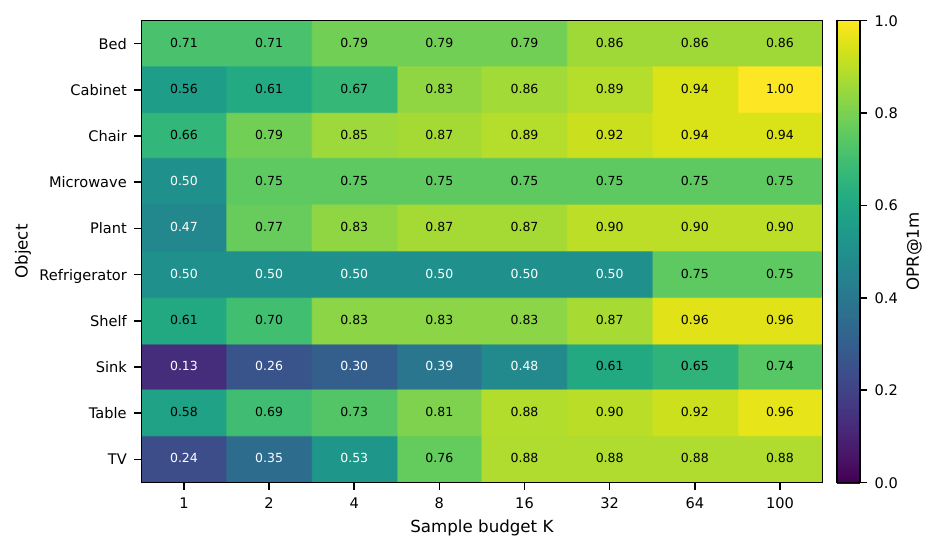}
\caption{Sample-budget OPR sensitivity by queried object.  Object categories
are shown vertically and generated-sample budgets horizontally.}
\label{fig:app_opr_budget_by_object}
\end{figure}

\begin{figure}[!htbp]
\centering
\includegraphics[width=0.62\textwidth]{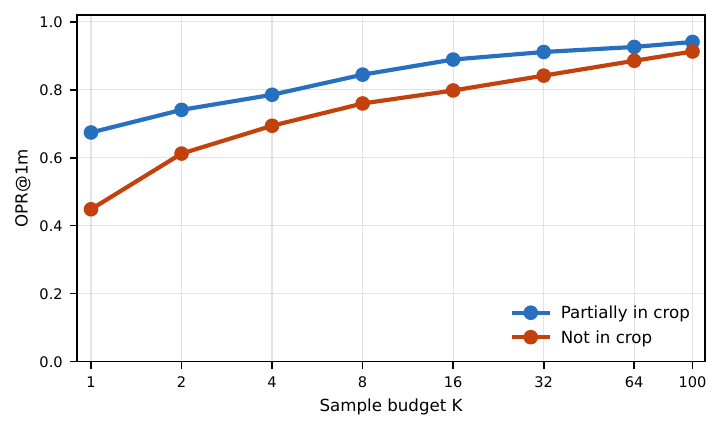}
\caption{Sample-budget OPR sensitivity by task type.  Out-of-crop targets need
more samples on average, while partially visible targets typically hit earlier.}
\label{fig:app_opr_budget_by_task}
\end{figure}

\section{Additional Result Breakdowns}
\label{app:additional_result_breakdowns}

This appendix supplements \cref{app:generative_ablations} with room-level,
object-level, sample-budget, gate-accounting, and planning-prior results for
MatterDoor-1k.  They give the details behind the main-paper results on the gate
policy, object appearance, object-box placement, $\Chamfer$, and ground-truth
replay behavior under the sampled planning prior.

The full ablation table also reports two degenerate sanity checks.
\emph{Object-everywhere} marks the queried object throughout the search region,
giving recall without usable localization.  \emph{Everything-everywhere} marks
every object class throughout the search region, illustrating a
high-recall/high-false-positive extreme.
These controls are metric extremes rather than realistic baselines, so the main
table omits them.

\Cref{tab:gen_breakdown}
summarizes scene- and object-level generative results.
\Cref{tab:app_current_metric_summary,tab:app_room_generation_accounting,tab:app_object_generation_accounting,tab:app_object_scene_generation_current}
summarize the MatterDoor-1k gate-accounting counts.
\Cref{tab:app_room_gate_accuracy_current,tab:app_object_gate_accuracy_current}
break down VLM gate behavior by room and object.
\Cref{tab:app_first_hit_global_current,tab:app_first_hit_scene_current,tab:app_first_hit_object_current}
report planner-facing first-hit coverage, while
\cref{tab:app_scene_gt_metrics,tab:app_object_gt_metrics} report strict 3D
generation scores by room and object.

\begin{table*}[t]
\centering
\scriptsize
\setlength{\tabcolsep}{3.2pt}
\renewcommand{\arraystretch}{0.90}
\caption{\textit{Generative 3D evaluation breakdown.}
Scene- and object-level results use the same metrics as \cref{tab:gen_overall}.
The last four columns report OPR@1m for the Uniform-random (U), Centroid (C),
Frontier (F), and proposed priors. Dashes denote metrics that are not
applicable.}
\label{tab:gen_breakdown}
\begin{tabularx}{\textwidth}{@{}l r *{8}{>{\centering\arraybackslash}X}@{}}
\toprule
& & & & & & \multicolumn{4}{c}{OPR@1m~$\uparrow$} \\
\cmidrule(l){7-10}
Breakdown & $n$ & $D_s~\downarrow$ & $p_d~\uparrow$ & $CD~\downarrow$ &
$\mathcal{E}~\downarrow$ & U & C & F & \textbf{Ours} \\
\midrule
\multicolumn{10}{@{}l}{\textit{By scene}} \\
\midrule
Bathroom & 11  & 0.332 & 0.664 & 2.10 & 1.96 & 0.005 & 0.000 & 0.000 & \textbf{0.909} \\
Bedroom  & 58  & 0.091 & 0.928 & 1.49 & 1.75 & 0.059 & 0.000 & 0.017 & \textbf{0.948} \\
Kitchen  & 81  & 0.071 & 0.849 & 1.88 & 2.31 & 0.031 & 0.012 & 0.000 & \textbf{0.889} \\
Living   & 108 & 0.091 & 0.923 & 2.13 & 2.50 & 0.026 & 0.009 & 0.000 & \textbf{0.954} \\
Office   & 60  & 0.136 & 0.949 & 1.93 & 2.27 & 0.062 & 0.000 & 0.000 & \textbf{0.900} \\
\midrule
\multicolumn{10}{@{}l}{\textit{By object}} \\
\midrule
bed     & 14 & --    & 0.976 & 1.13 & 1.70 & 0.125 & 0.000 & 0.071 & \textbf{0.857} \\
cabinet & 36 & --    & 0.941 & 1.84 & 2.18 & 0.057 & 0.028 & 0.000 & \textbf{1.000} \\
chair   & 89 & --    & 0.946 & 1.95 & 2.31 & 0.034 & 0.000 & 0.000 & \textbf{0.944} \\
micro.  & 4  & --    & 0.998 & 1.46 & 1.73 & 0.000 & 0.000 & 0.000 & \textbf{0.750} \\
plant   & 30 & --    & 1.000 & 2.02 & 2.83 & 0.020 & 0.000 & 0.000 & \textbf{0.900} \\
fridge  & 4  & --    & 0.835 & 1.71 & 1.45 & 0.075 & 0.000 & 0.000 & \textbf{0.750} \\
shelf   & 23 & --    & 0.836 & 2.06 & 2.39 & 0.076 & 0.043 & 0.000 & \textbf{0.957} \\
sink    & 23 & --    & 0.577 & 1.86 & 2.51 & 0.007 & 0.000 & 0.000 & \textbf{0.739} \\
table   & 78 & --    & 0.922 & 2.02 & 2.18 & 0.036 & 0.000 & 0.000 & \textbf{0.962} \\
TV      & 17 & --    & 0.765 & 1.74 & 1.65 & 0.006 & 0.000 & 0.000 & \textbf{0.882} \\
\bottomrule
\end{tabularx}
\end{table*}

\begin{table*}[tbp]
\centering
\scriptsize
\setlength{\tabcolsep}{4pt}
\renewcommand{\arraystretch}{1.04}
\caption{Additional MatterDoor-1k gate and support counts.  Gate accounting uses
the 1000 crop-side/object queries; strict 3D metrics use the 184 verified target
instances with Matterport geometry.}
\label{tab:app_current_metric_summary}
\begin{tabularx}{\textwidth}{@{}L{0.21\textwidth}Yrrr@{}}
\toprule
Quantity & Definition & Count & Denom. & Rate \\
\midrule
Queries & Doorway side paired with one object name & 1000 & -- & -- \\
Object-present queries & Doorway/object pairs where the object is in the target room & 144 & -- & -- \\
3D instances & Verified object instances with 3D geometry & 184 & -- & -- \\
VLM-pass generated & $p_{\mathrm{VLM}}\ge 0.40$ & 398 & 1000 & 0.398 \\
GT-present fallback & Real targets retained after VLM miss & 37 & 144 & 0.257 \\
Active generation/fallback & VLM-pass plus GT-present fallback & 435 & 1000 & 0.435 \\
GT-absent skipped & Absent queries not generated & 565 & 856 & 0.660 \\
Unlikely generated & Room-prior-zero generated absent queries & 8 & 573 & 0.014 \\
First joint hit by $K{=}4$ & Object appears and region has free space & 181 & 184 & 0.984 \\
BBox-local hit by $K{=}16$ & Generated object box within 0.5 m & 141 & 184 & 0.766 \\
\bottomrule
\end{tabularx}
\end{table*}

\begin{table*}[!tp]
\centering
\scriptsize
\setlength{\tabcolsep}{3pt}
\renewcommand{\arraystretch}{1.04}
\caption{MatterDoor-1k VLM/GT groups by target room.  Generated absent means
GT-absent queries that pass the VLM gate; skipped absent means GT-absent queries
below the gate; GT fallback means GT-present queries below the gate. Active
counts all GT-present queries plus generated GT-absent queries.}
\label{tab:app_room_generation_accounting}
\begin{tabularx}{\textwidth}{@{}l *{8}{>{\centering\arraybackslash}X}@{}}
\toprule
Room & Queries & GT queries & 3D instances & Gen. absent & Skip absent & Active & GT fallb. & Unlikely gen. \\
\midrule
Bathroom & 40 & 5 & 6 & 11 & 24 & 16 & 3 & 1 \\
Bedroom & 250 & 31 & 34 & 66 & 153 & 97 & 16 & 0 \\
Kitchen & 180 & 41 & 56 & 87 & 52 & 128 & 2 & 0 \\
Living room & 370 & 43 & 57 & 104 & 223 & 147 & 9 & 6 \\
Office & 160 & 24 & 31 & 23 & 113 & 47 & 7 & 1 \\
\midrule
\textbf{Total} & 1000 & 144 & 184 & 291 & 565 & 435 & 37 & 8 \\
\bottomrule
\end{tabularx}
\end{table*}

\begin{table*}[!t]
\centering
\scriptsize
\setlength{\tabcolsep}{3pt}
\renewcommand{\arraystretch}{1.04}
\caption{MatterDoor-1k VLM/GT groups by queried object.  Each object has 100
queries before gate and fallback are applied. Active counts all GT-present
queries plus generated GT-absent queries.}
\label{tab:app_object_generation_accounting}
\begin{tabularx}{\textwidth}{@{}l *{8}{>{\centering\arraybackslash}X}@{}}
\toprule
Object & Queries & GT queries & 3D instances & Gen. absent & Skip absent & Active & GT fallb. & Unlikely gen. \\
\midrule
bed & 100 & 7 & 8 & 18 & 75 & 25 & 3 & 1 \\
cabinet & 100 & 18 & 26 & 36 & 46 & 54 & 3 & 0 \\
chair & 100 & 28 & 41 & 56 & 16 & 84 & 7 & 0 \\
microwave & 100 & 3 & 5 & 17 & 80 & 20 & 0 & 2 \\
plant & 100 & 14 & 16 & 25 & 61 & 39 & 9 & 0 \\
refrigerator & 100 & 3 & 3 & 17 & 80 & 20 & 0 & 2 \\
shelf & 100 & 13 & 14 & 21 & 66 & 34 & 7 & 0 \\
sink & 100 & 12 & 13 & 12 & 76 & 24 & 1 & 2 \\
table & 100 & 33 & 45 & 65 & 2 & 98 & 1 & 0 \\
TV & 100 & 13 & 13 & 24 & 63 & 37 & 6 & 1 \\
\midrule
\textbf{Total} & 1000 & 144 & 184 & 291 & 565 & 435 & 37 & 8 \\
\bottomrule
\end{tabularx}
\end{table*}

\begin{table*}[!t]
\centering
\scriptsize
\setlength{\tabcolsep}{2.4pt}
\renewcommand{\arraystretch}{1.04}
\caption{VLM gate behavior by target room.  GT recall counts real objects found by the gate; plausibility precision also credits absent objects that are reasonable for the room.}
\label{tab:app_room_gate_accuracy_current}
\begin{tabularx}{\textwidth}{@{}l *{10}{>{\centering\arraybackslash}X}@{}}
\toprule
Room & Queries & GT & Pass & TP & FN & Gen. absent & Unlikely & Mean score & GT recall & Plaus. prec. \\
\midrule
Bathroom & 40 & 5 & 13 & 2 & 3 & 11 & 1 & 0.269 & 0.400 & 0.923 \\
Bedroom & 250 & 31 & 81 & 15 & 16 & 66 & 0 & 0.302 & 0.484 & 1.000 \\
Kitchen & 180 & 41 & 126 & 39 & 2 & 87 & 0 & 0.480 & 0.951 & 1.000 \\
Living room & 370 & 43 & 138 & 34 & 9 & 104 & 6 & 0.301 & 0.791 & 0.957 \\
Office & 160 & 24 & 40 & 17 & 7 & 23 & 1 & 0.253 & 0.708 & 0.975 \\
\bottomrule
\end{tabularx}
\end{table*}

\begin{table*}[thp]
\centering
\scriptsize
\setlength{\tabcolsep}{2.4pt}
\renewcommand{\arraystretch}{1.04}
\caption{VLM gate behavior by queried object.  The fallback column in the count tables keeps the missed real objects shown here.}
\label{tab:app_object_gate_accuracy_current}
\begin{tabularx}{\textwidth}{@{}l *{10}{>{\centering\arraybackslash}X}@{}}
\toprule
Object & Queries & GT & Pass & TP & FN & Gen. absent & Unlikely & Mean score & GT recall & Plaus. prec. \\
\midrule
bed & 100 & 7 & 22 & 4 & 3 & 18 & 1 & 0.224 & 0.571 & 0.955 \\
cabinet & 100 & 18 & 51 & 15 & 3 & 36 & 0 & 0.487 & 0.833 & 1.000 \\
chair & 100 & 28 & 77 & 21 & 7 & 56 & 0 & 0.542 & 0.750 & 1.000 \\
microwave & 100 & 3 & 20 & 3 & 0 & 17 & 2 & 0.115 & 1.000 & 0.900 \\
plant & 100 & 14 & 30 & 5 & 9 & 25 & 0 & 0.334 & 0.357 & 1.000 \\
refrigerator & 100 & 3 & 20 & 3 & 0 & 17 & 2 & 0.115 & 1.000 & 0.900 \\
shelf & 100 & 13 & 27 & 6 & 7 & 21 & 0 & 0.381 & 0.462 & 1.000 \\
sink & 100 & 12 & 23 & 11 & 1 & 12 & 2 & 0.203 & 0.917 & 0.913 \\
table & 100 & 33 & 97 & 32 & 1 & 65 & 0 & 0.579 & 0.970 & 1.000 \\
TV & 100 & 13 & 31 & 7 & 6 & 24 & 1 & 0.265 & 0.538 & 0.968 \\
\bottomrule
\end{tabularx}
\end{table*}

\begin{table*}[tbp]
\centering
\tiny
\setlength{\tabcolsep}{2pt}
\renewcommand{\arraystretch}{1.04}
\caption{Generation counts for all 50 object-room pairs in the 1000-query table.
These counts correspond to the active gate/fallback heatmap in
\cref{fig:object_scene_heatmaps}: generated absent queries are VLM-pass
GT-absent probes, and GT fallback keeps missed real objects in the evaluation
denominator.}
\label{tab:app_object_scene_generation_current}
\begin{tabularx}{\textwidth}{@{}l l *{7}{>{\centering\arraybackslash}X}@{}}
\toprule
Object & Room & Queries & GT queries & 3D inst. & Gen. absent & Skipped absent & GT fallback & Unlikely gen. \\
\midrule
bed & Bathroom & 4 & 0 & 0 & 0 & 4 & 0 & 0 \\
bed & Bedroom & 25 & 7 & 8 & 15 & 3 & 3 & 0 \\
bed & Kitchen & 18 & 0 & 0 & 0 & 18 & 0 & 0 \\
bed & Living room & 37 & 0 & 0 & 2 & 35 & 0 & 0 \\
bed & Office & 16 & 0 & 0 & 1 & 15 & 0 & 1 \\
cabinet & Bathroom & 4 & 0 & 0 & 3 & 1 & 0 & 0 \\
cabinet & Bedroom & 25 & 1 & 1 & 19 & 5 & 0 & 0 \\
cabinet & Kitchen & 18 & 12 & 20 & 6 & 0 & 0 & 0 \\
cabinet & Living room & 37 & 4 & 4 & 6 & 27 & 2 & 0 \\
cabinet & Office & 16 & 1 & 1 & 2 & 13 & 1 & 0 \\
chair & Bathroom & 4 & 1 & 1 & 1 & 2 & 1 & 0 \\
chair & Bedroom & 25 & 6 & 7 & 6 & 13 & 4 & 0 \\
chair & Kitchen & 18 & 5 & 8 & 13 & 0 & 0 & 0 \\
chair & Living room & 37 & 6 & 10 & 30 & 1 & 1 & 0 \\
chair & Office & 16 & 10 & 15 & 6 & 0 & 1 & 0 \\
microwave & Bathroom & 4 & 0 & 0 & 0 & 4 & 0 & 0 \\
microwave & Bedroom & 25 & 0 & 0 & 0 & 25 & 0 & 0 \\
microwave & Kitchen & 18 & 3 & 5 & 15 & 0 & 0 & 0 \\
microwave & Living room & 37 & 0 & 0 & 2 & 35 & 0 & 2 \\
microwave & Office & 16 & 0 & 0 & 0 & 16 & 0 & 0 \\
plant & Bathroom & 4 & 0 & 0 & 1 & 3 & 0 & 0 \\
plant & Bedroom & 25 & 5 & 5 & 4 & 16 & 5 & 0 \\
plant & Kitchen & 18 & 2 & 4 & 0 & 16 & 2 & 0 \\
plant & Living room & 37 & 6 & 6 & 18 & 13 & 1 & 0 \\
plant & Office & 16 & 1 & 1 & 2 & 13 & 1 & 0 \\
refrigerator & Bathroom & 4 & 0 & 0 & 0 & 4 & 0 & 0 \\
refrigerator & Bedroom & 25 & 0 & 0 & 0 & 25 & 0 & 0 \\
refrigerator & Kitchen & 18 & 3 & 3 & 15 & 0 & 0 & 0 \\
refrigerator & Living room & 37 & 0 & 0 & 2 & 35 & 0 & 2 \\
refrigerator & Office & 16 & 0 & 0 & 0 & 16 & 0 & 0 \\
shelf & Bathroom & 4 & 0 & 0 & 3 & 1 & 0 & 0 \\
shelf & Bedroom & 25 & 1 & 1 & 1 & 23 & 1 & 0 \\
shelf & Kitchen & 18 & 5 & 5 & 13 & 0 & 0 & 0 \\
shelf & Living room & 37 & 3 & 3 & 3 & 31 & 2 & 0 \\
shelf & Office & 16 & 4 & 5 & 1 & 11 & 4 & 0 \\
sink & Bathroom & 4 & 3 & 4 & 1 & 0 & 1 & 0 \\
sink & Bedroom & 25 & 0 & 0 & 0 & 25 & 0 & 0 \\
sink & Kitchen & 18 & 9 & 9 & 9 & 0 & 0 & 0 \\
sink & Living room & 37 & 0 & 0 & 2 & 35 & 0 & 2 \\
sink & Office & 16 & 0 & 0 & 0 & 16 & 0 & 0 \\
table & Bathroom & 4 & 1 & 1 & 1 & 2 & 1 & 0 \\
table & Bedroom & 25 & 5 & 6 & 20 & 0 & 0 & 0 \\
table & Kitchen & 18 & 2 & 2 & 16 & 0 & 0 & 0 \\
table & Living room & 37 & 17 & 27 & 20 & 0 & 0 & 0 \\
table & Office & 16 & 8 & 9 & 8 & 0 & 0 & 0 \\
TV & Bathroom & 4 & 0 & 0 & 1 & 3 & 0 & 1 \\
TV & Bedroom & 25 & 6 & 6 & 1 & 18 & 3 & 0 \\
TV & Kitchen & 18 & 0 & 0 & 0 & 18 & 0 & 0 \\
TV & Living room & 37 & 7 & 7 & 19 & 11 & 3 & 0 \\
TV & Office & 16 & 0 & 0 & 3 & 13 & 0 & 0 \\
\bottomrule
\end{tabularx}
\end{table*}

\begin{table*}[!th]
\centering
\scriptsize
\setlength{\tabcolsep}{2.4pt}
\renewcommand{\arraystretch}{1.03}
\caption{Averages by task type.  $\mathcal{E}$ is MatterDoor placement energy
in meters.  Best OPR@1m baselines use the same task split as MatterDoor, and
Uniform denotes the $K=100$ random search-volume control.}
\label{tab:main_task_averages}
\begin{tabular}{@{}lrrrrrrrrr@{}}
\toprule
Task type & $n$ & $p_{\mathrm{det}}\uparrow$ & $\Chamfer\downarrow$ & $\mathcal{E}\downarrow$ & Uniform & Centroid & Frontier & \textbf{best OPR@1m} & GT \\
\midrule
Partially in crop & 135 & 0.895 & 1.70 & 1.91 & 0.049 & 0.015 & 0.007 & \textbf{0.941} & 1.000 \\
Not in crop & 183 & 0.905 & 2.07 & 2.51 & 0.033 & 0.000 & 0.000 & \textbf{0.913} & 1.000 \\

\bottomrule
\end{tabular}
\end{table*}

\begin{table*}[thp]
\centering
\scriptsize
\setlength{\tabcolsep}{3pt}
\renewcommand{\arraystretch}{1.04}
\caption{Sample-count coverage for planner-facing joint hits on the 184
GT-present target instances.  A hit requires the generated target to be detected
and the sample to pass the search-volume occupancy check.}
\label{tab:app_first_hit_global_current}
\begin{tabular}{@{}lrrrrrrrrrrr@{}}
\toprule
Set & $n$ & $C_1$ & $C_2$ & $C_4$ & $C_8$ & $C_{16}$ & $C_{32}$ & $C_{100}$ & Median first & P90 first & Mean hits \\
\midrule
All GT-present instances & 184 & 0.897 & 0.940 & 0.984 & 1.000 & 1.000 & 1.000 & 1.000 & 1.0 & 1.7 & 90.3 \\
\bottomrule
\end{tabular}
\end{table*}

\begin{table*}[!th]
\centering
\scriptsize
\setlength{\tabcolsep}{3pt}
\renewcommand{\arraystretch}{1.04}
\caption{Planner-facing first-hit coverage by target room on the 184 GT-present
target instances.}
\label{tab:app_first_hit_scene_current}
\begin{tabular}{@{}lrrrrrrrrr@{}}
\toprule
Room & $n$ & $C_1$ & $C_4$ & $C_8$ & $C_{16}$ & $C_{100}$ & Median first & P90 first & Mean hits \\
\midrule
Bathroom & 6 & 0.833 & 1.000 & 1.000 & 1.000 & 1.000 & 1.0 & 2.0 & 71.3 \\
Bedroom & 34 & 1.000 & 1.000 & 1.000 & 1.000 & 1.000 & 1.0 & 1.0 & 94.8 \\
Kitchen & 56 & 0.821 & 0.964 & 1.000 & 1.000 & 1.000 & 1.0 & 2.5 & 87.8 \\
Living room & 57 & 0.877 & 0.982 & 1.000 & 1.000 & 1.000 & 1.0 & 2.0 & 89.8 \\
Office & 31 & 0.968 & 1.000 & 1.000 & 1.000 & 1.000 & 1.0 & 1.0 & 94.6 \\
\bottomrule
\end{tabular}
\end{table*}

\begin{table*}[tbp]
\centering
\scriptsize
\setlength{\tabcolsep}{3pt}
\renewcommand{\arraystretch}{1.04}
\caption{Planner-facing first-hit coverage by queried object on the 184
GT-present target instances.}
\label{tab:app_first_hit_object_current}
\begin{tabular}{@{}lrrrrrrrrr@{}}
\toprule
Object & $n$ & $C_1$ & $C_4$ & $C_8$ & $C_{16}$ & $C_{100}$ & Median first & P90 first & Mean hits \\
\midrule
bed & 8 & 1.000 & 1.000 & 1.000 & 1.000 & 1.000 & 1.0 & 1.0 & 99.9 \\
cabinet & 26 & 0.923 & 1.000 & 1.000 & 1.000 & 1.000 & 1.0 & 1.0 & 98.2 \\
chair & 41 & 0.927 & 1.000 & 1.000 & 1.000 & 1.000 & 1.0 & 1.0 & 94.0 \\
microwave & 5 & 1.000 & 1.000 & 1.000 & 1.000 & 1.000 & 1.0 & 1.0 & 99.8 \\
plant & 16 & 1.000 & 1.000 & 1.000 & 1.000 & 1.000 & 1.0 & 1.0 & 100.0 \\
refrigerator & 3 & 1.000 & 1.000 & 1.000 & 1.000 & 1.000 & 1.0 & 1.0 & 78.0 \\
shelf & 14 & 0.714 & 1.000 & 1.000 & 1.000 & 1.000 & 1.0 & 2.7 & 82.1 \\
sink & 13 & 0.538 & 0.846 & 1.000 & 1.000 & 1.000 & 1.0 & 4.6 & 60.3 \\
table & 45 & 0.933 & 0.978 & 1.000 & 1.000 & 1.000 & 1.0 & 1.0 & 91.0 \\
TV & 13 & 0.923 & 1.000 & 1.000 & 1.000 & 1.000 & 1.0 & 1.0 & 80.9 \\
\bottomrule
\end{tabular}
\end{table*}

\begin{table*}[tbp]
\centering
\scriptsize
\setlength{\tabcolsep}{3.5pt}
\renewcommand{\arraystretch}{1.04}
\caption{Full ablation details for the generative metrics.  Mean OPR@1m is the
expected single-draw target-placement hit rate; best OPR@1m is the best-of-budget
counterpart.  The uniform random control reports the $K=1$ random sweep for mean
OPR and 100 independent search-volume draws for best OPR; centroid and frontier
are deterministic one-hypothesis controls.  $E_{\mathrm{place}}$ is the
placement energy score in meters, which rewards target agreement while
penalizing implausible spread; lower is better.  The object-everywhere and
everything-everywhere controls are sanity checks: they give recall without usable
localization.}
\label{tab:app_full_ablation_details}
\begin{tabular}{@{}lrrrrrr@{}}
\toprule
Method & $D_{\mathrm{sem}}\downarrow$ & $p_{\mathrm{det}}\uparrow$ & mean OPR@1m $\uparrow$ & best OPR@1m $\uparrow$ & $\Chamfer\downarrow$ & $E_{\mathrm{place}}\downarrow$ \\
\midrule
Uniform object prior & 0.642 & -- & -- & -- & -- & -- \\
No generation & -- & 0.000 & 0.000 & 0.000 & -- & -- \\
Uniform random ($K=100$) & -- & -- & 0.001 & 0.039 & -- & 3.42 \\
Centroid (1 hyp.) & -- & -- & 0.006 & 0.006 & -- & 5.04 \\
Frontier (1 hyp.) & -- & -- & 0.003 & 0.003 & -- & 10.07 \\
Object-everywhere & -- & 1.000 & -- & 1.000 & -- & -- \\
Everything-everywhere & -- & 1.000 & -- & 1.000 & -- & -- \\
\textbf{MatterDoor sampled} & \textbf{0.144} & \textbf{0.903} & \textbf{0.901} & \textbf{0.925} & 1.81 & 2.26 \\
GT oracle & 0.000 & 1.000 & 1.000 & 1.000 & 0.00 & 0.00 \\
\bottomrule
\end{tabular}
\end{table*}

\begin{table*}[tbp]
\centering
\scriptsize
\setlength{\tabcolsep}{5pt}
\renewcommand{\arraystretch}{1.04}
\caption{Full sampling-budget ablation for OPR.  MatterDoor settings vary the
number of generated samples $K$; the uniform random control reports its $K=100$
seeded draw result, while centroid and frontier are deterministic one-hypothesis
controls shown only as fixed references.}
\label{tab:app_bbox_budget_current}
\begin{tabular}{@{}lrrrr@{}}
\toprule
Method / $K$ & $n$ & OPR@0.25m $\uparrow$ & OPR@0.5m $\uparrow$ & OPR@1m $\uparrow$ \\
\midrule
$K=1$ & 184 & 0.421 & 0.487 & 0.544 \\
$K=2$ & 184 & 0.535 & 0.579 & 0.667 \\
$K=4$ & 184 & 0.597 & 0.651 & 0.733 \\
$K=8$ & 184 & 0.657 & 0.698 & 0.796 \\
$K=16$ & 184 & 0.720 & 0.755 & 0.836 \\
$K=32$ & 184 & 0.783 & 0.814 & 0.871 \\
$K=64$ & 184 & 0.814 & 0.849 & 0.903 \\
$K=100$ & 184 & 0.849 & 0.874 & 0.925 \\
\midrule
No generation & 184 & 0.000 & 0.000 & 0.000 \\
Uniform random ($K=100$) & 184 & -- & -- & 0.039 \\
Centroid (1 hyp.) & 184 & 0.000 & 0.003 & 0.006 \\
Frontier (1 hyp.) & 184 & 0.000 & 0.000 & 0.003 \\
GT oracle & 184 & 1.000 & 1.000 & 1.000 \\
\bottomrule
\end{tabular}
\end{table*}

\begin{table*}[tbp]
\centering
\scriptsize
\setlength{\tabcolsep}{4pt}
\renewcommand{\arraystretch}{1.04}
\caption{3D scores by object on the 184 verified target instances.  The split
columns show how many targets are partially visible in the crop versus outside
the crop.  $\Chamfer$ compares generated geometry with Matterport3D geometry in
the scored region.}
\label{tab:app_object_gt_metrics}
\begin{tabular}{@{}lrrrrrr@{}}
\toprule
Object & $n$ & Partial & Not in crop & $p_{\mathrm{det}}\uparrow$ & $\Chamfer\downarrow$ & $\Chamfer_{\min}\downarrow$ \\
\midrule
bed & 8 & 8 & 0 & 0.999 & 1.00 & 0.69 \\
cabinet & 26 & 13 & 13 & 0.982 & 1.91 & 1.26 \\
chair & 41 & 18 & 23 & 0.940 & 1.80 & 0.86 \\
microwave & 5 & 1 & 4 & 0.998 & 1.40 & 0.81 \\
plant & 16 & 7 & 9 & 1.000 & 1.75 & 0.89 \\
refrigerator & 3 & 1 & 2 & 0.780 & 1.71 & 0.45 \\
shelf & 14 & 7 & 7 & 0.822 & 2.14 & 1.01 \\
sink & 13 & 8 & 5 & 0.603 & 1.77 & 0.87 \\
table & 45 & 25 & 20 & 0.910 & 1.99 & 0.85 \\
TV & 13 & 3 & 10 & 0.809 & 1.54 & 0.50 \\
\midrule
\textbf{Total} & 184 & 91 & 93 & 0.903 & 1.81 & 0.89 \\
\bottomrule
\end{tabular}
\end{table*}

\begin{table*}[tbp]
\centering
\scriptsize
\setlength{\tabcolsep}{3pt}
\renewcommand{\arraystretch}{1.04}
\caption{Object-box hit rate by target room.  A hit requires a generated object box near the ground-truth object box.}
\label{tab:app_bbox_scene_current}
\begin{tabular}{@{}lrrrrrrr@{}}
\toprule
Room & $n$ & Median gap & P90 gap & $\leq0.25$m & $\leq0.50$m & $\leq1.00$m & Median first $\leq0.5$m \\
\midrule
Bathroom & 6 & 0.000 & 0.031 & 1.000 & 1.000 & 1.000 & 7.5 \\
Bedroom & 34 & 0.000 & 0.072 & 0.912 & 0.941 & 1.000 & 1.0 \\
Kitchen & 56 & 0.000 & 0.730 & 0.786 & 0.857 & 0.929 & 2.0 \\
Living room & 57 & 0.000 & 0.790 & 0.860 & 0.877 & 0.947 & 2.0 \\
Office & 31 & 0.000 & 0.000 & 0.903 & 0.903 & 0.935 & 1.0 \\
\bottomrule
\end{tabular}
\end{table*}

\begin{table*}[tbp]
\centering
\scriptsize
\setlength{\tabcolsep}{3pt}
\renewcommand{\arraystretch}{1.04}
\caption{Object-box hit rate by queried object on the 184 verified target
instances.}
\label{tab:app_bbox_object_current}
\begin{tabular}{@{}lrrrrrrr@{}}
\toprule
Object & $n$ & Median gap & P90 gap & $\leq0.25$m & $\leq0.50$m & $\leq1.00$m & Median first $\leq0.5$m \\
\midrule
bed & 8 & 0.000 & 0.247 & 0.875 & 0.875 & 1.000 & 1.0 \\
cabinet & 26 & 0.000 & 0.337 & 0.846 & 0.923 & 1.000 & 2.0 \\
chair & 41 & 0.000 & 0.084 & 0.927 & 0.927 & 0.976 & 1.0 \\
microwave & 5 & 0.829 & 4.751 & 0.400 & 0.400 & 0.600 & 1.0 \\
plant & 16 & 0.000 & 0.312 & 0.812 & 0.938 & 0.938 & 2.0 \\
refrigerator & 3 & 0.000 & 0.038 & 1.000 & 1.000 & 1.000 & 1.0 \\
shelf & 14 & 0.000 & 0.679 & 0.857 & 0.857 & 0.929 & 1.0 \\
sink & 13 & 0.000 & 1.750 & 0.769 & 0.769 & 0.846 & 4.0 \\
table & 45 & 0.000 & 0.062 & 0.911 & 0.933 & 0.978 & 1.5 \\
TV & 13 & 0.000 & 0.660 & 0.769 & 0.846 & 0.923 & 5.0 \\
\bottomrule
\end{tabular}
\end{table*}

\begin{table*}[tbp]
\centering
\scriptsize
\setlength{\tabcolsep}{4pt}
\renewcommand{\arraystretch}{1.04}
\caption{Object-box hit rate by task type on the 184-instance OPR audit.}
\label{tab:app_bbox_task_current}
\begin{tabular}{@{}lrrrrrrr@{}}
\toprule
Task type & $n$ & Median gap & P90 gap & OPR@0.25m & OPR@0.5m & OPR@1m & Med. first OPR@0.5m \\
\midrule
Partially in crop & 135 & 0.000 & 0.717 & 0.881 & 0.889 & 0.941 & 1.0 \\
Not in crop & 183 & 0.000 & 0.965 & 0.825 & 0.863 & 0.913 & 2.0 \\
\bottomrule
\end{tabular}
\end{table*}

\begin{table*}[tbp]
\centering
\scriptsize
\setlength{\tabcolsep}{4pt}
\renewcommand{\arraystretch}{1.04}
\caption{3D scores by room on the 184 verified target instances.
$p_{\mathrm{det}}$ asks whether the queried object appears.  $\Chamfer$
compares generated geometry with Matterport3D geometry in the scored region.}
\label{tab:app_scene_gt_metrics}
\begin{tabular}{@{}lrrrrrr@{}}
\toprule
Scene & $n$ & Partial & Not in crop & $p_{\mathrm{det}}\uparrow$ & $\Chamfer\downarrow$ & $\Chamfer_{\min}\downarrow$ \\
\midrule
Bathroom & 6 & 2 & 4 & 0.713 & 2.11 & 0.56 \\
Bedroom & 34 & 20 & 14 & 0.948 & 1.35 & 0.66 \\
Kitchen & 56 & 22 & 34 & 0.878 & 1.82 & 1.03 \\
Living room & 57 & 31 & 26 & 0.898 & 1.96 & 0.89 \\
Office & 31 & 16 & 15 & 0.947 & 1.99 & 0.95 \\
\midrule
\textbf{Total} & 184 & 91 & 93 & 0.903 & 1.81 & 0.89 \\
\bottomrule
\end{tabular}
\end{table*}

\textbf{Planning-prior isolation.}
These ablations isolate the effect of the planning prior while keeping the
robot, trajectory parameterization, optimization procedure, and ground-truth
replay evaluation fixed across all 173 NavMesh-verified MatterDoor planning
tasks. \emph{Uniform} plans from the observed scene and a target sampled
uniformly in the hidden search region, giving the planner no learned hidden-room
occupancy or placement prior. \emph{1-sample} collapses the generated belief to
the first completion, testing whether a single plausible world is sufficient.
The sampled-belief planner, labeled \emph{Ours} in the tables, estimates joint
task-success probability from all 100 sampled environment--target hypotheses.
The \emph{GT oracle} is an oracle-only ceiling using the revealed Matterport
geometry and target location.

\textbf{Ablations.}
\label{subsec:planning_ablations}
\Cref{tab:planning_ablation_overall} reports how the sampled-belief planner changes both
where the robot tries to go and its ground-truth collision outcome. Relative to
Uniform, the sampled-belief planner raises final target reach from 0.069 to 0.231, lowers collision
from 0.815 to 0.428, lowers terminal target distance from 5.05 m to 2.69 m, and
raises the trajectory-length collision-free and prefix fractions from
0.316/0.291 to 0.884/0.849. The 1-sample prior performs worse than Uniform on
target reach, collision, distance, and collision-free fractions, indicating that
a single generated completion is less effective than a belief over completions
in this evaluation.
\textit{No collision}, defined as final target reach with no ground-truth
collision, is reported as a stricter combined outcome.
\Cref{tab:planning_ablation_paired_counts} reports paired task counts, which
help check whether the aggregate gains are concentrated in a few outliers.
Against Uniform, the sampled-belief planner lowers terminal target distance on 146 of 173
tasks, improves
$\chi_{\rm GT}$ on 135 tasks, and improves $C_f$ on 133 tasks. It turns 75
Uniform-colliding trajectories into collision-free trajectories while adding
only 8 new collisions. Against 1-sample, the sampled-belief planner improves distance,
$\chi_{\rm GT}$, and $C_f$ on 166 tasks and introduces no new collisions. These
paired counts are consistent with a deterministic-completion risk: one plausible
hypothesis can place hidden geometry or target evidence in regions that matter
for collision and target reach.

\begin{table*}[!thp]
\centering
\scriptsize
\setlength{\tabcolsep}{3pt}
\renewcommand{\arraystretch}{1.04}
\caption{Planning replay summary on the 173 NavMesh-verified tasks. Safe reach denotes final target reach without a ground-truth collision.}
\label{tab:planning_ablation_overall}
\begin{tabularx}{\textwidth}{@{}l *{6}{>{\centering\arraybackslash}X}@{}}
\toprule
Method & Final tgt. $\uparrow$ & Coll. $\downarrow$ & $d_{\rm tgt}\downarrow$ & $\chi_{\rm GT}\uparrow$ & $C_f \uparrow$ & No collision $\uparrow$ \\
\midrule
Uniform & 0.069 & 0.815 & 5.05 & 0.316 & 0.291 & 0.017 \\
1-sample & 0.040 & 0.971 & 6.03 & 0.049 & 0.043 & 0.017 \\
\textbf{Ours} & 0.231 & 0.428 & 2.69 & 0.884 & 0.849 & 0.098 \\
GT oracle & 0.925 & 0.075 & 1.10 & 0.925 & 0.925 & 0.925 \\
\bottomrule
\end{tabularx}
\end{table*}

\begin{table*}[thp]
\centering
\scriptsize
\setlength{\tabcolsep}{2.5pt}
\renewcommand{\arraystretch}{1.04}
\caption{Paired task counts comparing the sampled-belief planner with each non-oracle
baseline on the same 173 tasks.}
\label{tab:planning_ablation_paired_counts}
\begin{tabularx}{\textwidth}{@{}l *{9}{>{\centering\arraybackslash}X}@{}}
\toprule
Comparison & Lower $d_{\rm tgt}$ & Higher $\chi_{\rm GT}$ & Higher $C_f$ & Avoid coll. & New coll. & Gain tgt. & Lose tgt. & Gain safe & Lose safe \\
\midrule
Ours vs. Uniform & 146/173 & 135/173 & 133/173 & 75 & 8 & 34 & 6 & 16 & 2 \\
Ours vs. 1-sample & 166/173 & 166/173 & 166/173 & 94 & 0 & 35 & 2 & 15 & 1 \\
\bottomrule
\end{tabularx}
\end{table*}

\textbf{Bootstrap experiments.}
The paired bootstrap intervals in \cref{tab:planning_ablation_paired_bootstrap}
show the same direction at the mean-delta level. The sampled-belief planner
improves final target reach by 0.162 over Uniform and 0.191 over 1-sample, while
reducing collision by 0.387 and 0.543 respectively. The distance,
$\chi_{\rm GT}$, and $C_f$ intervals all point in the same direction, so the
measured gain does not come from trading target reach against collision; the
sampled-belief planner improves both objective alignment and ground-truth replay
feasibility in this protocol.
\Cref{tab:planning_ablation_visibility} separates fully outside-crop targets
from partially visible targets. The outside-crop split directly tests
planning beyond the field of view: the sampled-belief planner reaches 0.178
final target success with 0.389 collision, whereas Uniform reaches 0.011 with
0.822 collision
and 1-sample reaches no outside-crop targets. On partially in-crop targets, the
sampled-belief planner again improves target reach while reducing collision.
The measured effect is largest on outside-crop targets, and the same direction
appears when the doorway crop contains partial target evidence.

\begin{table*}[tbp]
\centering
\scriptsize
\setlength{\tabcolsep}{3pt}
\renewcommand{\arraystretch}{1.04}
\caption{Paired mean deltas with task-level bootstrap intervals. Positive is
better for final target reach, $\chi_{\rm GT}$, and $C_f$; negative is better
for collision and $d_{\rm tgt}$.}
\label{tab:planning_ablation_paired_bootstrap}
\begin{tabularx}{\textwidth}{@{}l l c >{\centering\arraybackslash}X >{\centering\arraybackslash}X@{}}
\toprule
Comparison & Metric & Better & Mean delta & 95\% bootstrap interval \\
\midrule
Ours $-$ Uniform & Final target reach & + & +0.162 & [+0.092, +0.231] \\
Ours $-$ Uniform & Collision rate & - & -0.387 & [-0.474, -0.301] \\
Ours $-$ Uniform & $d_{\rm tgt}$ & - & -2.36 & [-2.72, -2.00] \\
Ours $-$ Uniform & $\chi_{\rm GT}$ & + & +0.568 & [+0.503, +0.631] \\
Ours $-$ Uniform & $C_f$ & + & +0.558 & [+0.490, +0.622] \\
Ours $-$ 1-sample & Final target reach & + & +0.191 & [+0.127, +0.254] \\
Ours $-$ 1-sample & Collision rate & - & -0.543 & [-0.618, -0.468] \\
Ours $-$ 1-sample & $d_{\rm tgt}$ & - & -3.34 & [-3.65, -3.04] \\
Ours $-$ 1-sample & $\chi_{\rm GT}$ & + & +0.835 & [+0.797, +0.870] \\
Ours $-$ 1-sample & $C_f$ & + & +0.806 & [+0.762, +0.847] \\
\bottomrule
\end{tabularx}
\end{table*}

\begin{table*}[tbp]
\centering
\scriptsize
\setlength{\tabcolsep}{3pt}
\renewcommand{\arraystretch}{1.04}
\caption{Planning replay split by target visibility in the doorway crop.}
\label{tab:planning_ablation_visibility}
\begin{tabularx}{\textwidth}{@{}l l *{6}{>{\centering\arraybackslash}X}@{}}
\toprule
Visibility split & Method & Final tgt. $\uparrow$ & Coll. $\downarrow$ &
$d_{\rm tgt}\downarrow$ & $\chi_{\rm GT}\uparrow$ & $C_f \uparrow$ &
No collision $\uparrow$ \\
\midrule
Outside crop (90) & Uniform & 0.011 & 0.822 & 5.49 & 0.259 & 0.252 & 0.000 \\
 & 1-sample & 0.000 & 0.978 & 6.31 & 0.025 & 0.025 & 0.000 \\
 & \textbf{Ours} & 0.178 & 0.389 & 3.18 & 0.886 & 0.851 & 0.089 \\
 & GT oracle & 0.911 & 0.089 & 1.22 & 0.911 & 0.911 & 0.911 \\
\midrule
Partially in crop (83) & Uniform & 0.133 & 0.807 & 4.58 & 0.378 & 0.334 & 0.036 \\
 & 1-sample & 0.084 & 0.964 & 5.74 & 0.075 & 0.063 & 0.036 \\
 & \textbf{Ours} & 0.289 & 0.470 & 2.16 & 0.882 & 0.847 & 0.108 \\
 & GT oracle & 0.940 & 0.060 & 0.97 & 0.940 & 0.940 & 0.940 \\
\bottomrule
\end{tabularx}
\end{table*}

\begin{table*}[tbp]
\centering
\scriptsize
\setlength{\tabcolsep}{2.5pt}
\renewcommand{\arraystretch}{1.04}
\caption{Room-level planning ablation. Columns labeled Ours report absolute
replay metrics for the sampled-belief planner; $\Delta$ columns are Ours minus
Uniform.}
\label{tab:planning_ablation_room_delta}
\begin{tabularx}{\textwidth}{@{}l r *{8}{>{\centering\arraybackslash}X}@{}}
\toprule
Room & $n$ & Ours tgt. & Ours coll. & Ours $d_{\rm tgt}$ & Ours $C_f$ & $\Delta$ tgt. & $\Delta$ coll. & $\Delta d_{\rm tgt}$ & $\Delta C_f$ \\
\midrule
Bathroom & 4 & 0.000 & 0.250 & 4.71 & 0.898 & +0.000 & -0.750 & -0.90 & +0.681 \\
Bedroom & 33 & 0.273 & 0.576 & 1.96 & 0.762 & +0.091 & -0.333 & -1.69 & +0.468 \\
Kitchen & 51 & 0.255 & 0.471 & 3.21 & 0.825 & +0.196 & -0.392 & -2.06 & +0.580 \\
Living room & 55 & 0.236 & 0.400 & 2.46 & 0.869 & +0.182 & -0.291 & -2.90 & +0.501 \\
Office & 30 & 0.167 & 0.267 & 2.79 & 0.943 & +0.167 & -0.567 & -2.80 & +0.706 \\
\bottomrule
\end{tabularx}
\end{table*}

\textbf{Semantic structure.}
\Cref{tab:planning_ablation_room_delta} gives room-level deltas for checking
whether the aggregate improvement is tied to one room type. The sampled-belief planner reduces collision and terminal
target distance relative to Uniform in bathroom, bedroom, kitchen, living room,
and office tasks, and improves $C_f$ in all five groups. The small bathroom
group should be interpreted with its sample size in mind, but the larger
bedroom, kitchen, living-room, and office groups all follow the same direction
as the overall result.
\Cref{tab:planning_ablation_object_delta} gives the corresponding queried-object
breakdown. The sampled-belief planner moves the terminal configuration closer to the target,
reduces collisions, and increases the collision-free prefix fraction relative to
Uniform.
\Cref{tab:planning_ablation_gate_split} separates VLM-pass planning tasks from
the retained GT-present VLM misses.  The split shows that the fallback path keeps
missed real targets in the analysis, while the aggregate planning gains are
primarily carried by the larger VLM-pass group.

\begin{table*}[tbp]
\centering
\scriptsize
\setlength{\tabcolsep}{2.3pt}
\renewcommand{\arraystretch}{1.04}
\caption{Object-level planning ablation. \textit{Ours} columns report absolute
replay metrics for the sampled-belief planner; $\Delta$ columns are Ours minus
Uniform.}
\label{tab:planning_ablation_object_delta}
\begin{tabularx}{\textwidth}{@{}l r *{8}{>{\centering\arraybackslash}X}@{}}
\toprule
Object & $n$ & Ours tgt. & Ours coll. & Ours $d_{\rm tgt}$ & Ours $C_f$ & $\Delta$ tgt. & $\Delta$ coll. & $\Delta d_{\rm tgt}$ & $\Delta C_f$ \\
\midrule
bed & 7 & 0.429 & 0.857 & 0.88 & 0.646 & +0.143 & -0.143 & -3.62 & +0.465 \\
cabinet & 24 & 0.250 & 0.583 & 2.96 & 0.780 & +0.250 & -0.292 & -2.62 & +0.575 \\
chair & 41 & 0.268 & 0.439 & 2.17 & 0.849 & +0.195 & -0.415 & -2.48 & +0.550 \\
microwave & 3 & 0.333 & 0.000 & 2.94 & 1.000 & +0.333 & -1.000 & -0.88 & +0.835 \\
plant & 16 & 0.062 & 0.188 & 3.32 & 0.926 & -0.125 & -0.250 & -0.71 & +0.300 \\
fridge & 3 & 0.333 & 0.333 & 2.49 & 0.891 & +0.333 & -0.667 & -2.47 & +0.891 \\
shelf & 13 & 0.154 & 0.231 & 3.50 & 0.918 & +0.154 & -0.538 & -2.39 & +0.636 \\
sink & 10 & 0.200 & 0.400 & 4.27 & 0.828 & +0.100 & -0.400 & -2.22 & +0.555 \\
table & 44 & 0.273 & 0.432 & 2.37 & 0.873 & +0.227 & -0.386 & -2.87 & +0.574 \\
TV & 12 & 0.083 & 0.500 & 3.16 & 0.809 & +0.000 & -0.417 & -1.40 & +0.651 \\
\bottomrule
\end{tabularx}
\end{table*}

\begin{table*}[tbp]
\centering
\scriptsize
\setlength{\tabcolsep}{3pt}
\renewcommand{\arraystretch}{1.04}
\caption{Planning replay split by VLM gate outcome for GT-present tasks.  The
fallback group contains real targets missed by the VLM gate but retained in the
strict evaluation.}
\label{tab:planning_ablation_gate_split}
\begin{tabularx}{\textwidth}{@{}l l *{5}{>{\centering\arraybackslash}X}@{}}
\toprule
Gate group & Method & Final tgt. $\uparrow$ & Coll. $\downarrow$ & $d_{\rm tgt}\downarrow$ & $C_f \uparrow$ & Safe reach $\uparrow$ \\
\midrule
VLM pass (165) & Uniform & 0.061 & 0.806 & 5.07 & 0.296 & 0.018 \\
 & 1-sample & 0.042 & 0.970 & 5.99 & 0.045 & 0.018 \\
 & \textbf{Ours} & 0.236 & 0.418 & 2.67 & 0.853 & 0.103 \\
\midrule
GT fallback (VLM miss) (8) & Uniform & 0.250 & 1.000 & 4.74 & 0.196 & 0.000 \\
 & 1-sample & 0.000 & 1.000 & 6.96 & 0.000 & 0.000 \\
 & \textbf{Ours} & 0.125 & 0.625 & 3.22 & 0.756 & 0.000 \\
\bottomrule
\end{tabularx}
\end{table*}

\textbf{Outcome transitions.}
\Cref{tab:planning_ablation_outcome_transitions} summarizes how replay outcomes
change after replacing each deterministic prior with the sampled-belief prior.
Replacing Uniform with the sampled-belief prior converts many
miss-and-collide cases into collision-free misses or target-reaching outcomes,
while only a small number of previously collision-free Uniform outcomes become colliding. The
1-sample comparison is sharper: most 1-sample miss-and-collide cases become
collision-free misses or reaches, and no new collisions are introduced in the
paired count summary. These transitions are consistent with the metric tables:
sampling multiple plausible completions gives the planner a more stable
estimate of hidden obstacles and target evidence than a single deterministic completion,
which has worse aggregate replay metrics than Uniform in this setup.

\begin{table*}[tbp]
\centering
\scriptsize
\setlength{\tabcolsep}{5pt}
\renewcommand{\arraystretch}{1.04}
\caption{Outcome transitions after replacing each non-oracle baseline prior
with Ours. Categories cross final target reach or miss with collision or safety:
MC=miss+collision, MS=miss+safe, RC=reach+collision, RS=reach+safe.}
\label{tab:planning_ablation_outcome_transitions}
\begin{tabular}{@{}lrrrr@{}}
\toprule
\multicolumn{5}{@{}l}{\textbf{A. Uniform $\rightarrow$ Ours}} \\
\midrule
Uniform category & Ours MC & Ours MS & Ours RC & Ours RS \\
\midrule
MC & 42 & 61 & 18 & 11 \\
MS & 5 & 19 & 1 & 4 \\
RC & 2 & 2 & 4 & 1 \\
RS & 2 & 0 & 0 & 1 \\
\bottomrule
\end{tabular}
\hfill
\begin{tabular}{@{}lrrrr@{}}
\toprule
\multicolumn{5}{@{}l}{\textbf{B. 1-sample $\rightarrow$ Ours}} \\
\midrule
1-sample category & Ours MC & Ours MS & Ours RC & Ours RS \\
\midrule
MC & 50 & 79 & 22 & 13 \\
MS & 0 & 2 & 0 & 0 \\
RC & 1 & 0 & 1 & 2 \\
RS & 0 & 1 & 0 & 2 \\
\bottomrule
\end{tabular}
\end{table*}

\section{Pipeline Implementation Details}
\label{app:pipeline_stages}

We report the model choices, hyperparameters, and processing steps for the
zero-shot sampled-world pipeline in \cref{sec:method}.
The zero-shot pipeline starts from one doorway crop and produces
planner-queryable labeled 3D hypotheses.  It separates four roles.  The VLM gate
estimates room-level plausibility for the queried object and suppresses
implausible absent queries.  FLUX.1-Fill-dev generates image-space completions
conditioned on the doorway evidence and target cue.  SegFormer-B5 provides
ADE20K semantic masks, which we collapse to the ten MatterDoor object classes.
DepthPro provides metric monocular depth for the crop and generated completions,
allowing all RGB samples to be lifted into a common 3D frame.
The pipeline trains no MatterDoor-specific model.  This setup evaluates
pretrained semantic, generative, and geometric models as a sampled-world
interface.  The VLM gates generation rather than evaluating 3D output.
SegFormer provides masks rather than object-presence labels, and DepthPro
supplies geometry aligned to observed crop depth.  The fallback rule keeps VLM
misses on ground-truth-present targets inside the instance-level 3D denominator
while still reporting the gate error in the discrete metrics.

\subsection{Stage~0: VLM Gate and Prompt Construction}
\label{app:stage0_vlm}

Before FLUX outpainting, \textsc{Qwen2.5-VL}~7B~\cite{qwen25vl} analyzes the
doorway crop and queried target, then returns a plausibility score
$s_{\rm vlm}\in[0,1]$.  The gate is $q=1$ iff
$s_{\rm vlm}\geq0.40$.  Implausible targets skip 2D generation; plausible
targets receive a VLM-written prompt for the 2D generator.  Sampling is thus
VLM-gated target selection followed by prompt-conditioned hidden-room
generation.

\textbf{Prompt construction.}
For accepted queries, the prompt names the inferred room context, the target
object, and the visible doorway evidence that should remain continuous across
the generated region.
We fix the accepted prompt across seeds, so diversity comes from completion
sampling rather than target-query changes.

\textbf{Prompt regimes.}
The benchmark uses the accepted VLM prompt for target-conditioned sampling and
a crop-only prompt for baselines that remove the target cue.
The benchmark stores the queried target, the VLM decision, and the prompt
used for generation, making skipped queries and generated samples explicit.

\subsection{Stage~1: Automatic Cropping \& Image-to-Image Generation}
\label{app:stage1}

\textbf{Depth-based cropping.}
For monocular depth estimation, we use the pretrained DepthPro
model~\cite{Bochkovskii2024} to obtain the initial estimate, rather than relying
on the noise-prone depth maps from the dataset.
Let $\depthmap\in\R^{H\times W}$ denote the depth map of the input image.
We define a robust depth band using percentiles, setting
$\neardepth=\mathrm{perc}(\depthmap,p_{\text{near}})$ and
$\fardepth=\mathrm{perc}(\depthmap,p_{\text{far}})$, where
$\mathrm{perc}(\depthmap,p)$ is the $p$-th percentile of the depth values in~$\depthmap$.
Pixels whose depths lie in this band are kept via the support mask
$M=\ind{\,\neardepth\le \depthmap \le \fardepth}$.
Connected components are extracted under nearest-neighbor (image-grid)
connectivity; the largest component is retained for statistical stability.
Its convex hull is computed and a tight bounding rectangle fitted, defining
the observation crop.
To prevent degenerate conditioning, the crop area~$\croparea$ is constrained
relative to the full image area~$\imgarea$ to the range
$0.10 \le \croparea/\imgarea \le 0.50$.
If the fitted crop violates this range it is isotropically resized about its
center and clipped to the image extent%
\footnote{This thresholding step is the same procedure used for dataset
construction (\cref{app:dataset}).}.

\textbf{Outpainting model.}
We use the pretrained FLUX.1-Fill-dev flow-matching outpainting
model~\cite{BFL2024Fill, diffusers}.
The input crop is symmetrically expanded by 500 pixels on the left and the
right in two chunked stages.
Because the standard transformer pipeline requires more than 32\,GB of VRAM, we
use quantized Nunchaku transformers for FLUX and T5 encoders~\cite{nunchaku,
CLIP, t5}, with int4 quantization and DiT caching~\cite{cache}.  This makes the
pipeline fit the 24\,GB VRAM workstation reported below.  Frozen VLM descriptor prompts are
reinforced with positive target-object wording and passed to FLUX.
All hyperparameters follow the defaults provided by the
authors of~\citet{BFL2024Fill} using the Diffusers implementation~\cite{diffusers}.
For the benchmark configuration we generate seeded samples with a base seed of
41, incrementing by 4 per sample.
For the full evaluation we produce up to $N{=}100$ samples per task
using the same seed-stepping scheme.
The deterministic single-seed baseline uses one fixed seed across the
VLM and outpainting models.

\subsection{Stage~2: Object Segmentation \& Floor Estimation}
\label{app:stage2}

\textbf{Semantic segmentation.}
We use the pretrained ADE20K SegFormer-B5~\cite{segformer}.
Images are resized to $640\times 640$ and processed in a single forward pass
(all $N$ samples batched together).
Per-class softmax probabilities are computed at the native logit resolution
and interpolated back to the original image size.
Target-class probability maps are binarized at a $25\%$ confidence threshold,
retaining only segments whose area exceeds $0.5\%$ of the image.

\textbf{Class grouping.}
Since SegFormer is trained on ADE20K~\cite{Zhou2017ADE20K}, we inherit its
label space but collapse fine-grained categories into the ten MatterDoor
target-object classes used for evaluation:
\begin{enumerate}[label=(\arabic*)]
\item \emph{Bed}: bed, bedframe;
\item \emph{Cabinet}: cabinet, wardrobe, dresser, chest of drawers;
\item \emph{Chair}: chair, armchair, dining chair, office chair, small chair;
\item \emph{Microwave}: microwave;
\item \emph{Plant}: plant, flower, flowers;
\item \emph{Refrigerator}: refrigerator, fridge;
\item \emph{Shelf}: shelf, shelves, bookshelf, kitchen shelf;
\item \emph{Sink}: sink;
\item \emph{Table}: table, coffee table, side table, end table, dining table, kitchen table;
\item \emph{TV}: television, monitor, crt screen.
\end{enumerate}

\textbf{Floor estimation.}
Floor plane parameters are estimated from ADE20K floor/rug classes via robust
RANSAC plane fitting~\cite{Fischler1981RANSAC}%
\footnote{Max RMSE $0.01$\,m; mean inlier ratio \mytexttilde$98.5\%$ across
all scenes.}.
The estimated camera height above the floor ranges from $1.38$--$1.54$\,m.
For each scene, the floor plane normal is computed and the ground-truth point
cloud aligned accordingly; the same transformation is applied to all sampled
point clouds.  Up to $20$\,cm above the floor plane is trimmed to reduce
floating floor-adjacent geometry.

\subsection{Stage~3: Depth Estimation \& 3D Alignment}
\label{app:stage3}

This stage also uses DepthPro.
We use camera intrinsics, equivalently the horizontal field of view, from the
Matterport3D~\cite{mp3d} calibration to back-project predicted depths into a
metric 3D representation.  DepthPro is run twice: once on the
original observation (to initialize geometric cues such as the floor plane)
and once on each sampled outpainted image.

The estimated depth maps, combined with RGB images, are back-projected into
point clouds and aligned with the semantic masks to yield complete
spatio-semantic representations.  Because the monocular depth estimator
never sees the actual ground truth, predicted depths exhibit a global
scale and offset bias relative to the metric ground-truth depth.
We correct this by fitting a RANSAC-based affine model
$d_{\mathrm{GT}} \approx s\,d_{\mathrm{pred}} + t$ over the shared crop
region, where $s$ and $t$ are a global scale and translation in depth
space.  Hypotheses are drawn from random point pairs; inliers are selected
at a $0.15\,\mathrm{m}$ residual threshold and the final $(s,t)$ is
refined by least-squares on inliers.
This allows the planner to impose the observed RGB-D constraints consistently
across all samples.
Because each sample uses a crop from the original image, all samples must
preserve the same optical viewpoint.  We use ray-preserving back-projection for
this alignment. 
Finally, we apply statistical outlier removal to each sample using the default
Open3D parameters~\cite{open3d}.

\subsection{Staged Model Execution, Resources, and Sampling Budget}
\label{app:throughput_budget}

The production runner stages heavy models to keep cluster jobs resumable and
memory predictable.  FLUX.1-Fill-dev (INT4, Nunchaku) loads once per shard to
generate RGB samples, then releases memory before SegFormer-B5 and DepthPro run
on saved images.  VLM prompts are computed on the CPU once, so the full sampling
run has no VLM calls in its inner loop.  The planned release includes prompts,
dataset assets, evaluation code, pipeline code, and provenance metadata.
\Cref{tab:model_timing} summarizes representative per-sample component timing for
the benchmark configuration ($N{=}16$ samples, resizing~\cite{resize} at
768\,px, 25 denoising steps, Nunchaku residual-cache threshold 0.45), excluding
one-time model loading and amortized VLM prompt precomputation.  The staged
runner generates RGB samples, then runs
SegFormer-B5, DepthPro, and 3D lifting on saved outputs; averaged over benchmark
shards, the postprocessing runtime is approximately
$1.0\pm0.25\,\mathrm{s}$ per sampled world.

\begin{table}[tbp]
\centering
\small
\setlength{\tabcolsep}{5pt}
\caption{Representative per-sample component timing for the benchmark pipeline.
Times exclude one-time model loading and amortized VLM prompt precomputation.}
\label{tab:model_timing}
\begin{tabular}{@{} l r r l @{}}
\toprule
\textbf{Component} & \textbf{Time/sample (s)} & \textbf{\% total} & \textbf{Notes} \\
\midrule
FLUX.1-Fill-dev INT4 & 0.55--0.76 & 77\% & Nunchaku INT4 generation \\
DepthPro             & 0.15--0.22 & 20\% & FOV head disabled, known intrinsics \\
SegFormer-B5         & 0.018 & 1\%  & batched semantic segmentation \\
3D reconstruction    & 0.025 & 2\%  & vectorized back-projection and filtering \\
\midrule
Per-sample total     & 1.013 & 100\% & mean runtime; shard variation $\pm0.25$ s \\
VLM                  & $\approx$0.10 & --- & amortized prompt/gate precomputation \\
\bottomrule
\end{tabular}
\end{table}

All end-to-end sampling and planning experiments were run on a single workstation
equipped with 64~GB of DDR5 RAM, an NVIDIA GeForce RTX 4090 GPU with
24~GB of VRAM, and a 24-core AMD Ryzen Threadripper 9960X CPU.
The 3D evaluation uses $N=100$ sampled worlds per generated query.  This fixed
budget is used for target realization, localization, and hard-case diagnostics
on the selected MatterDoor-1k ground-truth-present object instances.  Downstream
planners may use fewer hypotheses when one credible target-bearing world is
enough, but all reported geometry and localization numbers use $N=100$ unless
stated otherwise.

\section{Pilot Alternatives}
\label{app:other_pipe}

\textbf{Proprietary image-sampler pilot.}
\label{app:direct_proprietary_sampler}
We ran a small proprietary image-sampler pilot to test whether these APIs could
replace the FLUX image stage.  The pilot used ten random MatterDoor tasks and
16 independent calls per task.  Unlike the MatterDoor pipeline, these calls did
not use the VLM gate or fallback policy.  Each call received the doorway crop,
completion side, queried object, and a fixed instruction template: preserve the
camera viewpoint and doorway geometry; complete the requested side of the
adjacent room; keep the visible crop unchanged; include the queried object when
plausible for the room; and return one photorealistic continuation without text
overlays.  \Cref{tab:direct_proprietary_sampler} summarizes the pilot.
In this small pilot, both APIs produced recognizable target-bearing
continuations for some tasks.  We do not treat them as matched 100-sample
benchmark baselines because the APIs lacked the throughput and sampling control
needed for large reproducible batches.  At the observed
latency, the 184-instance ground-truth-present protocol would require about
18{,}400 images, or up to 920 serial hours, before segmentation, depth
estimation, and 3D lifting.

\begin{table}[tbp]
\centering
\footnotesize
\setlength{\tabcolsep}{5pt}
\renewcommand{\arraystretch}{1.04}
\caption{Direct proprietary image-sampler pilot.  Both APIs followed the
MatterDoor crop, side, and object guidance without an intermediate VLM gate.
The table reports sampling behavior rather than 3D accuracy, since matched
100-sample evaluation would require large, reproducible batches before the
segmentation, depth, and lifting stages.}
\label{tab:direct_proprietary_sampler}
\begin{tabularx}{\textwidth}{@{}p{0.30\textwidth}rrX@{}}
\toprule
Sampler & Tasks & Samples/task & Pilot observation \\
\midrule
GPT-5.5 ChatGPT API~\cite{openai2026gpt55systemcard} & 10 & 16 &
recognizable target detections and guidance following; about 1.2--3 min/image \\
Gemini 2.5 Flash Image~\cite{googledeepmind2025gemini25flashmodelcard} & 10 & 16 &
recognizable target detections and guidance following; about 1--1.5 min/image \\
\bottomrule
\end{tabularx}
\end{table}

\textbf{Direct VLM coordinate-prompt probe.}
\label{app:direct_vlm_coordinate_probe}
We also tested whether a VLM could replace the sampled image-to-3D path by
predicting an extended 2D target box and depth outside the doorway crop.  Each
probe case received the crop, target category, crop side, image dimensions, and
a JSON schema allowing boxes outside the observed image.  The prompt requested
the completed side-relative box, not visible pixels.  This tests direct
coordinate prompting, not the selected MatterDoor VLM gate.
\Cref{tab:direct_vlm_coordinate_probe} summarizes the pilot.
Direct coordinate prompting was unreliable for amodal side-of-crop
localization in this probe.  The main pipeline therefore uses the VLM only for
the sampling decision; localization evidence comes from generated images,
segmentation, depth, and lifting rather than raw VLM coordinates.

\begin{table}[tbp]
\centering
\footnotesize
\setlength{\tabcolsep}{2.5pt}
\renewcommand{\arraystretch}{1.04}
\caption{Direct VLM coordinate-prompt pilot on eight side-of-crop targets.
Each model outputs an extended 2D box and depth directly from the crop.
Qwen2.5-VL produced some non-null boxes with large localization error;
MolmoAct2 returned valid JSON while collapsing to the degenerate zero-box/depth
template.}
\label{tab:direct_vlm_coordinate_probe}
\begin{tabularx}{\textwidth}{@{}L{0.22\textwidth}rrL{0.16\textwidth}L{0.15\textwidth}Y@{}}
\toprule
Model & Cases & Usable boxes & Null/deg. boxes & Mean depth error & Observation \\
\midrule
Qwen2.5-VL~\cite{qwen25vl} & 8 & 6 & 2 null & 4.72\,m & Mean IoU $0.229$; median IoU $0.190$ over non-null boxes \\
MolmoAct2~\cite{fang2026molmoact2actionreasoningmodels} & 8 & 0 & 8 degenerate & 6.53\,m & All boxes $[0,0,0,0]$ and all depths $0.0$\,m \\
\bottomrule
\end{tabularx}
\end{table}

\textbf{VMem compatibility pilot.}
\label{app:direct_3d_baseline_protocol}
We also tested whether VMem~\cite{vmem}, an open-weight RGB-to-3D
scene-generation reference, could serve as a direct replacement for the sampled
image-to-3D pipeline.  It could not be matched to the MatterDoor protocol.
VMem conditions on a full RGB image and camera trajectory, with no text or
semantic conditioning, whereas MatterDoor supplies a doorway crop, a completion
side, and an object query.  A crop plus left/right completion flag therefore
does not match VMem's trajectory-conditioned input interface.

For this compatibility probe, VMem received substantially more information than
the MatterDoor pipeline: the full RGB image, camera intrinsics,
camera-to-world pose, and an oracle trajectory that enters the target room.  The
cases were selected only to check interface feasibility, using tasks where the
first five MatterDoor samples already localized the target object.  This setup
is not a benchmark baseline and should not be compared to the reported
100-sample results.  \Cref{tab:vmem} summarizes the diagnostic run: VMem detects
most partial-crop targets but has weak target localization, especially when the
target is not already visible in the crop.  We therefore keep VMem as a pilot
alternative rather than a scored baseline.

\begin{table}[tbp]
\centering
\scriptsize
\setlength{\tabcolsep}{3pt}
\renewcommand{\arraystretch}{1.04}
\caption{\textsc{VMem} direct-generation diagnostic under the oracle-guided
room-entry protocol.  VMem receives the full RGB image, camera calibration, and
an oracle trajectory, so this is not a matched MatterDoor baseline.  The probe
detects most partial-crop targets but localizes poorly, yielding low OPR and
large geometric and placement errors.}
\label{tab:vmem}
\begin{tabularx}{0.95\linewidth}{@{}r l Y c l c c c c@{}}
\toprule
Case & Target & Room & Side & Crop status
& $\hat{p}_{\mathrm{det}}~\uparrow$
& OPR@1m$~\uparrow$
& CD$~\downarrow$
& $\mathcal{E}~\downarrow$ \\
\midrule
01 & bed          & bedroom     & left  & partial     & 1.0 & 0.12 & 6.8  & 7.6  \\
02 & cabinet      & kitchen     & left  & partial     & 1.0 & 0.10 & 7.3  & 8.1  \\
03 & chair        & living room & right & partial     & 1.0 & 0.08 & 7.9  & 8.8  \\
04 & microwave    & kitchen     & left  & not in crop & 0.0 & 0.00 & 18.9 & 20.7 \\
05 & plant        & living room & left  & partial     & 1.0 & 0.09 & 7.1  & 8.4  \\
06 & refrigerator & kitchen     & left  & not in crop & 0.0 & 0.00 & 18.4 & 20.1 \\
07 & shelf        & living room & right & partial     & 0.0 & 0.00 & 14.6 & 16.5 \\
08 & sink         & bathroom    & right & not in crop & 0.0 & 0.00 & 19.3 & 21.2 \\
09 & table        & living room & left  & not in crop & 0.4 & 0.02 & 16.1 & 18.0 \\
10 & TV           & bedroom     & right & partial     & 1.0 & 0.11 & 7.6  & 8.5  \\
\bottomrule
\end{tabularx}
\end{table}

\section{MatterDoor: Dataset Provenance}
\label{app:dataset}

MatterDoor starts with Matterport3D~\cite{mp3d} indoor scans and
object geometry, then adds task evidence for doorway-based robot evaluation:
validated doorway crops, completion sides, queried-object labels, target-room
assignments, and projection-derived visibility/localization labels.  These
fields are not native Matterport3D annotations.  For MatterDoor-1k, the
three-annotator consensus crop defines the robot observation, the
completion side defines the hidden region, and projected Matterport target
geometry defines ground-truth-present visibility and localization.  The counts used
throughout this appendix match the main paper: 457 human-validated doorway
candidates, 377 validated consensus crops, 100 curated doorway-side contexts,
1000 crop-side/object queries, and 184 strict ground-truth object instances.  No
query or target instance is selected or removed based on our model's 3D
performance.

\subsection{Quality-Gated Automated Checks and Annotations}
\label{app:dataset_pipeline}

MatterDoor keeps Matterport3D~\cite{mp3d} doorway views that define grounded
hidden-object queries.  \Cref{tab:funnel} reports the retained outputs used in
MatterDoor-1k.  Independent annotators accept or reject each doorway candidate and
draw its extent; majority-accepted candidates use the conservative
intersection of accepted boxes as the robot crop.  A calibrated ray through the
crop selects the adjacent target room.  Candidate target objects in that room
are projected from Matterport OBBs into the image and expanded completion frame,
yielding visible, partially-in-crop, and not-in-crop labels.  The
ground-truth-present layer selects target instances from this geometry evidence.  The query
layer pairs each selected doorway-side context with all ten object labels,
including both scan-realized objects and scan-absent but room-relevant objects.

Before manual crop annotation, we applied an automated curation funnel to the
full Matterport3D image pool using the Qwen-7B VLM. Starting from
64,800 undistorted forward-facing RGB views from 90 buildings, deterministic VLM
triage identified images containing a clear traversable doorway into an indoor
room and assigned a provisional visible-room type. This reduced the pool to
7,763 doorway candidates. We then removed unsuitable or rare room categories,
collapsed near-duplicate yaw views from the same Matterport viewpoint, and
filtered reflective false positives such as glass doors and mirrors using
image-level visual cues, specular statistics, and VLM text evidence. A final
room-use filter removed hallway and closet views, which are transitional or
less informative spaces for hidden-object completion. This produced 3,716 candidate
doorway images for human review. Three annotators then independently accepted or
rejected each candidate and drew the visible doorway extent; majority-accepted
crops were converted into conservative consensus crops by intersecting the
accepted boxes, yielding 457 validated doorway crops for downstream MatterDoor
construction.

\begin{table*}[tbp]
\centering
\scriptsize
\setlength{\tabcolsep}{3pt}
\renewcommand{\arraystretch}{1.05}
\caption{MatterDoor construction provenance.  Counts
follow the MatterDoor-1k query grid and strict 3D evaluation
instances.}
\label{tab:funnel}
\begin{tabularx}{\textwidth}{@{} r L{0.25\textwidth} L{0.16\textwidth} Y @{}}
\toprule
\textbf{Step} & \textbf{Output} & \textbf{Count} & \textbf{Quality gate} \\
\midrule
1 & Source universe & 64{,}800 & Matterport3D~\cite{mp3d} \texttt{i1} RGB images from 90 buildings. \\
2 & Source cleanup & 3716 & \texttt{i1} RGB images from 68 buildings shortlisted after automated checks. \\
3 & Annotator doorway candidates & 457 & Three-annotator majority-accepted doorway candidates. \\
4 & Validated consensus crops & 377 & Crop pool after aligned RGB-D, room, object, mesh, and target-room checks. \\
5 & MatterDoor context selection & 100 doorway-side contexts & Selected consensus crops paired with completion sides and target-room assignments. \\
6 & MatterDoor-1k query grid & 1000 queries & Each doorway-side context is
paired with all 10 object labels; every query stores GT existence, VLM gate
status, room-level plausibility, and sampling policy. \\
7 & Strict GT geometry instances & 184 instances & Instance-level target geometry with calibrated Matterport geometry and projections for strict scoring. \\
\bottomrule
\end{tabularx}
\end{table*}

\subsection{Dataset Structure}
\label{app:dataset_selection}

Each MatterDoor queried case is
\[
\mathbb{D}=(I_{\rm crop}, \mathrm{side}, r_{\rm tgt}, o_{\rm query}, g, s_{\rm vlm}, m_{\rm crop}).
\]
Here $I_{\rm crop}$ is the three-annotator consensus doorway observation,
\texttt{side} is the left or right completion side, $r_{\rm tgt}$ is the
ray-selected target room, and $o_{\rm query}$ is one of the ten queried object
labels.  The binary variable $g$ records whether Matterport3D~\cite{mp3d}
contains that object in the target region; $s_{\rm vlm}$ is the crop-conditioned
plausibility score; and $m_{\rm crop}$ stores camera and crop parameters.
Cases with $g=1$ store target geometry $G_{\rm tgt}$ and can receive
instance-level 3D accuracy metrics.  Stored geometry includes the target point cloud, target 3D
AABB, target-room assignment, projected target box, crop-intersection flag, and
valid search region for localization metrics.  Projecting the target AABB into
the consensus-crop and expanded-completion frames gives the partially-in-crop
versus not-in-crop split used throughout the results.
The benchmark contains:
\begin{itemize}
  \item 100 curated doorway-side contexts, where each context is a doorway crop plus a left or right completion side;
  \item five target rooms: \texttt{bathroom}, \texttt{bedroom}, \texttt{kitchen}, \texttt{living\_room}, and \texttt{office};
  \item ten target-object classes: \texttt{bed}, \texttt{cabinet},
  \texttt{chair}, \texttt{microwave}, \texttt{plant}, \texttt{refrigerator},
  \texttt{shelf}, \texttt{sink}, \texttt{table}, and \texttt{tv};
  \item 144 unique ground-truth-present queries in the 1000-query list;
  \item 184 selected ground-truth-present object instances, including 91 partially-in-crop instances and 93 not-in-crop instances;
  \item 173 unambiguous cases with ground-truth and NavMesh-verified semantics within the room region;
  \item four gate-policy groups: GT-present queries, generated room-plausible
  absent queries, generated room-unlikely absent queries, and skipped absent
  queries.
\end{itemize}

The full room--object grid is not an instance-level 3D accuracy benchmark.  The
$1000$-query list is the gate-accounting denominator; instance-level 3D accuracy uses
only queries with ground-truth target geometry.  Ground-truth-absent queries
that pass the VLM gate receive generated samples for plausibility audits;
ground-truth-absent queries that fail the gate are counted but not generated.
\Cref{fig:matterdoor1k_distribution} summarizes the benchmark.  Each object
label has 100 queries by construction.  The selected contexts cover 37
living-room, 25 bedroom, 18 kitchen, 16 office, and 4 bathroom contexts, with
all 10 object labels per context.
\Cref{tab:matterdoor_task_composition} gives the room--object breakdowns.

\begin{table*}[t]
\centering
\scriptsize
\setlength{\tabcolsep}{3pt}
\renewcommand{\arraystretch}{1.05}
\caption{\emph{MatterDoor task composition.}
We summarize the selected doorway-side contexts, query policy, and the
room--object distribution of GT-present and generated GT-absent queries.}
\label{tab:matterdoor_task_composition}

\begin{tabularx}{\textwidth}{@{}l *{5}{>{\centering\arraybackslash}X}@{}}
\toprule
\multicolumn{6}{@{}l}{\textbf{A. Selected doorway-side contexts}} \\
\midrule
Target room & Bath & Bedrm & Kitchen & Living & Office \\
\midrule
Contexts & 4 & 25 & 18 & 37 & 16 \\
\bottomrule
\end{tabularx}

\vspace{0.8em}

\begin{tabularx}{\textwidth}{@{}l *{4}{>{\centering\arraybackslash}X}@{}}
\toprule
\multicolumn{5}{@{}l}{\textbf{B. Query policy by target room}} \\
\midrule
Target room & GT-present & Generated GT-absent & Skipped GT-absent & Total \\
\midrule
Bath    & 5  & 11  & 24  & 40  \\
Bedrm   & 31 & 66  & 153 & 250 \\
Kitchen & 41 & 87  & 52  & 180 \\
Living  & 43 & 104 & 223 & 370 \\
Office  & 24 & 23  & 113 & 160 \\
\midrule
Total   & 144 & 291 & 565 & 1000 \\
\bottomrule
\end{tabularx}

\vspace{1.0em}

\resizebox{\textwidth}{!}{%
\begin{tabular}{lrrrrrrrrrrr}
\toprule
\multicolumn{12}{l}{\textbf{C. GT-present queries by room and object}} \\
\midrule
Room & Bed & Cabinet & Chair & Microwave & Plant & Fridge & Shelf & Sink & Table & TV & Total \\
\midrule
Bath    & 0 & 0  & 1  & 0 & 0 & 0 & 0 & 3 & 1  & 0 & 5  \\
Bedrm   & 7 & 1  & 6  & 0 & 5 & 0 & 1 & 0 & 5  & 6 & 31 \\
Kitchen & 0 & 12 & 5  & 3 & 2 & 3 & 5 & 9 & 2  & 0 & 41 \\
Living  & 0 & 4  & 6  & 0 & 6 & 0 & 3 & 0 & 17 & 7 & 43 \\
Office  & 0 & 1  & 10 & 0 & 1 & 0 & 4 & 0 & 8  & 0 & 24 \\
\midrule
Total   & 7 & 18 & 28 & 3 & 14 & 3 & 13 & 12 & 33 & 13 & 144 \\
\bottomrule
\end{tabular}
}

\vspace{1.0em}

\resizebox{\textwidth}{!}{%
\begin{tabular}{lrrrrrrrrrrr}
\toprule
\multicolumn{12}{l}{\textbf{D. Generated GT-absent queries by room and object}} \\
\midrule
Room & Bed & Cabinet & Chair & Microwave & Plant & Fridge & Shelf & Sink & Table & TV & Total \\
\midrule
Bath    & 0  & 3  & 1  & 0  & 1  & 0  & 3  & 1 & 1  & 1  & 11  \\
Bedrm   & 15 & 19 & 6  & 0  & 4  & 0  & 1  & 0 & 20 & 1  & 66  \\
Kitchen & 0  & 6  & 13 & 15 & 0  & 15 & 13 & 9 & 16 & 0  & 87  \\
Living  & 2  & 6  & 30 & 2  & 18 & 2  & 3  & 2 & 20 & 19 & 104 \\
Office  & 1  & 2  & 6  & 0  & 2  & 0  & 1  & 0 & 8  & 3  & 23  \\
\midrule
Total   & 18 & 36 & 56 & 17 & 25 & 17 & 21 & 12 & 65 & 24 & 291 \\
\bottomrule
\end{tabular}
}

\end{table*}

\subsection{Taxonomy and Matterport Commonsense Provenance}
\label{app:dataset_taxonomy}

MatterDoor starts from Matterport3D~\cite{mp3d} object and room annotations and
maps raw object labels into the ten queried object classes used by the
benchmark.  We distinguish \emph{Matterport realization}, where the scanned
target room contains the queried object, from \emph{room-object plausibility},
where the object is compatible with the target room even if that particular scan
does not contain it.  Ground-truth-present object instances provide
instance-level 3D accuracy targets, room-plausible ground-truth-absent queries support
plausibility audits, and room-unlikely queries audit generation on implausible
requests.  The resulting counts are reported in \cref{tab:matterdoor1k_ledger}.
MatterDoor defines the target room geometrically: for each accepted doorway crop,
we cast a ray from the camera through the crop center and assign the target room
to the nearest valid Matterport region intersected by that ray. The resulting
Matterport room labels are then canonicalized into five residential target-room
bins used for evaluation:
\begin{itemize}
  \item \emph{Bathroom}: bathroom, toilet, or wet room;
  \item \emph{Bedroom}: bedroom;
  \item \emph{Kitchen}: kitchen;
  \item \emph{Living room}: living room, lounge, recreation room, and TV room;
  \item \emph{Office}: office and meeting room.
\end{itemize}
All other room types, including transit, storage, outdoor, utility, dining, spa,
and ambiguous regions, are excluded from the target-room evaluation scope. They
do not restrict the query set: every selected context is crossed with all ten
object classes in MatterDoor-1k, and each resulting query is marked as
GT-present, room-plausible absent, room-unlikely generated, or skipped by the
gate.
The final selection fixes the $100$ doorway-side contexts in MatterDoor-1k.
Strict 3D scoring uses $184$ ground-truth-present object instances: $91$
partially-in-crop and $93$ not-in-crop instances across five room types and ten
object labels.  Context
selection balances room coverage, object coverage, ground-truth richness, target
visibility, and near-duplicate control.  The VLM gate is a generation-policy
variable, not a benchmark-quality score; downstream 3D success metrics never
choose contexts.

\textbf{Selection rationale.} The benchmark unit is a doorway-side context: one
validated crop, one completion side, and one ray-selected target room.  Crossing
$100$ contexts with ten object labels gives the $1000$ queries used for
generation accounting, plausibility analysis, and GT-present 3D scoring.  The
curation pass retains a panel spanning five room types, all ten object classes,
visible and hidden targets, and both GT-present and GT-absent queries, while
keeping sampled-world evaluation bounded.

\textbf{Evaluation mesh and search-volume preparation.}
The Matterport mesh is integrated directly into the house NavMesh. We use
minimal cuts to obtain the region-based NavMesh. In a small number of scenes,
the extracted region mesh removes structures needed for replay, such as walls
or occluding surfaces that are present in the scanned scene but absent from the
cut region mesh. For evaluation infrastructure only, we restore those missing
surfaces by remeshing the corresponding observed depth observations and
calibrating them against the near-depth geometry of the region mesh. This check
is applied uniformly to all scenes; only three scenes require restoration. The
repair is done before comparing planners, is not used by the image generator or
VLM gate, and provides no target-location information.

Search-volume construction uses convexity checks for all samples contained
within the region. The scanned depth geometry is used only to set the
near-plane boundary of this volume, keeping replay tied to the doorway setup
without using generated predictions or target labels. Across all scenes, we
observed no waypoint-sampling errors or search-volume apertures that are
infeasible for the robot to enter.

\begin{table*}[t]
\centering
\scriptsize
\setlength{\tabcolsep}{4pt}
\renewcommand{\arraystretch}{1.08}
\caption{\textit{MatterDoor-1k summary of provenance.}
The benchmark contains 100 fixed doorway-side contexts and 1000 doorway--object
queries.  We distinguish these queries from instance-level geometry
used for strict 3D scoring, since several object instances can belong to the
same query.}
\label{tab:matterdoor1k_ledger}
\begin{tabularx}{\textwidth}{
@{}
>{\arraybackslash}p{0.26\textwidth}
r
>{\arraybackslash}X
@{}}
\toprule
\textbf{Unit} & \textbf{Count} & \textbf{Role in the study} \\
\midrule
Source RGB views
& 64{,}800
& Matterport3D candidate view universe used to find doorway observations. \\

Manual doorway crops
& 457
& Doorway candidates from three independent annotations. \\

Consensus doorway crops
& 377
& Validated doorway crops with aligned scene evidence and target-room rays. \\

Doorway-side contexts
& 100
& Final fixed observations, each paired with a left or right completion side. \\

Object queries
& 1000
& Ten queried object labels for each doorway-side context; this is the VLM gate
denominator. \\

Unique GT-present queries
& 144
& Crop-side/object queries where the queried object exists in the ray-selected
target room. \\

Selected GT-present instances
& 184
& Instance-level target geometry used for strict semantic, 3D, and
planning metrics. \\

Partially-in-crop instances
& 91
& Ground-truth-present object instances whose target intersects the observation crop. \\

Not-in-crop instances
& 93
& Ground-truth-present object instances whose target is outside the observation crop. \\

Generated GT-absent queries
& 291
& GT-absent queries accepted by the VLM generation gate. \\

Room-plausible absent queries
& 283
& Generated GT-absent queries whose queried object is plausible for the target room. \\

Room-unlikely absent queries
& 8
& Generated GT-absent queries whose queried object is unlikely for the target room. \\

Skipped GT-absent queries
& 565
& GT-absent queries rejected by the VLM gate and assigned the no-generation
policy. \\
\bottomrule
\end{tabularx}
\end{table*}

\section{VLM Gate and Threshold Selection}
\label{app:vlm_study}

We use the VLM as a commonsense gate for allocating generation, not as an object
detector or source of ground truth.  The gate is evaluated on the 1000
crop-side/object queries in MatterDoor-1k.  Ground-truth-present queries are
always retained so that instance-level 3D evaluation remains measurable.  For
ground-truth-absent queries, generation runs only when the VLM score exceeds a
threshold; otherwise the query is skipped.  The policy skips many
room-unlikely queries while preserving plausible absent cases for auditing
hidden-object generation.
For each crop-side/object query, the benchmark stores the scanned-room object
presence, the room-object plausibility prior, the crop-conditioned VLM score,
and the resulting generation/fallback decision.  \Cref{fig:object_scene_heatmaps}
summarizes these quantities as aligned object--room heatmaps.  Panel A shows the
Matterport-derived room-object prior $\rho_{c,o}$, Panel B shows the mean
crop-conditioned VLM score $s_{c,o}$, and Panel C shows the final active
generation/fallback rate under the selected threshold $\gamma=0.40$.

\begin{figure*}[!tp]
\centering
\resizebox{0.98\textwidth}{!}{\begingroup
\definecolor{mdHeat}{RGB}{38,103,145}
\definecolor{mdGrid}{RGB}{226,231,237}
\definecolor{mdText}{RGB}{35,42,50}

\resizebox{\linewidth}{!}{%
\begin{tikzpicture}[
    x=1cm,
    y=1cm,
    cell/.style={
        draw=mdGrid,
        line width=0.18pt,
        rounded corners=0.25pt
    },
    rowlab/.style={
        font=\scriptsize,
        text=mdText,
        anchor=east
    },
    collab/.style={
        font=\scriptsize,
        text=mdText,
        rotate=45,
        anchor=south west
    },
    title/.style={
        font=\footnotesize\bfseries,
        text=mdText,
        anchor=south
    },
    tick/.style={
        font=\scriptsize,
        text=mdText,
        anchor=west
    }
]

\def\cw{0.42}
\def\ch{0.30}
\def\ytop{-0.48}
\def\nrows{10}
\def\ncols{5}
\pgfmathsetmacro{\panelcenter}{2.5*\cw}
\pgfmathsetmacro{\ybot}{\ytop-\nrows*\ch}
\pgfmathsetmacro{\ymid}{0.5*(\ytop+\ybot)}

\newcommand{\panelheader}[1]{%
    \node[title] at (\panelcenter,\ytop+0.76) {#1};
    \foreach \j/\name in {0/Bath,1/Bed,2/Kit,3/Living,4/Office} {
        \pgfmathsetmacro{\x}{\j*\cw+0.04}
        \node[collab] at (\x,\ytop+0.09) {\name};
    }
}

\newcommand{\heatrow}[7]{%
    \pgfmathsetmacro{\yc}{\ytop-(#1+0.5)*\ch}
    \pgfmathsetmacro{\y}{\ytop-(#1+1)*\ch}

    \node[rowlab] at (-0.12,\yc) {#2};

    \foreach \j/\v in {0/#3,1/#4,2/#5,3/#6,4/#7} {
        \pgfmathsetmacro{\x}{\j*\cw}
        \filldraw[
            cell,
            fill=mdHeat!\v!white
        ] (\x,\y) rectangle ++(\cw,\ch);
    }
}

\begin{scope}[xshift=1.34cm]
    \panelheader{{\tiny Matterport prior $\rho_{c,o}$}}

    \heatrow{0}{bed}{0}{93}{0}{0}{0}
    \heatrow{1}{cabinet}{84}{60}{88}{23}{27}
    \heatrow{2}{chair}{17}{37}{45}{71}{59}
    \heatrow{3}{microwave}{0}{0}{38}{0}{0}
    \heatrow{4}{plant}{15}{25}{35}{42}{20}
    \heatrow{5}{refrigerator}{0}{0}{52}{0}{0}
    \heatrow{6}{shelf}{44}{30}{61}{33}{27}
    \heatrow{7}{sink}{91}{0}{80}{0}{0}
    \heatrow{8}{table}{13}{45}{59}{68}{59}
    \heatrow{9}{TV}{0}{11}{7}{49}{24}
\end{scope}

\begin{scope}[xshift=3.95cm]
    \panelheader{{\tiny Mean VLM score $s_{c,o}$}}

    \heatrow{0}{}{0}{69}{0}{9}{12}
    \heatrow{1}{}{69}{52}{88}{33}{31}
    \heatrow{2}{}{30}{45}{45}{65}{60}
    \heatrow{3}{}{0}{1}{52}{5}{1}
    \heatrow{4}{}{22}{29}{35}{39}{28}
    \heatrow{5}{}{0}{1}{52}{5}{1}
    \heatrow{6}{}{41}{31}{61}{35}{30}
    \heatrow{7}{}{68}{1}{80}{7}{2}
    \heatrow{8}{}{27}{51}{59}{65}{60}
    \heatrow{9}{}{12}{20}{7}{41}{29}
\end{scope}

\begin{scope}[xshift=6.56cm]
    \panelheader{{\tiny Active policy rate, $\gamma=0.40$}}

    \heatrow{0}{}{0}{88}{0}{5}{6}
    \heatrow{1}{}{75}{80}{100}{27}{19}
    \heatrow{2}{}{50}{48}{100}{97}{100}
    \heatrow{3}{}{0}{0}{100}{5}{0}
    \heatrow{4}{}{25}{36}{11}{65}{19}
    \heatrow{5}{}{0}{0}{100}{5}{0}
    \heatrow{6}{}{75}{8}{100}{16}{31}
    \heatrow{7}{}{100}{0}{100}{5}{0}
    \heatrow{8}{}{50}{100}{100}{100}{100}
    \heatrow{9}{}{25}{28}{0}{70}{19}
\end{scope}

\begin{scope}[xshift=9.10cm]
    \pgfmathsetmacro{\barw}{0.16}
    \pgfmathsetmacro{\seg}{(\ytop-\ybot)/20}

    \foreach \k in {0,...,19} {
        \pgfmathsetmacro{\yy}{\ybot+\k*\seg}
        \pgfmathtruncatemacro{\pct}{round(100*\k/19)}
        \fill[fill=mdHeat!\pct!white, draw=none]
            (0,\yy) rectangle ++(\barw,\seg);
    }

    \draw[mdGrid, line width=0.25pt]
        (0,\ybot) rectangle ++(\barw,\ytop-\ybot);

    \draw[mdGrid, line width=0.25pt] (\barw,\ybot) -- ++(0.06,0);
    \node[tick] at (\barw+0.09,\ybot) {$0$};

    \draw[mdGrid, line width=0.25pt] (\barw,\ytop) -- ++(0.06,0);
    \node[tick] at (\barw+0.09,\ytop) {$1$};
\end{scope}

\end{tikzpicture}%
}
\endgroup}
\caption{Object-by-room VLM gate accounting for MatterDoor-1k.  Panel A shows
the room-object prior $\rho_{c,o}$; Panel B shows the mean crop-conditioned VLM
score $s_{c,o}$; Panel C shows the active generation/fallback rate at
$\gamma=0.40$.  Ground-truth-present queries are counted as active because they
are always retained, while ground-truth-absent queries are active only when they
pass the VLM gate.
All panels use a shared 0--1 scale.}
\label{fig:object_scene_heatmaps}
\end{figure*}

\textbf{Room-conditioned scoring.}
Doorway crops are often too narrow to consistently support a binary
object-presence decision.  We therefore score object plausibility through room context.  Given a
crop $c$, the VLM provides soft support over candidate room classes
$\mathcal R$.  The queried object $o$ is then scored by marginalizing over the
room posterior:
\begin{equation}
\label{eq:vlm_room_conditioned_gate}
s_{\rm vlm}(c,o)
=
\sum_{r\in\mathcal R}
p_\eta(r\mid c)\,\pi(o\mid r),
\end{equation}
where $p_\eta(r\mid c)$ is the crop-conditioned room posterior and
$\pi(o\mid r)$ is an offline calibrated room-object plausibility table.  The
score increases when the crop supports room hypotheses in which the queried
object is plausible.  This preserves room ambiguity while avoiding an unconditional
object prior that ignores the visual crop.

\textbf{Gate policy.}
For a threshold $\gamma$, a ground-truth-absent query receives generation when
$s_{\rm vlm}(c,o)\ge\gamma$ and is skipped otherwise.  A
ground-truth-present query is never discarded: if it falls below threshold, it
enters the ground-truth-present fallback path.  The fallback path prevents VLM
misses from removing real targets from the evaluation denominator.  The
threshold therefore controls only generation allocation, not whether a real
target remains part of the benchmark.
\Cref{fig:gamma_threshold_study} shows the threshold sweep.  The active count is
the number of retained ground-truth-present queries plus ground-truth-absent
queries that pass the gate.  The remaining curves separate skipped absent
queries, plausible generated absent queries, room-unlikely generated absent
queries, and ground-truth-present fallback cases.

\begin{figure*}[tbp]
\centering
\resizebox{0.98\textwidth}{!}{\begingroup
\definecolor{mdBlue}{RGB}{42,111,151}
\definecolor{mdSlate}{RGB}{84,96,111}
\definecolor{mdGreen}{RGB}{47,133,90}
\definecolor{mdOrange}{RGB}{217,119,6}
\definecolor{mdRed}{RGB}{190,64,62}
\definecolor{mdAxis}{RGB}{80,88,96}
\definecolor{mdGrid}{RGB}{226,230,236}
\begin{tikzpicture}[x=1cm,y=1cm]
\tikzset{mdaxis/.style={draw=mdAxis, line width=0.35pt}, mdgrid/.style={draw=mdGrid, line width=0.2pt}, mdlabel/.style={font=\scriptsize}, mdsmall/.style={font=\tiny}}
\draw[mdgrid] (0,0) -- (6.2,0);
\draw[mdaxis] (-0.05,0) -- (0,0);
\node[mdsmall, anchor=east] at (-0.12,0) {0};
\draw[mdgrid] (0,1) -- (6.2,1);
\draw[mdaxis] (-0.05,1) -- (0,1);
\node[mdsmall, anchor=east] at (-0.12,1) {200};
\draw[mdgrid] (0,2) -- (6.2,2);
\draw[mdaxis] (-0.05,2) -- (0,2);
\node[mdsmall, anchor=east] at (-0.12,2) {400};
\draw[mdgrid] (0,3) -- (6.2,3);
\draw[mdaxis] (-0.05,3) -- (0,3);
\node[mdsmall, anchor=east] at (-0.12,3) {600};
\draw[mdgrid] (0,4) -- (6.2,4);
\draw[mdaxis] (-0.05,4) -- (0,4);
\node[mdsmall, anchor=east] at (-0.12,4) {800};
\draw[mdgrid] (0,5) -- (6.2,5);
\draw[mdaxis] (-0.05,5) -- (0,5);
\node[mdsmall, anchor=east] at (-0.12,5) {1000};
\draw[mdaxis] (0.0,-0.05) -- (0.0,0);
\node[mdsmall, anchor=north] at (0.0,-0.16) {0.0};
\draw[mdaxis] (2.0,-0.05) -- (2.0,0);
\node[mdsmall, anchor=north] at (2.0,-0.16) {0.2};
\draw[mdaxis] (3.0,-0.05) -- (3.0,0);
\node[mdsmall, anchor=north] at (3.0,-0.16) {0.3};
\draw[mdaxis] (4.0,-0.05) -- (4.0,0);
\node[mdsmall, anchor=north] at (4.0,-0.16) {0.4};
\draw[mdaxis] (5.0,-0.05) -- (5.0,0);
\node[mdsmall, anchor=north] at (5.0,-0.16) {0.5};
\draw[mdaxis] (6.0,-0.05) -- (6.0,0);
\node[mdsmall, anchor=north] at (6.0,-0.16) {0.6};
\draw[mdaxis] (0,0) -- (6.25,0);
\draw[mdaxis] (0,0) -- (0,5.15);
\node[mdlabel, anchor=north] at (3.1,-0.48) {VLM threshold $\gamma$};
\node[mdlabel, rotate=90, anchor=south] at (-0.68,2.55) {query count};
\draw[mdAxis, dashed, line width=0.3pt] (4.0,0) -- (4.0,5.05);
\node[mdsmall, anchor=south] at (4.0,5.05) {$\gamma=0.40$};
\draw[mdBlue, line width=0.85pt] (0.00,5.000) -- (2.00,3.295) -- (3.00,2.765) -- (4.00,2.175) -- (5.00,1.725) -- (6.00,1.395);
\fill[mdBlue] (0.00,5.000) circle (1.45pt);
\fill[mdBlue] (2.00,3.295) circle (1.45pt);
\fill[mdBlue] (3.00,2.765) circle (1.45pt);
\fill[mdBlue] (4.00,2.175) circle (1.45pt);
\fill[mdBlue] (5.00,1.725) circle (1.45pt);
\fill[mdBlue] (6.00,1.395) circle (1.45pt);
\draw[mdSlate, line width=0.65pt] (0.00,0.000) -- (2.00,1.705) -- (3.00,2.235) -- (4.00,2.825) -- (5.00,3.275) -- (6.00,3.605);
\fill[mdSlate] (0.00,0.000) circle (1.15pt);
\fill[mdSlate] (2.00,1.705) circle (1.15pt);
\fill[mdSlate] (3.00,2.235) circle (1.15pt);
\fill[mdSlate] (4.00,2.825) circle (1.15pt);
\fill[mdSlate] (5.00,3.275) circle (1.15pt);
\fill[mdSlate] (6.00,3.605) circle (1.15pt);
\draw[mdGreen, line width=0.65pt] (0.00,3.315) -- (2.00,2.490) -- (3.00,2.005) -- (4.00,1.415) -- (5.00,0.970) -- (6.00,0.660);
\fill[mdGreen] (0.00,3.315) circle (1.15pt);
\fill[mdGreen] (2.00,2.490) circle (1.15pt);
\fill[mdGreen] (3.00,2.005) circle (1.15pt);
\fill[mdGreen] (4.00,1.415) circle (1.15pt);
\fill[mdGreen] (5.00,0.970) circle (1.15pt);
\fill[mdGreen] (6.00,0.660) circle (1.15pt);
\draw[mdOrange, line width=0.65pt] (0.00,0.000) -- (2.00,0.040) -- (3.00,0.080) -- (4.00,0.185) -- (5.00,0.315) -- (6.00,0.440);
\fill[mdOrange] (0.00,0.000) circle (1.15pt);
\fill[mdOrange] (2.00,0.040) circle (1.15pt);
\fill[mdOrange] (3.00,0.080) circle (1.15pt);
\fill[mdOrange] (4.00,0.185) circle (1.15pt);
\fill[mdOrange] (5.00,0.315) circle (1.15pt);
\fill[mdOrange] (6.00,0.440) circle (1.15pt);
\draw[mdRed, line width=0.65pt] (0.00,0.965) -- (2.00,0.085) -- (3.00,0.040) -- (4.00,0.040) -- (5.00,0.035) -- (6.00,0.015);
\fill[mdRed] (0.00,0.965) circle (1.15pt);
\fill[mdRed] (2.00,0.085) circle (1.15pt);
\fill[mdRed] (3.00,0.040) circle (1.15pt);
\fill[mdRed] (4.00,0.040) circle (1.15pt);
\fill[mdRed] (5.00,0.035) circle (1.15pt);
\fill[mdRed] (6.00,0.015) circle (1.15pt);
\draw[mdBlue, line width=0.85pt] (6.55,4.80) -- ++(0.36,0);
\fill[mdBlue] (6.73,4.80) circle (1.15pt);
\node[mdsmall, anchor=west] at (7.00,4.80) {Active (gen. or fallback)};
\draw[mdSlate, line width=0.65pt] (6.55,4.38) -- ++(0.36,0);
\fill[mdSlate] (6.73,4.38) circle (1.15pt);
\node[mdsmall, anchor=west] at (7.00,4.38) {Skipped GT-absent};
\draw[mdGreen, line width=0.65pt] (6.55,3.96) -- ++(0.36,0);
\fill[mdGreen] (6.73,3.96) circle (1.15pt);
\node[mdsmall, anchor=west] at (7.00,3.96) {Plausible absent generated};
\draw[mdOrange, line width=0.65pt] (6.55,3.54) -- ++(0.36,0);
\fill[mdOrange] (6.73,3.54) circle (1.15pt);
\node[mdsmall, anchor=west] at (7.00,3.54) {GT-present fallback};
\draw[mdRed, line width=0.65pt] (6.55,3.12) -- ++(0.36,0);
\fill[mdRed] (6.73,3.12) circle (1.15pt);
\node[mdsmall, anchor=west] at (7.00,3.12) {Unlikely absent generated};
\node[mdsmall, align=left, anchor=north west] at (6.55,2.35) {$\gamma=0.40$ keeps all 144\\{}GT-present queries, skips 565\\{}GT-absent queries, and leaves\\{}8 unlikely absent generations.};
\end{tikzpicture}
\endgroup}
\caption{Threshold sweep for the VLM gate on the 1000-query MatterDoor-1k table.
The selected operating point is $\gamma=0.40$.}
\label{fig:gamma_threshold_study}
\end{figure*}

\begin{table*}[tbp]
\centering
\footnotesize
\setlength{\tabcolsep}{5pt}
\renewcommand{\arraystretch}{1.04}
\caption{Breakpoint accounting for the VLM gate. ``Active'' counts all
ground-truth-present queries plus generated ground-truth-absent queries. ``GT
fallback'' is the subset of ground-truth-present queries below threshold.}
\label{tab:gamma_threshold_breakpoints}
\begin{tabular}{@{}lrrrrr@{}}
\toprule
$\gamma$ & Active & Skipped absent & GT fallback & Plausible gen. & Unlikely gen. \\
\midrule
0.00 & 1000 & 0   & 0  & 663 & 193 \\
0.20 & 659  & 341 & 8  & 498 & 17  \\
0.30 & 553  & 447 & 16 & 401 & 8   \\
0.40 & 435  & 565 & 37 & 283 & 8   \\
0.50 & 345  & 655 & 63 & 194 & 7   \\
0.60 & 279  & 721 & 88 & 132 & 3   \\
\bottomrule
\end{tabular}
\end{table*}

\textbf{Threshold choice.}
The threshold is chosen relative to the five-room scoring structure of the
gate.  With five target-room bins, a maximally ambiguous crop induces a
room-confusion floor of approximately $1/5=0.20$: scores near this level are
consistent with diffuse room uncertainty rather than object-specific support
behind the doorway.  We therefore treat
$\gamma=0.20$ as a permissive lower reference point, not as a suitable operating
threshold.  The selected threshold $\gamma=0.40$ requires support above this
confusion floor, so generation is reserved for queries whose
room-conditioned object plausibility is stronger than what would arise from
near-uniform room ambiguity.
The breakpoint counts show the tradeoff behind this choice
(\cref{fig:gamma_threshold_study}).
At $\gamma=0.20$, the policy keeps 659 active queries and still generates 17
room-unlikely absent cases, so many weakly supported absent queries remain
active.  Raising the threshold to $\gamma=0.40$ reduces the active set to 435,
skips 565 ground-truth-absent queries, and limits room-unlikely absent
generation to 8 cases, while preserving 283 plausible absent generations for
auditing hidden-object synthesis.  Moving higher to $\gamma=0.50$ is more
restrictive for this benchmark: it reduces plausible absent probes from 283 to
194 and increases ground-truth-present fallback from 37 to 63.  Thus
$\gamma=0.40$ is the selected operating point: it stays above the five-room
confusion floor, avoids generation for low-scoring room-ambiguous queries, and
retains plausible absent cases for evaluating commonsense hidden-object
generation.

\endgroup
\bibliography{main}

\end{document}